\documentclass{article}

\usepackage{arxiv}

\usepackage{hyperref}
\usepackage{url}

\usepackage[utf8]{inputenc} 
\usepackage[T1]{fontenc} 

\usepackage{xcolor}
\usepackage{array}
\usepackage{wrapfig}
\usepackage{booktabs}
\usepackage{multirow}
\usepackage{amsmath}
\usepackage{amsthm}
\usepackage{amssymb}
\usepackage{graphicx}

\usepackage{natbib}
\usepackage{amsfonts}       
\usepackage{nicefrac}       
\usepackage{graphicx}
\renewcommand{\citet}{\citep}
\usepackage{multirow}
\usepackage{amsmath,bm}
\usepackage{verbatim}

\usepackage{algorithm,algorithmic}

\theoremstyle{plain}
\newtheorem{lem_boglem1}{Lemma}

\newtheorem{lem_sigmalimit}[lem_boglem1]{Lemma}

\newtheorem{lem_sigmalimitb}[lem_boglem1]{Lemma}

\newtheorem{lem_sigmaconverge}[lem_boglem1]{Corollary}

\newtheorem{lem_srilem52pre}[lem_boglem1]{Lemma}

\newtheorem{lem_neginfnorm}[lem_boglem1]{Lemma}

\newtheorem{lem_srilem52hmm}[lem_boglem1]{Lemma}

\newtheorem{th_totregbndb}[lem_boglem1]{Theorem}

\usepackage[textsize=tiny]{todonotes}

\newcommand{\tsp}[0]{{\rm T}}
\newcommand{\infset}[1]{{\mathbb{#1}}}

\usepackage{subfig}

\title{BO-Muse: A human expert and AI teaming framework for accelerated experimental design}

\author{
	{Sunil Gupta$^1$\thanks{Mail Correspondence: sunil.gupta@deakin.edu.au;$\quad^\dagger$This research work was completed during Majid's affiliation with $\text{A}^2\text{I}^2$, Deakin University, Australia}  $\:$, Alistair Shilton$^1$, Arun Kumar A V$^1$, Shannon Ryan$^1$, Majid Abdolshah$^{2\dagger}$, Hung Le$^1$,} \\
		\textbf{Santu Rana$^1$, Julian Berk$^1$, Mahad Rashid$^1$, Svetha Venkatesh$^1$}  \\
	$^1$ Applied Artificial Intelligence Institute ($\text{A}^2\text{I}^2$), Deakin University, Australia \\
	$^2$ Amazon, Australia
}

\date{}

\renewcommand{\undertitle}{}
\renewcommand{\shorttitle}{BO-Muse: A human expert and AI teaming framework for accelerated experimental design}

\begin{document}

\maketitle

\begin{abstract}
In this paper we introduce BO-Muse, a new approach to human-AI teaming
for the optimization of expensive black-box functions. Inspired by
the intrinsic difficulty of extracting expert knowledge and distilling
it back into AI models and by observations of human behavior in real-world
experimental design, our algorithm lets the human expert take the
lead in the experimental process. The human expert can use their domain
expertise to its full potential, while the AI plays the role of a
muse, injecting novelty and searching for areas of weakness to break
the human out of over-exploitation induced by cognitive entrenchment.
With mild assumptions, we show that our algorithm converges sub-linearly,
at a rate faster than the AI or human alone. We validate our algorithm
using synthetic data and with human experts performing real-world
experiments.
\end{abstract}

\section{Introduction}

Bayesian Optimization (BO) \citet{shahriari2015taking} is a popular
sample-efficient optimization technique to solve problems where the
objective is expensive. It has been successfully applied in diverse
areas \citet{Greenhill_etal_20Bayesian} including material discovery
\citet{Li_etal_17rapid}, alloy design \citet{barnett2020scrap} and
molecular design \citet{gomez2018automatic}. However, standard BO
typically operates \emph{tabula rasa}, building its model of the objective
from minimal priors that do not include domain-specific information.  While
there has been some progress made incorporating domain-specific knowledge
to accelerate BO \citet{Li_etal_18Accelerating,Hva1} or transfer learning
from previous experiments \citet{Shilton_etal_17Regret}, it remains
the case that there is a significant corpus of knowledge and expertise
that could potentially accelerate BO even further but which remain
largely untapped due to the inherent complexities involved in knowledge
extraction and exploitation. In particular, this often arises from
the fact that experts tend to organize their knowledge in complex
schema containing concepts, attributes and relationships \citet{rousseau2001schema},
making the elicitation of relevant expert knowledge, both quantitative
and qualitative, a difficult task.

Experimental design underpins the discovery of new materials, processes
and products. However, experiments are costly, the target function
is unknown and the search space unclear.
To be sample-efficient, the least number of experiments must be performed.
Traditionally experimental design is guided by (human) experts who
use their domain expertise and intuition to formulate an experimental
design, test it, and iterate based on observations. Living beings
from fungi \citet{watkinson2005new} to ants \citet{pratt2006tunable}
and humans \citet{daw2006cortical,cohen2007should} face a dilemma
when they make these decisions: exploit the information they have,
or explore to gather new information.  How humans balance this 
dilemma has been studied in \cite{daw2006cortical} -- examining human
choices in a multi-arm bandit problem,  they showed that humans were highly skewed towards exploitation.  Moreover,  when the task requires specialized experts,  cognitive entrenchment is heightened and the balance between expertise and flexibility swings further towards remaining in known paradigms.

\begin{figure}
	\centering
	\includegraphics[width=0.8\columnwidth]{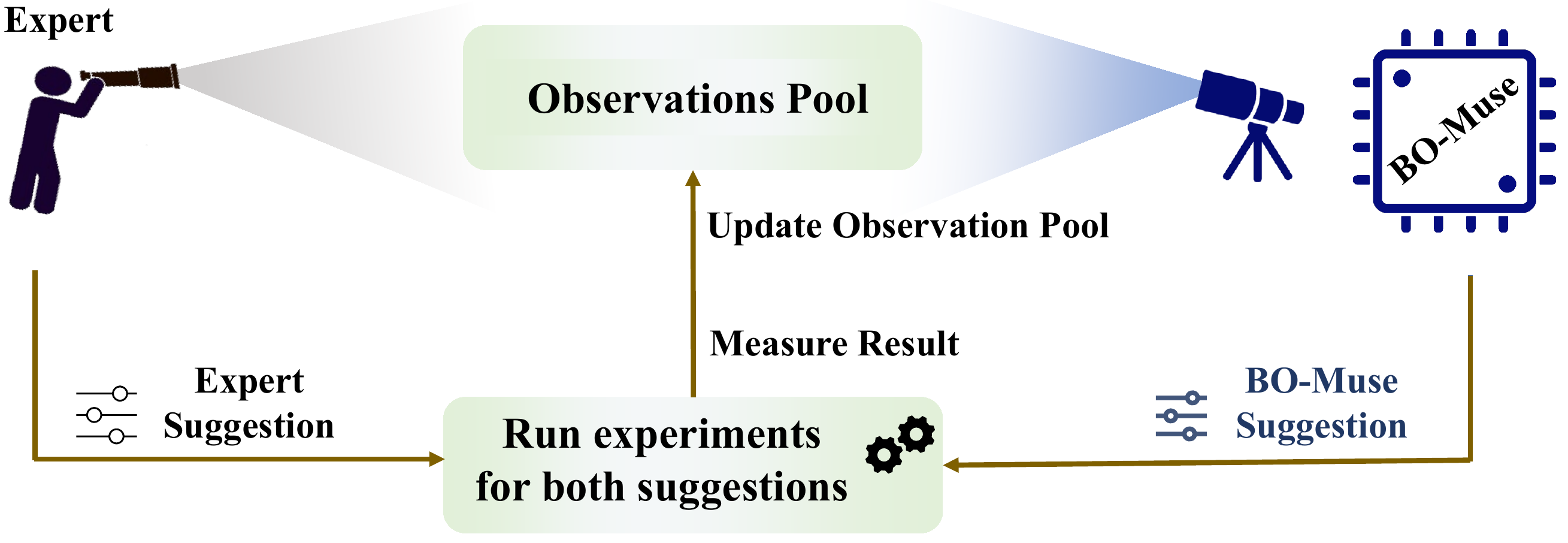}\label{fig:design}\caption{BO-Muse Workflow.}
\end{figure}

To break out of this, dynamic environments of engagement are needed
to force experts to incorporate new points of view \citet{dane2010reconsidering}.
For such lateral thinking to catalyze creativity, \cite{beaney2005imagination}
has further confirmed that external stimuli are crucial.  For example,
using random stimuli to boost creativity has been attempted in the
context of games \citet{yannakakis2014mixed}.  In Sentient Sketchbook,
a machine creates sketches that the human can refine, and sketches
are readily created through machine learning models trained on ample
data. Other approaches use machine representations to learn models
of human knowledge, narrowing down options for the human to consider.
Recently, \cite{vasylenko2021element} constructs a variational auto-encoder
from underlying patterns of chemistry based on structure/composition
and then a human generated hypothesis guides possible solutions. An
entirely different approach refines a target function by allowing
machine learning to discover relations between mathematical objects,
and guides humans to make new conjectures \citet{davies2021advancing}.
Note, however, that there is still a requirement for large datasets
to formulate representations of mathematical objects, which is antithetical
to sample-efficiency as typically in experimental design we have a
budget on the number of experiments, data from past designs is lean,
and formulation of hypothesis is difficult in this lean data space.

The use of BO for experimental design overcomes the problems of over-exploitation
and cognitive entrenchment,  and provides mathematically rigorous guarantees
of convergence to the optimal design.  However, as noted previously,
this often means that domain-specific knowledge and expertise is lost.
In this paper, motivated by our observations, rather than attempting
to enrich AI models using expert knowledge to accelerate BO,
we propose the BO-Muse algorithm that lets the human expert take the
lead in experimental design with the aid of an AI ``muse'' whose
job is to augment the expert's intuition through AI suggestions.  Thus
the AI's role is to provide dynamism to break
an expert's cognitive entrenchment and go beyond the
state-of-the-art in new problems, while the expert's
role is to harness their vast knowledge and extensive experience to
produce state-of-the-art designs. Combining these roles in a formal
framework is the main contribution of this paper. 

BO-Muse is a formal framework that inserts BO into the
expert's workflow (see Figure 1), allowing adjustment
of the AI exploit/explore strategy in response to the human expert
suggestions. This process results in a batch of suggestions from 
the human expert and the AI at each iteration. This batch of designs
is experimentally evaluated and shared with both the human and the AI. 
The AI model is updated and the process iterated until the
target is reached or the design budget is depleted.  We theoretically 
analyze the sample-efficiency of
BO-Muse and provide a sub-linear regret bound under appropriate assumptions. We validate BO-Muse
using optimization benchmarks and teaming with experts to perform
complex real-world tasks.  
Our contributions are:

\begin{itemize}
\item Design of a framework (BO-Muse) for a human expert and an AI to work
in concert to accelerate experimental design, taking advantage of
the human's deeper insight into the problem and the AI's advantage
in using rigorous models to complement the expert to achieve sample-efficiency; 
\item Design of an algorithm that compensates for the human tendency to
be overly exploitative by appropriately boosting the AI exploration;
\item Provide a sub-linear regret bound for BO-Muse to demonstrate the accelerated
convergence due to the human-AI teaming in the optimization process;
and
\item Provide experimental validation both using optimization benchmark
functions and with human experts to perform complex real-world tasks.
\end{itemize}

\section{Background}

\subsection{Human machine partnerships}

Mixed initiative creative interfaces propose a tight coupling of human
and machine to foster creativity. Thus far, however, research has
been largely restricted to game design \citep{deterding2017mixed},
where the authors identified open challenges including \textquotedblleft what
kinds of human-AI co-creativity can we envision across and beyond
creative practice?\textquotedblright . Our work is the first example
of the use of such a paradigm to accelerate experimental design. Also of
importance, though beyond the scope of this study, is the design of
interfaces for such systems \citep{rezwana2022designing} and how
the differing ways human and machine express confidence affects
performance \citep{steyvers2022bayesian}.

\subsection{Bayesian Optimization}

Bayesian Optimization (BO, \citet{Bro2}) is an optimization method for solving 
the problems of the form:
\[
 \begin{array}{c}
  {\bf x}^{\star}={\rm argmax}_{{\bf x}\in\mathbb{X}}  f^{\star} \left(\mathbf{x} \right)
 \end{array}
\]
in the least possible number of iterations when 
$f^{\star}$ is an expensive blackbox function. Bayesian optimization models 
$f^\star$ as a draw from a Gaussian Process ${\rm GP} (0,K)$ with prior 
covariance (kernel) $K$ \citep{ras1}.  At each iteration $t$, BO recommends 
the next function evaluation point ${\bf x}_t$ by optimizing a (cheap) 
acquisition function $a_t : \mathbb{X} \to \mathbb{R}$ based on the posterior 
mean and variance given a dataset of observations $\mathbb{D}_{t-1} = \{ ({\bf 
x}_i, y_i) : i \in \mathbb{N}_{t-1} \}$:
\[
 \begin{array}{rl}
  \!\!\mu_{t-1} \!\left( {\bf x} \right) &\!\!\!\!\!=\! {\bf y}_{t-1}^{\rm T} \left( {\bf K}_{t-1} + \sigma^2 {\bf I} \right)^{-1} {\bf k}_{t-1} \left( {\bf x} \right) \\
  \!\!\sigma_{t-1}^2 \!\left( {\bf x} \right) &\!\!\!\!\!=\! K_{t-1} \!\left( {\bf x}, {\bf x} \right) - {\bf k}_{t-1}^{\rm T} \!\left( {\bf x} \right) \!\left( {\bf K}_{t-1} \!+\! \sigma^2 {\bf I} \right)^{-1} \!{\bf k}_{t-1} \!\left( {\bf x} \right) \\
 \end{array}
\]
where ${\bf y}_{t-1} = [ y_i ]_{i \in \mathbb{N}_{t-1}}$ is the set of 
observed outputs, ${\bf K}_{t-1} = [ K ({\bf x}_i, {\bf x}_j) ]_{i,j 
\in \mathbb{N}_{t-1}}$ and ${\bf k}_{t-1} ({\bf x}) = [ K ({\bf x}_i, 
{\bf x}) ]_{i \in \mathbb{N}_{t-1}}$. 
Experiments evaluate $y_t = f^\star ({\bf x}_t)+\nu_t$, where $\nu_t$
is noise, the GP model is updated to include the new observation,
and the process repeats either until a convergence criteria is met
or a fixed budget of evaluations is exhausted. Several acquisition
functions exist e.g. Expected Improvement (EI, \citet{Jon1}), GP-UCB
\citet{Sri1}, Entropy search schemes \citet{hennig2012entropy,hernandez2014predictive,wang2017max}, Knowledge gradient \citet{frazier2009knowledge} etc. We use GP-UCB as it is particularly amenable for theoretical analysis.  GP-UCB uses:
\begin{equation}
 \begin{array}{l}
  a_t \left( {\bf x} \right) = \mu_{t-1} \left( {\bf x} \right) + \beta_t^{1/2} \sigma_{t-1} \left( {\bf x} \right)
 \end{array}
\end{equation}
Here $\beta_t$  is a variable controlling the trade-off between
exploitation of known maxima (if $\beta_t$ is small) and exploration
(if $\beta_t$ is large). The optimization performance depends on
$\beta_t$. To achieve sub-linear convergence \cite{Cho7} recommend
$\beta_t = \chi_t$, where: \begin{equation} \begin{array}{l} \chi_t = \left( \frac{\Sigma}{\sqrt{\sigma}} \sqrt{ 2 \ln \left( \frac{1}{\delta} \right) + 1 + \gamma_t   } + \| f^\star \|_{{\mathcal{H}}_K} \right)^2 \end{array} \label{eq:define_chi_base} \end{equation}
and $\gamma_t$ is the maximum information gain (maximum mutual information
between $f^{\star}$ and any $t$ observations \citet{Sri1}). When the 
assumption that $f^\star \sim {\rm GP} (0,K)$ (or more generally $f^\star \in 
\mathcal{H}_K$, where $\mathcal{H}_K$ is the reproducing kernel 
Hilbert space of the kernel $K$ \citep{Sri1}) is not true the problem is {\em mis-specified}.
One approach to mis-specified BO is enlarged confidence
GP-UCB (EC-GP-UCB, \citet{Bog2}). In EC-GP-UCB, the function closest
to the objective $f^\star$ at iteration $t$ is denoted $f_t^\# = {\rm argmin}_{f \in \mathcal{H}_K} \| f^\star - f \|_{\infty}$,
and the acquisition function is:
\begin{equation}
 \begin{array}{l}
  a_t \left( {\bf x} \right) = \mu_{t-1} \left( {\bf x} \right) + \left( \beta_t^{1/2} + \frac{\epsilon_t \sqrt{t}}{\sigma} \right) \sigma_{t-1} \left( {\bf x} \right)
 \end{array}
\label{eq:ecgpucbacq} 
\end{equation}
where $\epsilon_t = \| f_t^\# - f^\star \|_\infty$ is the mis-specification
gap. The additional term in the acquisition function is required to
ensure sub-linear convergence in this case.

\section{The Human Optimization Model}
\label{sec:humanmodel}


It is difficult to make general statements about how a human expert may go 
about modeling or optimizing the objective function $f^\star$.  Nevertheless 
we think it reasonable to posit that the human expert will maintain, 
explicitly or implicitly, an {\em evolving} model of $f^\star$ (note that we 
do not presume to know the details of this, only that it exists, explicitly or 
otherwise):
\[
 \begin{array}{l}
  f_t \left( {\bf x} \right) = g_t \left( {\bf p}_t \left( {\bf x} \right) \right)
 \end{array}
\]
where ${\bf p}_t : \mathbb{R}^n \to \mathbb{R}^{m_t}$ represents the human's 
understanding of relevant features $g_t : \mathbb{R}^m \to \infset{R}$ is in 
some sense simple.  Such a model is well-approximated by a GP model if we let 
the form of $g_t$ dictate the kernel $K_t$ - for example if $g_t$ is well 
approximated by a $d^{\rm th}$-order Taylor series then we may model $f_t$ as 
a draw from ${\rm GP} (0,K_t)$ with a $d^{\rm th}$ order polynomial kernel 
$K_t ({\bf x},{\bf x}') = (1+{\bf p}_t^{\rm T} ({\bf x}) {\bf p}_t ({\bf x}') 
)^d$.\footnote{Precisely, using a $d^{\rm th}$ order Taylor series 
approximation $f_t ({\bf x}) \approx {\bf w}_t^{\rm T} [\sqrt{{}^d C_q} {\bf 
p}_t^{\otimes q} ({\bf x})]_{q = 0,1,\ldots,d}$ and using the kernel trick:
\[
 \begin{array}{rl}
  K_t \left( {\bf x}, {\bf x}' \right) 
  &\!\!\!\!= \sum_{q=0}^d {}^d C_q {\bf p}_t^{\otimes q} \left( {\bf x} \right)^{\rm T} {\bf p}_t^{\otimes q} \left( {\bf x}' \right) \\
  &\!\!\!\!= \left( 1 + {\bf p}_t^{\rm T} \left( {\bf x} \right) {\bf p}_t \left( {\bf x}' \right) \right)^d \\
 \end{array}
\]}
Alternatively if the expert uses a similarity-based model then this may be 
modeled using a GP equipped with an SE or similar kernel, and so on.  We 
use a GP model here as it is both flexible enough to encompass a wide range of 
possible human models and amenable to analysis.  We do not presume to know the 
specifics of the human expert's model, and we do not assume that it suffices to 
precisely replicate $f^\star$, particularly in the earlier stages of the 
algorithm.  We do, however, assume that the human expert is capable of learning 
from observations of $f^\star$ to refine their model, closing the gap between 
their model and reality as $t$ increases.

With regard to the optimization methodology applied by the human expert it is 
challenging to say precisely what method will be used.  However, motivated by 
\citep{Bor2,dane2010reconsidering}, we believe it is reasonable to assume that 
some form of trade-off between exploitation of presumed ``good'' regions 
(colloquially, ``based on my model, this experiment should yield good 
results'') and exploration of areas of uncertainty will be applied; and 
moreover that the human will typically favor exploitation over exploration.  
Hence for the purposes of convergence analysis, acknowledging the imperfect 
nature of the human's model, we model human optimization using the EC-GP-UCB 
variant (\ref{eq:ecgpucbacq}) of the BO algorithm, where the gap $\epsilon_t$ 
and trade-off parameter $\beta_t$ are unknown but $\epsilon_t$ is assumed to 
converge toward $0$ and while we make no assumptions regarding $\beta_t$ we allow for the possibility that this may be small due to cognitive entrenchment.

Finally, it is important to note that these assumptions regarding human 
behavior are for analytical purposes only.  Even if all assumptions do not hold or 
the human expert attempts to actively sabotage convergence we can still prove that 
BO-Muse will converge with a sub-linear regret bound big-O comparable to 
standard GP-UCB.  The purpose of these assumptions is to allow us to analyze 
how convergence will be accelerated, in terms of regret, given a typical human 
expert working to the best of their abilities.

\section{Framework}

Our goal is to solve 
${\bf x}^\star = {\rm argmax}_{{\bf x} \in \mathbb{X}} f^\star ({\bf x})$, 
where $f^\star : \mathbb{X} \to \mathbb{R}$ is expensive and evaluation is 
noisy, with results (observations) $y = f^\star ({\bf x}) + \nu$, where $\nu$ is 
$\Sigma$-sub-Gaussian noise.  To do this, we follow a 
BO methodology, running a sequence of 
experimental batches, indexed by $s = 1,2,\ldots$, consisting of one human 
recommendation ${\hat{\bf x}}_s$ and one AI recommendation ${\breve{\bf 
x}}_s$ ({{\bf hatted} variables relate to the human expert, {\bf breved} 
variables to the AI}), after which human and AI update their models based on 
the new data and the process repeats until the budget of $S$ batches is 
exhausted.

As discussed in section \ref{sec:humanmodel}, for analytical purposes we 
assume the human expert maintains an implicit GP model of $f^*$ with an 
unknown kernel $\hat{K}_s$.  We also explicitly model 
$f^*$ as a draw from a GP with prior covariance $\breve{K}_s$ (the AI 
model), where $\breve{K}_s$ and $\hat{K}_s$ may be updated after each 
batch.  The posterior means and variances given dataset $\mathbb{D}_s = \{ 
({\hat{\bf x}}_i, {\hat{y}}_i), ({\breve{\bf x}}_i, {\breve{y}}_i) : 
i \leq s \}$ are:
\[
 \begin{array}{l}
 \left\{
 \begin{array}{rl}
  {  \hat{{\mu}}}_{s}        \left( {\bf x} \right) &\!\!\!\!=                                                     {      {{\bf y}}}_{s}^{\rm T}                        ( {  \hat{{\bf K}}}_{s} + \sigma^2 {\bf I} )^{-1} {  \hat{{\bf k}}}_{s} \left( {\bf x} \right) \\
  {  \hat{{\sigma}}}_{s}^{2} \left( {\bf x} \right) &\!\!\!\!= {  \hat{{K}}}_{s} \left( {\bf x}, {\bf x} \right) - {  \hat{{\bf k}}}_{s}^{\rm T} \left( {\bf x} \right) ( {  \hat{{\bf K}}}_{s} + \sigma^2 {\bf I} )^{-1} {  \hat{{\bf k}}}_{s} \left( {\bf x} \right) \\
 \end{array} \right. \\
 \left\{
 \begin{array}{rl}
  {\breve{{\mu}}}_{s}        \left( {\bf x} \right) &\!\!\!\!=                                                     {      {{\bf y}}}_{s}^{\rm T}                        ( {\breve{{\bf K}}}_{s} + \sigma^2 {\bf I} )^{-1} {\breve{{\bf k}}}_{s} \left( {\bf x} \right) \\
  {\breve{{\sigma}}}_{s}^{2} \left( {\bf x} \right) &\!\!\!\!= {\breve{{K}}}_{s} \left( {\bf x}, {\bf x} \right) - {\breve{{\bf k}}}_{s}^{\rm T} \left( {\bf x} \right) ( {\breve{{\bf K}}}_{s} + \sigma^2 {\bf I} )^{-1} {\breve{{\bf k}}}_{s} \left( {\bf x} \right) \\
 \end{array} \right. \\
 \end{array}
\]
respectively for (implicit) human and (explicit) AI models.

We assume $f^\star \in 
\breve{\mathcal{H}}_s = \mathcal{H}_{\breve{K}_s}$ lies in the RKHS of the AI's 
kernel $\breve{K}_s$.  We do not make this assumption for the human expert, 
so the is problem mis-specified from their perspective.  As discussed previously, 
for the purposes of regret (convergence) analysis, borrowing from \citep{Bog2}, we assume that for batch $s$ the human attempts to 
maximize the closest function to $f^\star$ in $\hat{\mathcal{H}}_s$:
\[
 \begin{array}{l}
  {  \hat{{f}}}_{s}^{\#} = {{\rm argmin}}_{
f \in {{  \hat{{\mathcal{H}}}}_{s}} : \left\| f \right\|_{{{  \hat{{\mathcal{H}}}}_{{s}}}} \leq {\hat{B}} 
} \left\| f - f^\star \right\|_\infty 
\\
 \end{array}
\]
where $\hat{\mathcal{H}}_s = \mathcal{H}_{\hat{K}_s}$.  
As discussed in section \ref{sec:humanmodel}, we assume the gap between 
$f^\star$ and ${\hat{{f}}}_s^{\#}$ is bounded as $\| {\hat{{f}}}_{s}^{\#} - 
f^\star \|_\infty \leq {\hat{{\epsilon}}}_{s}$, and that the expert is able to 
learn (in effect, update their kernel) to ``close the 
gap'', so $\mathop{\lim}\limits_{s \to \infty} {\hat{{\epsilon}}}_s \to 0$.

The AI generates recommendations using GP-UCB and, as discussed in section 
\ref{sec:humanmodel}, for analysis we assume the human in effect 
generates recommendations using EC-GP-UCB \citep{Bog2}, so:
\begin{equation}
{\!\!\!\!\!\!\!\!\!{
 \begin{array}{rrl}
  {  \hat{{\bf x}}}_{s} &\!\!\!\!\!\!=\! 
  \mathop{\rm argmax}\limits_{{\bf x} \in \mathbb{X}} {  \hat{{a}}}_s \!\left( {\bf x} \right) 
  &\!\!\!\!\!=\! 
  {  \hat{{\mu}}}_{s-1} \!\left( {\bf x} \right) \!+\! ( {  \hat{{\beta}}}_{s}^{1/2} \!\!+\! \frac{{\hat{{\epsilon}}}_s}{\sigma} \!\sqrt{2s} ) {  \hat{{\sigma}}}_{s-1} \left( {\bf x} \right) \\

  {\breve{{\bf x}}}_{s} &\!\!\!\!\!\!=\! 
  \mathop{\rm argmax}\limits_{{\bf x} \in \mathbb{X}} {\breve{{a}}}_s \!\left( {\bf x} \right) 
  &\!\!\!\!\!=\! 
  {\breve{{\mu}}}_{s-1} \!\left( {\bf x} \right) \!+\!   {\breve{{\beta}}}_{s}^{1/2}                                                   {\breve{{\sigma}}}_{s-1} \left( {\bf x} \right) \\
 \end{array}
}\!\!\!\!\!\!\!\!\!}
\end{equation}
We do not presume to know the precise trade-off $\hat{\beta}_s$ used by the 
human but based on \citep{Bor2,dane2010reconsidering}, as discussed previously, 
it appears likely that ${\hat{\beta}}_s$ will be small.  We therefore use the 
AI trade-off ${\breve{{\beta}}}_s$, which we control, to compensate by adding 
much needed exploration.

It is convenient to specify the exploration/exploitation trade-off parameters 
${\hat{\beta}}_s$ and ${\breve{\beta}}_s$ relative to 
(\ref{eq:define_chi_base}) \citep{Cho7,Bog2}.  Without loss of generality we 
require that $\hat{\beta}_s \in (\hat{\zeta}_{s\downarrow} \hat{\chi}_s, 
\hat{\zeta}_{s\uparrow} \hat{\chi}_s )$ and ${{\breve{{\beta}}}_{s}} = 
{\breve{{\zeta}}}_{s} {{\breve{{\chi}}}_{s}}$, where $0 \leq {\hat{\zeta}}_{s 
\downarrow} \leq 1 \leq {\hat{\zeta}}_{s\uparrow} \leq \infty$, 
${\breve{{\zeta}}}_{s} \geq 1$ and:
\begin{equation}
 {\!\!\!\!\!\!{
 \begin{array}{l}
  {  \hat{{\chi}}}_{s} = \big( \frac{\Sigma}{\sqrt{\sigma}} \sqrt{ 2 \ln \left( {1}/{\delta} \right) + 1 + {  \hat{{\gamma}}}_{s}
  } + \| {  \hat{{f}}}_{s}^{\#} \|_{{  \hat{{\mathcal{H}}}}_{s}} \big)^2 \\
  {\breve{{\chi}}}_{s} = \big( \frac{\Sigma}{\sqrt{\sigma}} \sqrt{ 2 \ln \left( {1}/{\delta} \right) + 1 + {\breve{{\gamma}}}_{s}
  } + \| f^\star                \|_{{\breve{{\mathcal{H}}}}_{s}} \big)^2 \\
 \end{array}
 }\!\!\!\!\!\!}
\label{eq:chichidefdef}
\end{equation}
where ${\hat{{\gamma}}}_{s}$ and ${\breve{{\gamma}}}_{s}$ are, respectively, 
the max information-gain for human expert and the AI.  We show in section 
\ref{sec:converge} that the this suffices to ensure sub-linear convergence for 
any human expert selections, and moreover that the convergence rate can be 
improved beyond what can be achieved by standard GP-UCB if the human operates 
as described here.

\subsection{The BO-Muse Algorithm}

The proposed BO-Muse method is shown in algorithm \ref{mainalg}, where 
$f^\star$ is optimized using a sequence of batches $s = 1,2,\ldots,S$, each 
containing one human and one AI recommendation, respectively ${\hat{\bf x}}_s$ 
and ${\breve{\bf x}}_s$.  The AI recommendation minimizes the GP-UCB acquisition 
function on the AI's GP posterior, with the exploration/exploitation trade-off 
${\breve{\beta}}_s$ given (see section \ref{sec:betaselect}).  
The human recommendation is assumed to be {\em implicitly} selected to 
minimize the EC-GP-UCB acquisition function (\ref{eq:ecgpucbacq}), where the 
exploitation/exploration trade-off ${\hat{\beta}}_s$ is unknown 
but assumed to lie in the range $\hat{\beta}_s \in ( \hat{\zeta}_{s \downarrow} 
\hat{\chi}_s, \hat{\zeta}_{s\uparrow} \hat{\chi}_s)$, and the gap 
${\hat{\epsilon}}_s$ is unknown but assume to converge to $0$ at the rate 
specified in theorem \ref{th:totregbndb_loc}).

We use two approximations in our definition of ${\breve{\beta}}_s$.  
Following the standard practice, we approximate the max information gain as:
\begin{equation}
 \begin{array}{l}
  {\breve{\gamma}}_s = \sum_{\left({\bf x},\ldots \right) \in \mathbb{D}_{s-1}} \ln \left( 1+\sigma^{-2} {\breve{\sigma}}_i^2 \left( {\bf x} \right) \right)
 \end{array}
 \label{eq:migapprox}
\end{equation}
and we approximate the 
RKHS norm bound ${\breve{B}} = \| f^\star \|_{{\mathcal{H}}_{s}}$ 
as $\| {\breve{\mu}}_s \|_{{\mathcal{H}}_{s}}$, which is updates 
using $\max$ to ensure it is non-decreasing with $s$ (this 
will tend to over-estimate the norm, but this should not affect the 
algorithm's rate of convergence).  Finally, for simplicity we assume 
that $\sigma = \Sigma$ (the model parameter matches the noise).

\begin{algorithm}

\caption{BO-Muse}

\begin{algorithmic}[H]

\STATE \textbf{Input:} Initial observations $\mathbb{D}_0 = \left\{ \left( {\bf x}_{1}, y_1 \right), \left( {\bf x}_{2}, y_2 \right), \ldots \right\}$, AI prior ${\rm GP} (0,K)$.  Let $\breve{B}=1$.
\FOR {$s \in 1,2,\ldots,S$}
\STATE Set ${\breve{\beta}}_s = 7 \big( \sqrt{\sigma} \sqrt{ 2 \ln ( {1}/{\delta} ) + 1 + {\breve{\gamma}}_s } + \breve{B} \big)^2$ as per (\ref{eq:chichidefdef}), (\ref{eq:migapprox}) and (\ref{eq:approxphibnd}).
\STATE AI recommends ${\breve{\bf x}}_s$ as ${\breve{\bf x}}_s = {\rm argmax}_{{\bf x} \in \mathbb{X}} {\breve{a}}_s \left( {\bf x} \right) = {\breve{\mu}}_{s-1} \left( {\bf x} \right) + {\breve{\beta}}_{s}^{1/2} {\breve{\sigma}}_{s-1} \left( {\bf x} \right)$.
\STATE Human recommends  ${\hat{\bf x}}_s$.
\STATE Run experiments to obtain ${\hat{y}}_s = f^\star \left( {\hat{\bf x}}_s \right) + \hat{\nu}_s$ and ${\breve{y}}_s = f^\star \left( {\breve{\bf x}}_s \right) + \breve{\nu}_s$.
\STATE Set $\mathbb{D}_s = \mathbb{D}_{s-1} \cup \left\{ \left( {\hat{\bf x}}_s, {\hat{y}}_s \right), \left( {\breve{\bf x}}_s, {\breve{y}}_s \right) \right\}$ and update the AI GP posterior.
\STATE Set $\breve{B} = \max \{ \breve{B}, {\bf y}_s^\tsp \left( {\bf K}_s + \sigma^2 {\bf I} \right)^{-1} {\bf y}_s \}$.
\ENDFOR

\end{algorithmic}
\label{mainalg}
\end{algorithm}

\subsection{Convergence and Regret Bounds} \label{sec:converge}

We now discuss the convergence properties of BO-Muse.  Our goal here 
is twofold: first we show that BO-Muse converges even in the worst-case where 
the human operates arbitrarily; and second, assuming the human behaves 
according to our assumptions, we analyze how convergence is accelerated.  
Our approach is based on regret analysis \citep{Sri1,Cho7,Bog2}.  As 
experiments are batched we use instantaneous regret per 
batch, not per experiment.  The instantaneous regret for batch $s$ is defined as:
\[
 \begin{array}{ll}
  {r}_s = \min \left\{ 
f^\star \left( {\bf x}^\star \right) - f^\star \left( {  \hat{{\bf x}}}_s \right), 
f^\star \left( {\bf x}^\star \right) - f^\star \left( {\breve{{\bf x}}}_s \right)
\right\} \\
 \end{array}
\]
and the cumulative regret up to and including batch $S$ is $R_S = {\sum}_{s 
=1}^{S} {r}_s$.  If $R_S$ grows sub-linearly then the minimum instantaneous 
regret will converge to $0$ as $S \to \infty$.

Consider first the worst-case scenario, i.e. an arbitrary, non-expert human.  
In this case the BO-Muse algorithm \ref{mainalg} effectively 
involves the AI using a GP-UCB acquisition function (the constant 
scaling factor on $\breve{\beta}_s$ does not change the convergence properties 
of GP-UCB) to design a sequence of experiments $\breve{\bf x}_s$, where each 
experiment costs twice as much as usual to evaluate and yields two 
observations, $(\breve{\bf x}_s, \breve{y}_s = f^\star (\breve{\bf x}_s) + 
\breve{\eta}_s)$ and $(\hat{\bf x}_s, \hat{y}_s = f^\star (\hat{\bf x}_s) + 
\hat{\eta}_s)$, where $\hat{\bf x}_s$ is arbitrary.  Additional observations 
can only improve the accuracy of the posterior, so using standard methods 
(e.g. \cite{Cho7,Bog2,Sri1}), we see that $R_S = \mathcal{O} (B\sqrt{ 
\breve{\gamma}_{2S} 2S}+\sqrt{(\ln ({1}/{\delta})+\breve{\gamma}_{2S}) 
\breve{\gamma}_{2S} 2S})$, which is sub-linear for well-behaved kernels.


Thus we see that the worst-case convergence of BO-Muse is the same, in the 
big-O sense, as that of GP-UCB.  However this is pessimistic: in reality we 
assume a human is generating experiments $\hat{\bf x}_s$ using an EC-GP-UCB 
style trade-off between exploitation and exploration, and moreover that the 
human is an expert with an implicit or explicit evolving model of the system 
that is superior to the AI's generic prior.  For this case we have the 
following result:
\begin{th_totregbndb}
 Fix $\delta > 0$, $\theta_\downarrow > 0$, $\theta_s \in [\theta_\downarrow, 
 \infty]$, $S \in \infset{N}_+$.  Assume ${\hat{{\zeta}}}_{s\downarrow} \in 
 (0,1]$, ${\breve{\zeta}}_s = {\breve{{\zeta}}} \geq 1$ so:
 \begin{equation}
  \begin{array}{r}
   M_{\theta_s} \big(
   \exp ( {{  \hat{\chi}}_{s}^{1/2}} ( 1 - {{  \hat{{\zeta}}}_{s\downarrow }^{1/2}} )_+ {{\sigma}}_{s-1  \uparrow} ), \ldots \;\;\;\;\;\;\;\;\;\;\;\;\;\; \\
   \exp ( {{\breve{\chi}}_{s}^{1/2}} ( 1 - {{\breve{{\zeta}}}_{            }^{1/2}} )_- {{\sigma}}_{s-1\downarrow} )
   \big)
   \leq 1
  \end{array}
  \label{eq:theta_constrain}
 \end{equation}
 for all $s \leq S$, where 
 ${{\sigma}}_{s-1  \uparrow} = \max ( {\hat{{\sigma}}}_{s-1} ({\bf x}^\star), {\breve{{\sigma}}}_{s-1} ({\bf x}^\star))$, 
 ${{\sigma}}_{s-1\downarrow} = \min ( {\hat{{\sigma}}}_{s-1} ({\bf x}^\star), {\breve{{\sigma}}}_{s-1} ({\bf x}^\star))$, 
 $M_\theta (a,b) = (\frac{1}{2} (a^\theta + b^\theta))^{1/\theta}$ is the generalized mean, and ${\hat{\chi}}_{s}, {\breve{\chi}}_{s}$ as per (\ref{eq:chichidefdef}). 
 If ${\hat{\epsilon}}_{2s} = \mathcal{O} (s^{-\frac{1}{2}})$ then:
 \begin{equation}
  {\!\!\!\!\!\!\!\!\!\!{
  \begin{array}{r}
   {\frac{1}{S} {{R}}_S}
   \leq 
   \ln \! \Big( \! M_{-\theta_\downarrow} \! \Big(
   \! \exp \! \Big( \!
   \sqrt{ {  \hat{C}}
   \frac{{  \hat{{\chi}}}_{S} {  \hat{{\gamma}}}_{S}}{S}
   \!+\! \mathcal{O} \left( \frac{1}{S} \right) } 
   \Big)\!, \ldots \;\;\;\;\;\;\;\;\;\; \\
   \exp \! \Big(
   \sqrt{ {\breve{C}} 
   \big( \frac{1}{2} ({1+{{\breve{{\zeta}}}^{1/2}}}) \big)^2 \frac{{\breve{{\chi}}}_{S} {\breve{{\gamma}}}_{S}}{S} }
   \Big) 
   \! \Big) \! \Big)
  \end{array}
  }\!\!\!\!\!}
  \label{eq:regret_bound_form}
 \end{equation}
 with probability $\geq 1-\delta$, where ${\hat{C}}, {\breve{C}} > 0$ are constants.
\label{th:totregbndb_loc}
\end{th_totregbndb}
\textbf{We present the proof of this theorem in the appendix}.  In effect, 
this theorem tells us that the regret bound for BO-Muse is the average 
(generalized mean) of the regret bounds resulting from GP-UCB governed by the 
AI's model (ie. governed by the max information gain $\breve{\gamma}_S$ 
associated with the AI's kernel) and the same bound but governed by the human 
expert's (implicit) kernel (through the max information gain $\hat{\gamma}_S$).  
Assuming for the moment that the latter bound is superior to the former (that 
is, $\hat{\gamma}_S < \breve{\gamma}_S$), the ``free parameter'' $\theta_s$ 
controls the degree to which the superior regret bound dominates the inferior 
bound.  If $\theta_s$ is large then the overall regret will approach the 
superior regret bound ($M_{-\theta} (a,b) \to \min(a,b)$ as 
$\theta \to \infty$), and if $\theta_s$ is small then the bound 
will be a mix of both.  The caveat here is that, when $\theta_s$ is large, 
the AI trade-off parameter ${\breve{\zeta}}$ must be large to satisfy 
(\ref{eq:theta_constrain}), which corresponds to a very explorative AI, so 
while the regret bound may be asymptotically superior in this case, 
the factor $(\frac{1}{2} (1 + {\breve{\zeta}}^{1/2}))^2$ will also 
be large, so the regret bound may only be superior for very large $S$.  The 
``ideal'' trade-off is unclear, but the important observation from this 
theorem is that the human's expertise, and subsequent superior maximum 
information gain, will improve the regret bound and thus convergence.

With regard to max information gain, we can reasonably assume that the 
human expert (the expertise is important) starts with a better understanding 
of $f^\star$ than the AI - the underlying physics of the system, the 
behavior one might expect in similar experiments, etc.  So the 
human expert may begin with an incomplete but informative set of features that 
relate to their knowledge of the system or similar systems, or an 
understanding of the covariance structure of design space, so:
\vspace{-\topsep}
\begin{itemize}
  \setlength\itemsep{0em}
 \item The prior variance of the human expert's GP will vary between a 
       zero-knowledge base level in regions that are a mystery to the expert, 
       and much lower in regions where the expert has a good understanding 
       from past experience, understanding of underlying physics etc.
 \item The prior covariance of the expert's GP will 
       have a structure informed by the expert's understanding and knowledge, 
       for example, of the ``region A will behave like region B as they 
       have feature/attribute C in common'' type, so an experiment in region 
       A will reduce the human expert's posterior variance both in regions.
\end{itemize}
\vspace{-\topsep}
By comparison, vanilla BO starts with a generic kernel prior - SE, Matern or 
similar - that is ``flat'' over the design space, with no areas of lower 
prior variance and no non-trivial ``region A will behave like B'' behavior, so that an 
experiment will only reduce the AI's local variance.\footnote{We do not 
consider the possibility of transfer learning of this structure, as this 
requires the expert to distill their knowledge in an amenable form which, 
as noted previously, is highly non-trivial.}  Thus as the algorithm 
progresses an expert's posterior variance will start from a lower prior 
and decrease more quickly than standard BO.  Max information gain is 
bounded as the sum of the logs of the pre-experiment posterior variance 
\citep[Lemma 5.3]{Sri1}, so we may reasonably assume that the expert's 
max information gain will be lower than standard BO.  Finally, we argue 
that, as the human expert's kernel is built on relatively few, highly 
informative features, $\| f^{\#}_t \|_{{\hat{K}}_t} = \| \hat{\bf w} 
\|_2$ (the RKHS norm of the human model) will be less than $\| f^\star 
\|_{\breve{K}_t} = \| \breve{\bf w} \|_2$ (the corresponding RKHS norm using 
the standard BO's kernel).

%
%
%



\subsection{Tuning Parameter Selection} \label{sec:betaselect}

We now consider the selection of the tuning parameter ${\breve{\beta}}_s$ to 
ensure faster convergence than standard BO alone.  The conditions that ${\breve{\beta}}_s$ 
must meet are specified by (\ref{eq:theta_constrain}).  Equivalently, 
taking the maximally pessimistic view of human under-exploration 
(i.e. ${\hat{\zeta}}_{s \downarrow} = 0$), we require that 
  $\exp ( \theta_s \sqrt{{\breve{{\chi}}}_{s}} ( 1 - {{\breve{{\zeta}}}_{s}^{1/2}} ) {{\sigma}}_{s-1\downarrow} )
  \leq 2 -
  \exp ( \theta_s \sqrt{{  \hat{{\chi}}}_{s}} {{\sigma}}_{s-1 \uparrow} )$ 
or, equivalently,
  ${\breve{{\zeta}}}_s
  \geq 
  \big( 1 + 
  (\frac{{{  \hat{{\chi}}}_{s}}}{{{\breve{{\chi}}}_{s}}})^{1/2} \frac{{{\sigma}}_{s-1 \uparrow}}{{{\sigma}}_{s-1\downarrow}} 
  \frac{1}{\phi_s} \ln \big( \frac{1}{2 - e^{\phi_s}} \big) \big)^2$, 
where\footnote{We need $\phi_s > 0$ to ensure $\theta_s > 0$, and note that 
$h (\phi) = \frac{1}{\phi} \ln (\frac{1}{2-e^\phi})$ is increasing and $h(\ln 2) 
= \infty$.} $\phi_s = \theta_s \sqrt{{  \hat{{\chi}}}_{s}} {{\sigma}}_{s-1 
\uparrow} \in (0,\ln 2)$, and, noting that the right of the inequality is 
strictly increasing, $\phi_s = 0$ corresponds to ${\breve{\zeta}}_s 
= 1 + (\frac{{{  \hat{{\chi}}}_{s}}}{{{\breve{{\chi}}}_{s}}})^{1/2} 
\frac{{{\sigma}}_{s-1 \uparrow}}{{{\sigma}}_{s-1\downarrow}}$ and $\phi_s = \ln 2$ 
to ${\breve{\zeta}}_s = \infty$.


As discussed in section \ref{sec:humanmodel}, it is reasonable to assume that 
the max information gains satisfy ${\breve{\gamma}}_s \geq {\hat{\gamma}}_s$, 
and that $\| f^{\#}_s \|_{{\hat{K}}_s} \leq \| f^\star \|_{{\breve{K}}_s}$.  
With these assumptions ${\sqrt{{\hat{{\chi}}}_{s}}} \leq {\sqrt{{\breve{{\chi}}}_{s}}}$.  
Furthermore we prove in the Appendix that $\lim_{s \to \infty} \frac{\sigma_{s-1 
\uparrow}}{\sigma_{s-1\downarrow}} = 1$, so we 
approximate the bound on ${\breve{\zeta}}_s$ as ${\breve{{\zeta}}}_s \geq (1 + 
\frac{1}{\phi_s} \ln (\frac{1}{2 - e^{\phi_s}}))^2$.  Recalling $\phi_s \in (\ln 1, 
\ln 2)$ we finally, somewhat arbitrarily select $\phi_s = \ln {3}/{2}$ 
(the middle of the range in the log domain), which leads to the following heuristic 
used in the BO-Muse algorithm:
\begin{equation}
 \begin{array}{r}
  {\breve{{\zeta}}}_s
  \geq 
  \left( 1 + \frac{\ln 2}{\ln {3}/{2}} \right)^2 \approx 7
 \end{array}
\label{eq:approxphibnd}
\end{equation}

\section{Experiments}

We validate the performance of our proposed BO-Muse algorithm in the
optimization of synthetic benchmark functions, and the real-world
tasks involving human experts. In all our experiments, we have used
Squared Exponential (SE) kernel with associated
hyper-parameters estimated using maximum-likelihood estimation. We measure
the sample-efficiency of BO-Muse framework and other
standard baselines in terms of the \emph{simple regret} ($r_{t}$):
$r_{t}=f^{\star}(\mathbf{x}^{\star})-\max_{\mathbf{x}_{t}\in\mathcal{D}_{t}}f^{\star}(\mathbf{x}_{t})$,
where $f^{\star}(\mathbf{x}^{\star})$ is the true global optima and
$f^{\star}(\mathbf{x}_{t})$ is the best solution observed in $t$
iterations. Experiments were run on an Intel Xeon CPU@ 3.60GHz workstation with 16 GB RAM capacity. 

\subsection{Experiments with Optimization Benchmark Functions}

We have evaluated BO-Muse on synthetic test functions covering a range
of dimensions, as detailed in Table \ref{tab:syn_func_details}. We
compare the sample-efficiency of BO-Muse with\textbf{ (i) Generic
BO:} A standard GP-UCB based BO algorithm with the exploration-exploitation
trade-off factor ($\beta$) set as per \cite{Sri1}; \textbf{(ii)
Simulated Human: }A simulated human with access to higher level properties
(refer to Section 3.1 and high level features in Table \ref{tab:syn_func_details})
that may help to model the optimization function more accurately;
and \textbf{(iii)} \textbf{Simulated Human + PE}: A simulated human
teamed with an AI agent using a pure exploration strategy (that is,
an AI policy with $\breve{\zeta}_{s}=\infty$). To simulate a human
expert (with high exploitation), we use a standard BO algorithm with
small exploration factor maximizing $\hat{a}_{s}({\bf x})=\hat{\mu}_{s}({\bf x})+0.001\hat{\sigma}_{s}({\bf x})$.  Furthermore, we have ensured to allocate the same function evaluation budget for all the competing methods.
For a $d$ dimensional problem, we use $d+1$ initial observations and optimize for $10\times d$ iterations i.e., the budget allocated for our synthetic experiments is set to $10\times d$ function evaluations. 


\begin{table}
\centering{}\caption{Synthetic benchmark functions. Analytical forms are provided in the
second column and the last column depicts the high level features
used by a simulated human expert.\label{tab:syn_func_details}} 
{\small{}}%
\begin{tabular}{ccc}
\toprule 
{\scriptsize{}Functions} & {\small{}$f^{\star}(\mathbf{x})$} & {\small{}High Level Features}\tabularnewline
\midrule
\multirow{2}{*}{{\scriptsize{}Matyas-2D}} & \multirow{2}{*}{{\footnotesize{}$0.26*(x_{1}^{2}+x_{2}^{2})-0.48*(x_{1}*x_{2})$}} & {\small{}$x'_{1}=x_{1}^{2}$},{\small{} $x'_{2}=x_{2}^{2}$}\tabularnewline
&  & {\small{}$x'_{3}=x_{1}*x_{2}$}\tabularnewline
\midrule
\multirow{2}{*}{{\scriptsize{}Ackley-4D}} & \multirow{2}{*}{{\small{}$\!\!\!\!$$-ae^{-b\sqrt{\frac{1}{d}||x||_{2}^{2}}}-e^{\sqrt{\frac{1}{d}\sum_{i}\cos(cx_{i})}}+a+e^{1}$$\!\!$}} & {\footnotesize{}$\!\!\!\!$$x'_{1}=\cos(x_{1})$, $x'_{2}=\cos(x_{2})$$\!\!$}\tabularnewline
 &  & {\footnotesize{}$\!\!\!\!$$x'_{3}=\cos(x_{3})$,$x'_{4}=\cos(x_{4})$,$x'_{5}=||\mathbf{x}||_{2}$$\!\!$}\tabularnewline
\midrule
\multirow{2}{*}{\scriptsize{}Rastrigin-5D} & \multirow{2}{*}{\small{}$10d+\sum_{i}^{d}[{x}_{i}^{2}-10\cos(2\pi{x}_{i})]$ 
$\forall i\in\mathbb{N}_{5}$} & ${x}'_{i}={x}_{i}^{2}$ $\forall i\in\mathbb{N}_{5}$\tabularnewline
 &  & ${x}'_{j+5}=\cos{x}_{j}$ $\forall j\in\mathbb{N}_{5}$\tabularnewline
\midrule
\multirow{3}{*}{{\scriptsize{}Levy-6D}} & {\footnotesize{}$\sin^{2}(\pi w_{1})+s+(w_{d}-1)^{2}[1+\sin^{2}(2\pi w_{d})]$} & {\small{}$x'_{1}=(\sin x_{1})^{2}$,}\tabularnewline
 & {\footnotesize{}where $s=\sum_{i}^{d-1}(w_{1}-1)^{2}[1+10\sin^{2}(\pi w_{i}+1)]$} & {\small{}$x'_{j+1}=x_{j}^{2}(\sin x_{j})^{2}$}\tabularnewline
 & {\footnotesize{}and $w_{i}=1+\frac{\mathbf{x}_{i}-1}{4}$ $\forall i\in\mathbb{N}_{6}$} & {\small{}$\forall j\in\mathbb{N}_{6}$}\tabularnewline
\bottomrule
\end{tabular}
\end{table}

\begin{figure}
\begin{centering}
\subfloat[]{\includegraphics[width=0.47\columnwidth]{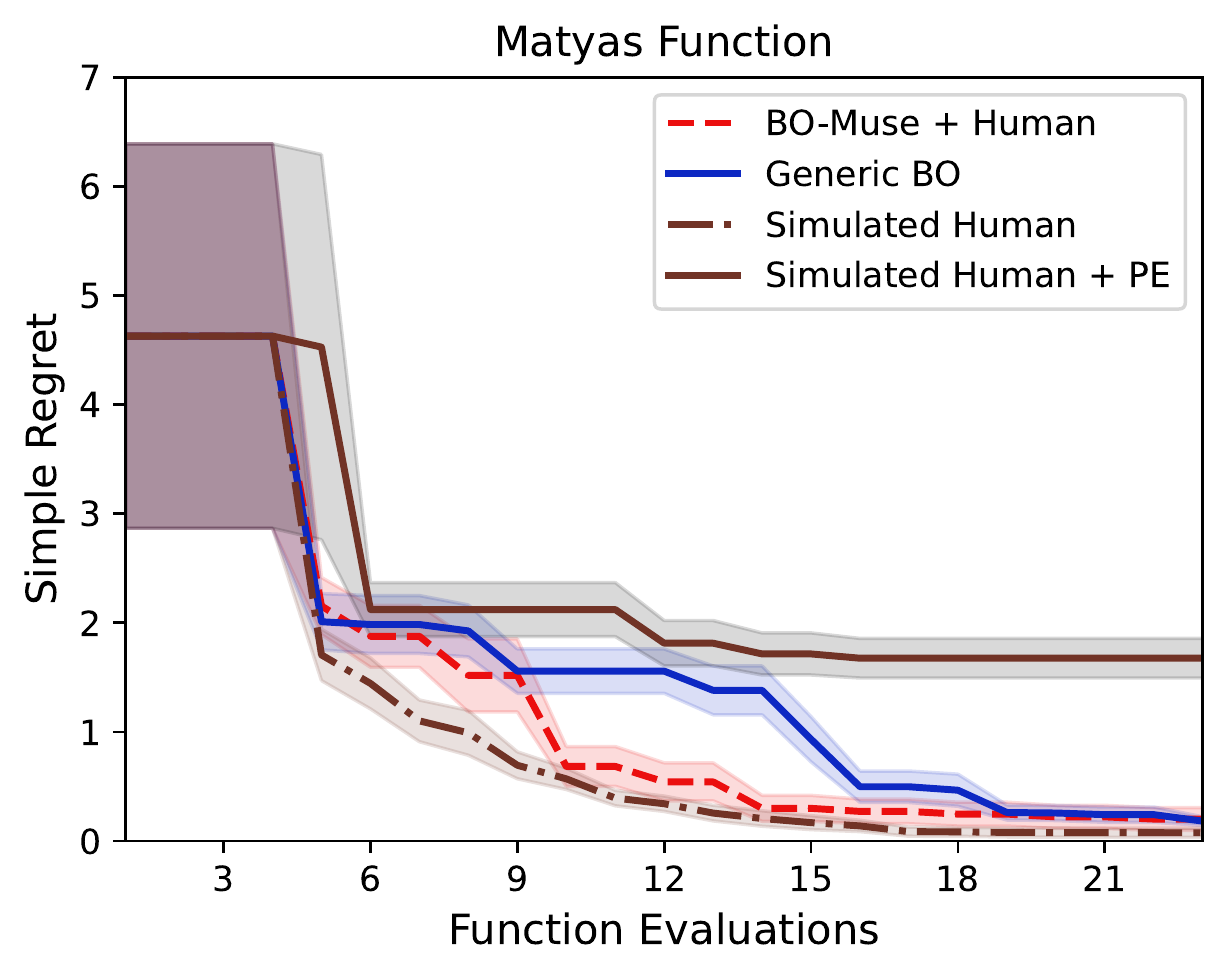}\label{fig:synthetic-matyas}}
\subfloat[]{\includegraphics[width=0.47\columnwidth]{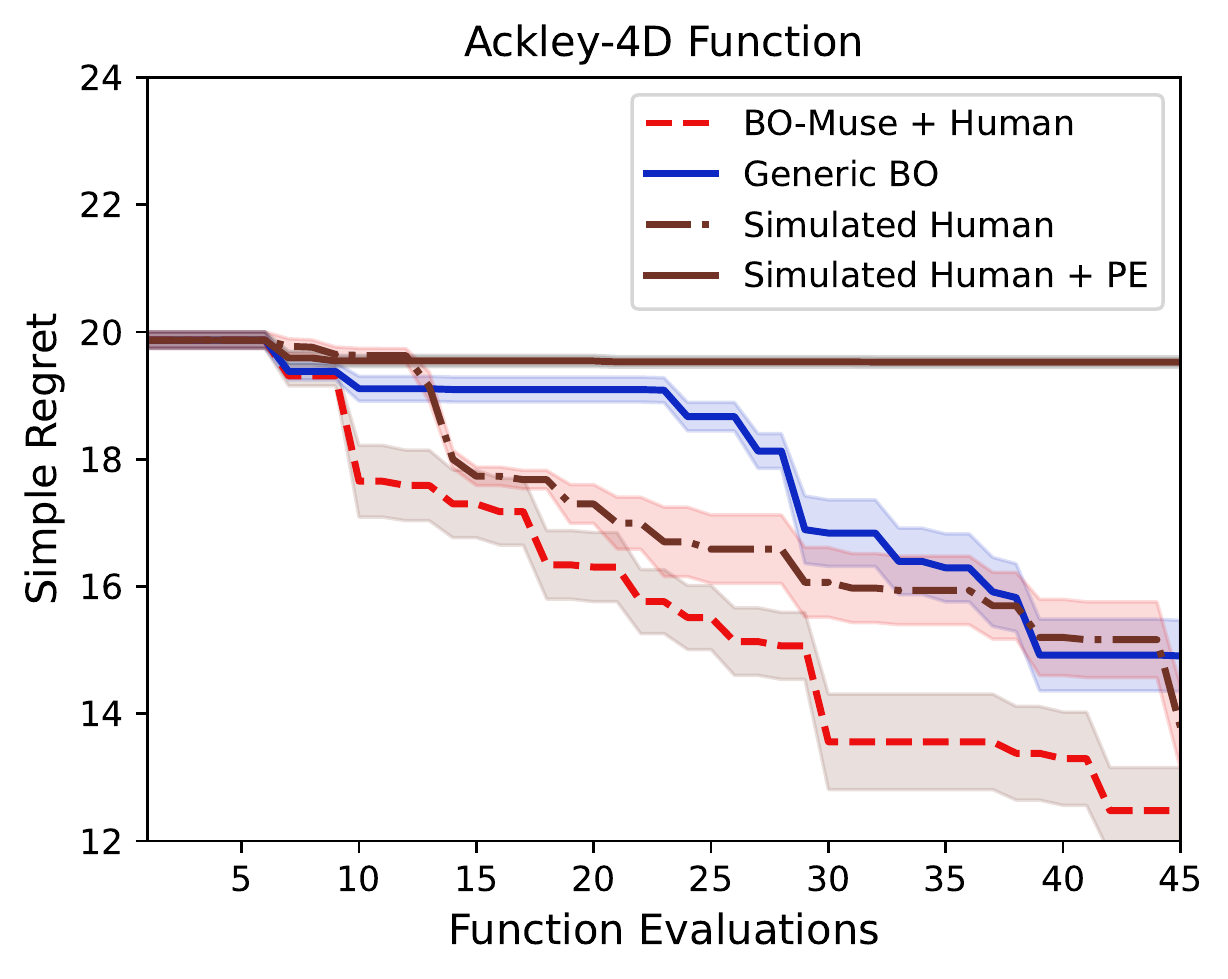}\label{fig:synthetic-ackley}}\\

\subfloat[]{\includegraphics[width=0.47\columnwidth]{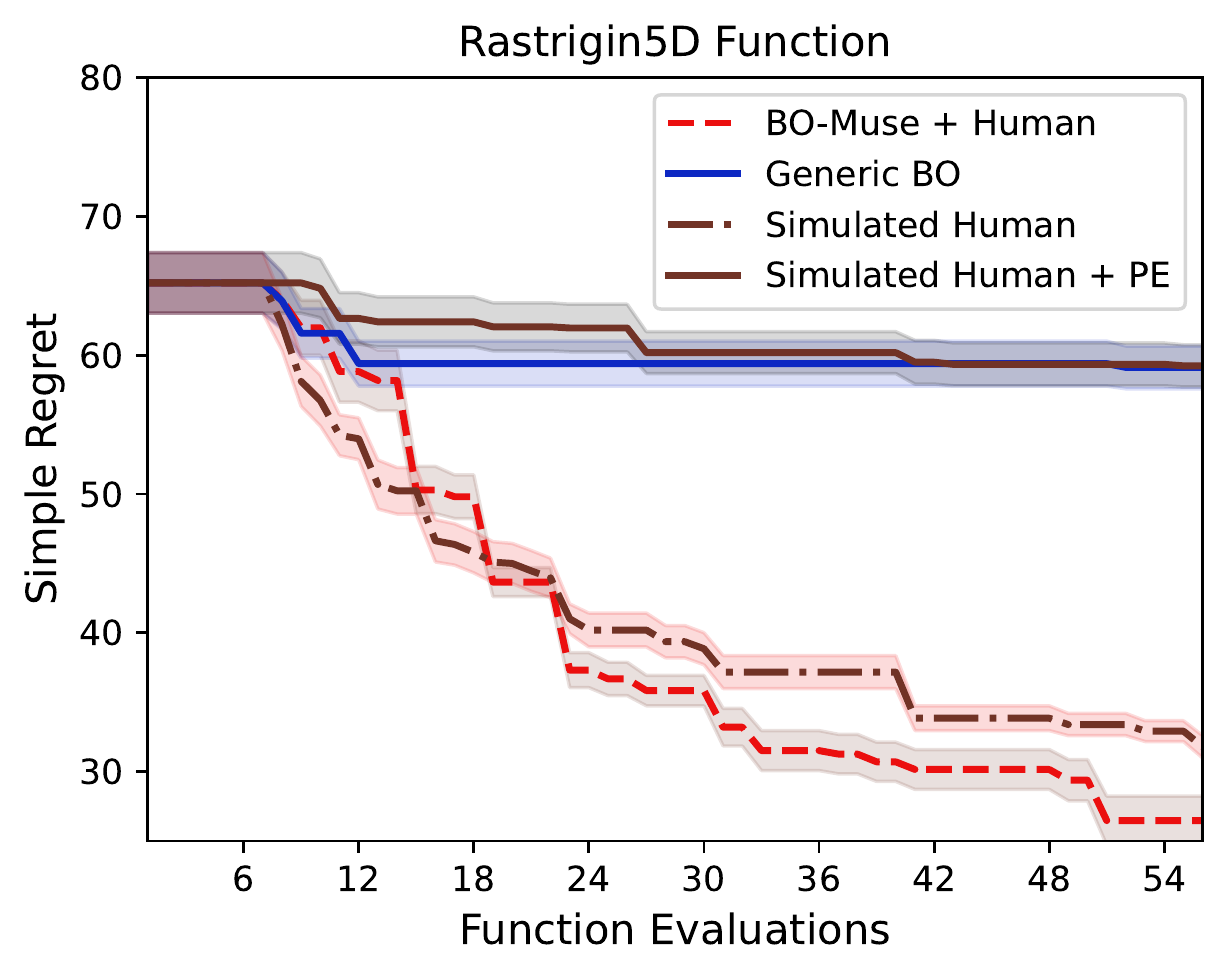}\label{fig:synthetic-rastrigin}}
\subfloat[]{\includegraphics[width=0.47\columnwidth]{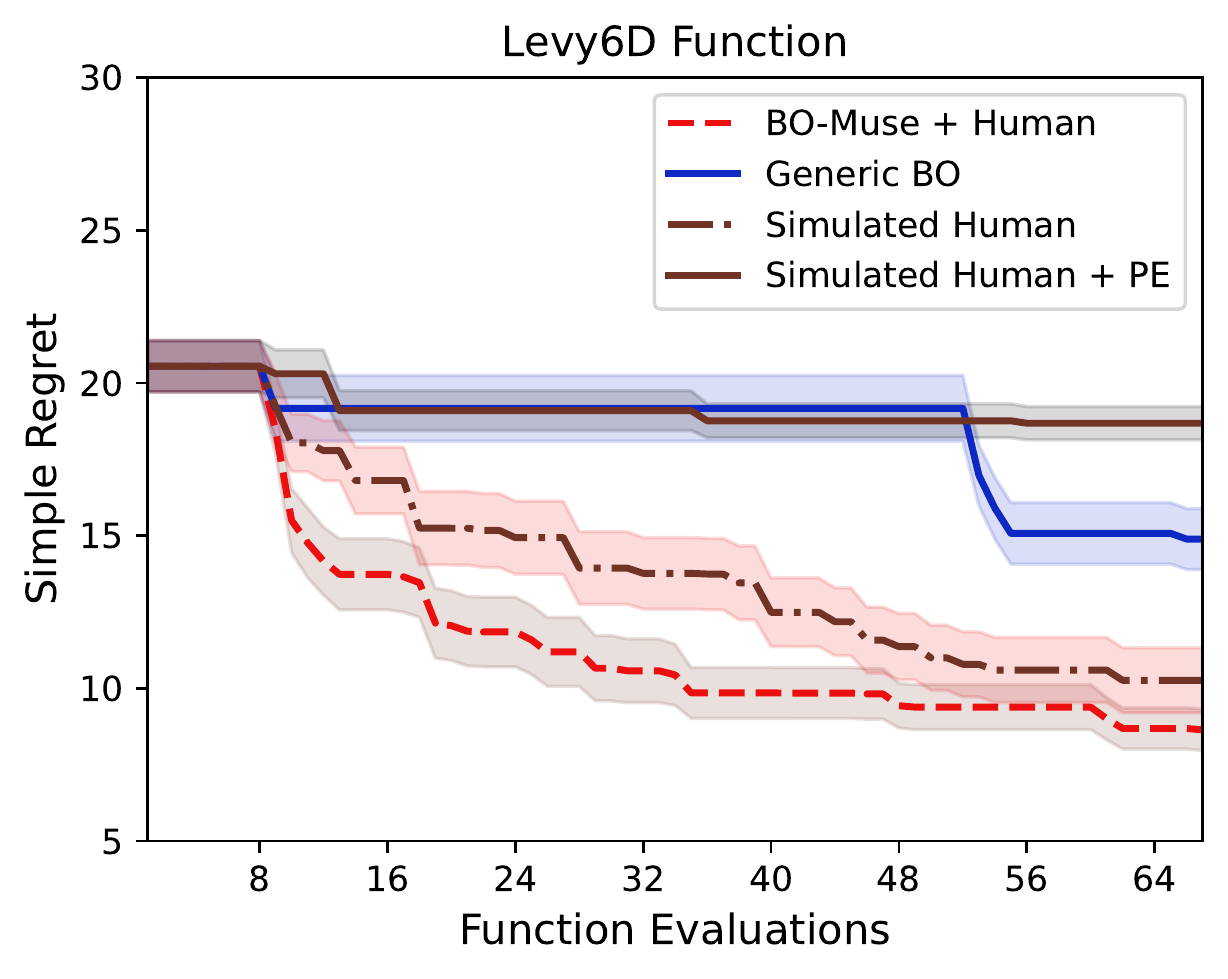}\label{fig:synthetic-levy}}

\caption{Simple regret versus iterations for (a) Matyas-2D, (b) Ackley-4D,  (c) Rastrigin-5D,  and (d) Levy-6D functions. We plot the mean regret along with its standard error obtained after 10 random repeated runs. \label{fig:synthetic-1}}

\end{centering}
\end{figure}

\begin{figure}
\begin{centering}
\subfloat[]{\includegraphics[width=0.47\columnwidth]{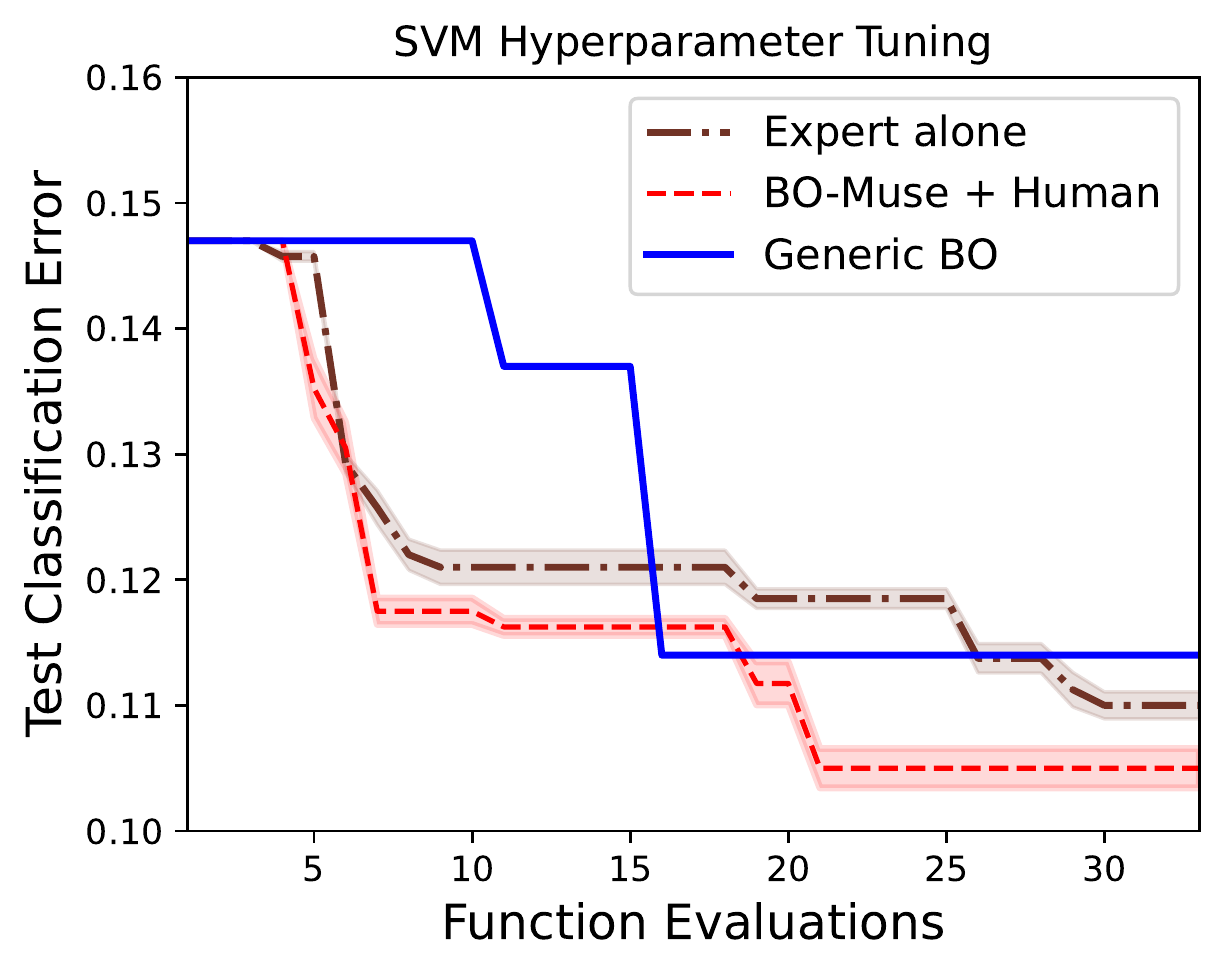}\label{fig:svm_real}}
\subfloat[]{\includegraphics[width=0.47\columnwidth]{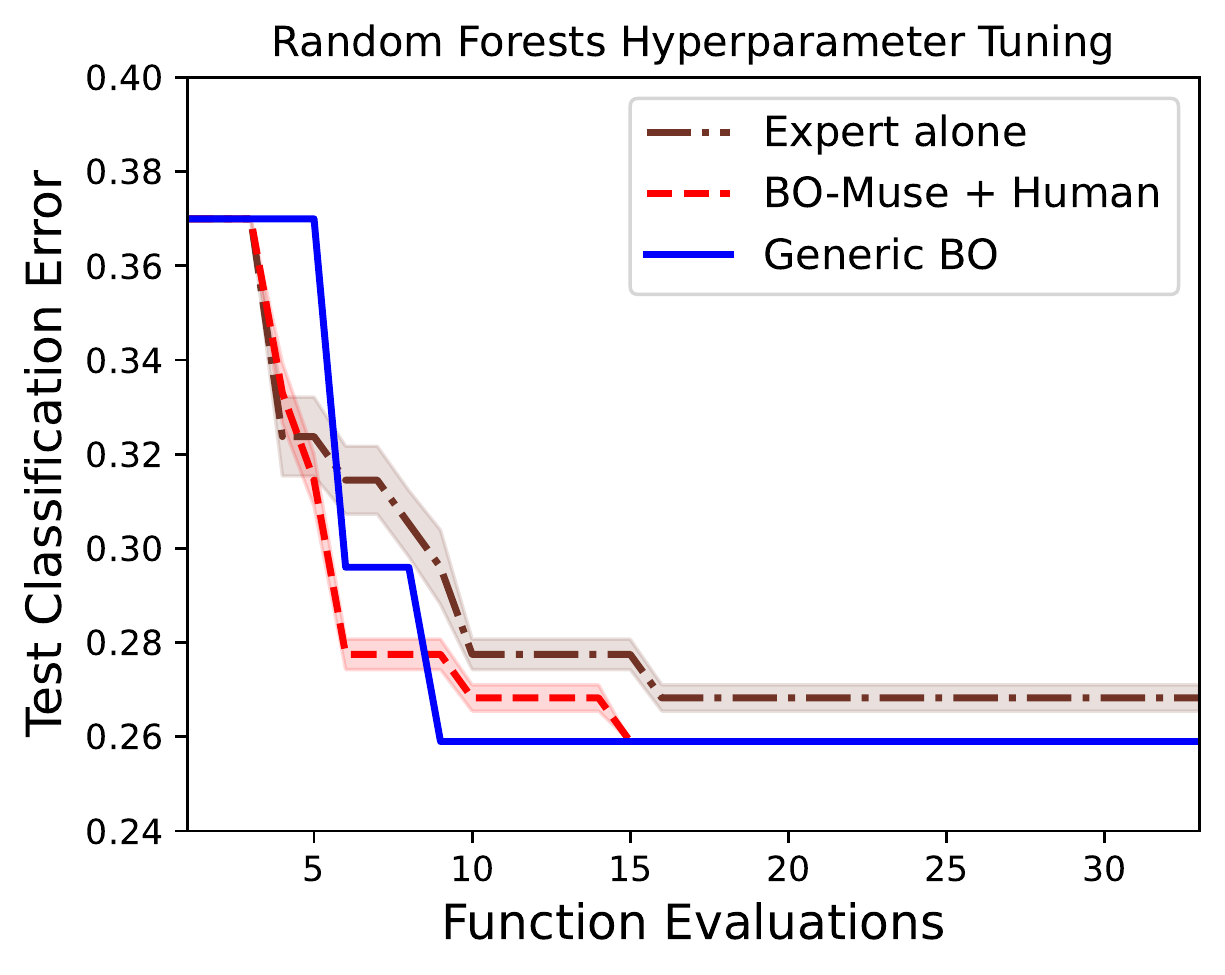}\label{fig:rf_real}} 
\caption{Simple regret vs iterations
for the hyper-parameter tuning of classifiers comparing Team 1 (BO-Muse
+ Human) (red) vs Team 2 (Human only) (black) vs AI alone (blue). We report the simple
regret mean (along with its standard deviation) for (a) Support Vector
Machines (SVM) and (b) Random Forests (RF) classifiers.} 
\label{fig:realworld_clf}
\end{centering}
\end{figure}

Figure \ref{fig:synthetic-1} shows the simple regret computed for Matyas-2D , Ackley-4D, Rastrigin-5D and Levy-6D functions
averaged over 10 randomly initialized runs.  As seen, BO-Muse consistently performs better than most baselines.  The poor performance of Simulated Human + PE is due to the over-exploration used by pure exploration strategy which maximizes only predictive uncertainty.  Additionally,  we have conducted an ablation study by varying the human exploitation-exploration trade-off ($\hat{\beta_{s}}$) to simulate a range of over-exploitative to over-explorative experts.  These ablation study results are provided in Appendix (see section \ref{subsec:ablation_beta}.) To further understand the behavior of BO-Muse framework with other human acquisition functions, we have also simulated an Expected Improvement (EI) acquisition function based human expert.  The results of this ablation study are provided in Appendix (see section \ref{subsec:ablation_ei}).

\subsection{Real-world Experiments}

We now present our experiments for complex
real-world optimization tasks. 

\subsubsection{Classification Tasks -- Support Vector Machines and Random Forests}

\paragraph{Experimental Set-up.}

In this experiment, our task is to choose hyper-parameters for Support Vector Machine
(SVM) and Random Forest (RF) classifiers operating on real-world \textit{Biodeg
}dataset from UCI repository \citep{Dua_2019}. We divide the dataset
into random 80/20 train/test splits. We set up two human expert teams.
Each member of Team 1 works in partnership with BO-Muse, whilst
members of Team 2 work individually without BO-Muse. We recruited
8 participants\footnote{Necessary ethics approval obtained.} consisting
of 4 postdocs and 4 postgraduate students. 2 postdocs and 2 students
are allocated to each team randomly so that each team has 4 participants,
with roughly similar expertise. Each participant is given the same
budget, 3 random initial designs + 30 further iterations. At the end
of each iteration, the test classification error for the suggested
hyper-parameter set is computed. We measure the overall performance
using simple regret, which will be the minimum test classification
error observed so far. The individual results from each team are averaged
to compare (Team 1) BO-Muse + Human vs (Team 2) Human alone (baseline). Additionally, we also report the performance of Generic BO (AI alone) method to demonstrate the efficacy of our approach. Further, we have set the same seed initialization and allocated the same evaluation budget for all the algorithms. 

\paragraph{Interfacing with Experts.}

The human experts perform two hyper-parameter tuning experiments to
minimize the test classification error of SVM and RF.  Each expert is
provided with a simple graphical interface that shows accumulated observations of classifier performance as
a function of hyper-parameters along with the best result thus far.  Experts suggest the next hyper-parameter set by clicking at a point of their choice inside the plot.  Same interface is provided to both teams.

\paragraph{Experiment 1 -- SVM Classification.}

In this experiment we have considered C-SVM classifiers with Radial
Basis Function (RBF) kernel. We have used LibSVM \citep{libsvm}
implementation of SVM with hyper-parameters kernel scale $\gamma$
and the cost parameter $C$. The SVM hyper-parameters \textit{i.e.,}
$\gamma$ and $C$ are tuned in the exponent space of $[-3,3]$.

\paragraph{Experiment 2 -- Random Forest Classification.}

In the classification tasks with random forests we tune the hyper-parameters \textit{maximum depth of the
decision tree }and the \textit{number of samples per split} in the
range $(0,100]$, and $(1,50]$, respectively. 

The results of our classification experiments are depicted in Figure
\ref{fig:svm_real} and Figure \ref{fig:rf_real}. In both experiments,
the BO-Muse + Human team outperforms the experts working on their
own. 

\subsubsection{Space shield design application}

Our third experiment applies BO-Muse to a real-world applied engineering
problem. We consider the design of a shield for protecting spacecraft
against the impact of space debris particles by partnering with a
world leading impact expert.  Due to the expensive nature of this experiment, it was not feasible to do perform multiple experiments or have access to many human experts.  Therefore, this experiment was primarily done to showcase an application rather than an evaluation.  We include the details of this experiment in Appendix \ref{subsec:SpacecraftShieldingDesign}. 

In the experiment (see Table \ref{tab:Results} in Appendix \ref{subsec:SpacecraftShieldingDesign})
we observe the human expert initially exploring solutions based on
the state-of-the-art for more typical debris impact problems (which
are normally simplified to spherical aluminium projectiles). The expert
is observed to rapidly exploit their initial 3 designs to identify
two feasible shielding solutions within the first four batch iterations
(ID \textcolor{blue}{4--11}, marked in \textcolor{blue}{blue}). The
expert performs further exploitation of these successful designs (Result=0)
over the next four batch iterations (ID \textcolor{brown}{12-19},
marked in \textcolor{brown}{brown}) in an attempt to reduce the weight,
but is unsuccessful. To this point the expert does not appear to have
been influenced at all by the BO suggestions. However, in the next
four batch iterations (ID \textcolor{green}{20-27}, marked in \textcolor{green}{green})
we can observe the expert taking inspiration from the previous BO-Muse
suggestions. One such exploitation results in a successful solution
(ID 27), further exploitation of which provides the best solution
identified by the experiment ID 29. This solution is highly \emph{unique}
for spacecraft debris shields, utilizing a polymer outer layer in
contact with a metallic backing to disrupt the debris particle. Such
a design is not reminiscent of any established flight hardware, see
e.g., \cite{MMODhandbook2009}. Thus, the BO-Muse is demonstrated
to have performed its role as hypothesized, inspiring the human expert
with novel designs that are subsequently subject to exploitation by
the human expert. The experimental set-up and results are discussed
in detail in Appendix \ref{subsec:SpacecraftShieldingDesign}.

\section{Conclusion}

We have presented a novel framework for human-AI teaming to accelerate expensive experimental design tasks.  Our algorithm lets the human expert take the lead
in the experimental process thus allowing them to fully use their
domain expertise, while the AI plays the role of a muse, injecting
novelty and searching for regions the human may have overlooked to
break the human out of over-exploitation induced by cognitive entrenchment.
We theoretically analyzed our algorithm to show that it converges sub-linearly and faster than
either the AI or human expert alone.  We demonstrated the utility of our algorithm using both synthetic and real world experimental design tasks.

\bibliography{main,universal}
\bibliographystyle{ArxivBib}

\newpage
\appendix
\onecolumn
\appendix

\section{Appendix}

\subsection{Proofs of Regret Bounds}


In this appendix we consider a generalized version of the framework.  For clarity 
(we favor brevity in the paper body, but clarity is essential here) we also require 
some additional notations.  As in the main paper, our goal is to solve:
\[
 \begin{array}{l}
  {\bf x}^\star = \mathop{\rm argmax}\limits_{{\bf x} \in \mathbb{X}} f^\star \left( {\bf x} \right)
 \end{array}
\]
where $f^\star$ is only measurable via an expensive and noisy process:
\[
 \begin{array}{l}
  y = f^\star \left( {\bf x} \right) + \nu
 \end{array}
\]
where $\nu$ is $\Sigma$-sub-Gaussian noise.

Let's assume that we have a series of experimental batches $1,2,\ldots$, where 
batch $s$ contains $\hat{p}$ human generated experiments and $\breve{p}$ AI 
generated experiments, giving a total of $p = \hat{p}+\breve{p}$ experiments.  
We use a hat with an index $i \in \hat{\mathbb{M}}$ to indicate a property 
relating to human $i$, and a breve with an index $j \in \breve{\mathbb{M}}$ to 
indicate a property relating to AI $j$, where:
\[
 \begin{array}{rll}
  {  \hat{\mathbb{M}}} &\!\!\!= \mathbb{N}_{  \hat{p}} & \mbox{(set of humans)} \\
  {\breve{\mathbb{M}}} &\!\!\!= \mathbb{N}_{\breve{p}} & \mbox{(set of AIs)} \\
 \end{array}
\]

We assume the batches are run sequentially, and wlog that the experiments within each 
batch are nominally ordered, so the set of all experiments may be indexed with $t$.  
We use a bar to differentiate between a property indexed by experiment number $t$, 
which is unbarred (e.g. $c_t$), and a property indexed by batch number $s$, which 
is barred (e.g. $\bar{d}_s$).  For 
simplicity we define:
\[
 \begin{array}{rll}
  s_t                       &\!\!\!\!= \left\lceil \frac{t}{p} \right\rceil    & \mbox{(batch $s$ in which experiment $t$ occurs)} \\
  {      {\underline{t}}}_s &\!\!\!\!= p \left( s   - 1 \right) + 1            & \mbox{(first experiment in batch $s$)} \\
  {      {\underline{t}}}   &\!\!\!\!= p \left( s_t - 1 \right) + 1            & \mbox{(first experiment in batch $s_t$)} \\
  {  \hat{\underline{t}}}_s &\!\!\!\!= p \left( s   - 1 \right) + 1            & \mbox{(first human experiment in the batch $s$)} \\
  {  \hat{\underline{t}}}   &\!\!\!\!= p \left( s_t - 1 \right) + 1            & \mbox{(first human experiment in the batch $s_t$)} \\
  {\breve{\underline{t}}}_s &\!\!\!\!= p \left( s   - 1 \right) + 1 + \hat{p}  & \mbox{(first AI experiment in the batch $s$)} \\
  {\breve{\underline{t}}}   &\!\!\!\!= p \left( s_t - 1 \right) + 1 + \hat{p}  & \mbox{(first AI experiment in the batch $s_t$)} \\
  \mathbb{T}_{     s}       &\!\!\!\!= \mathbb{N}_p + p \left( s-1 \right) + 1 & \mbox{(set of experiments in batch $s$)} \\
  \mathbb{T}_{\leq s}       &\!\!\!\!= \mathbb{N}_{sp} + 1                     & \mbox{(set of experiments up to and including batch $s$)} \\
 \end{array}
\]

We assume that humans and AIs maintain a GP model that is updated after each 
batch.  So, after batch $s$, the posterior means and variances are, 
respectively:
\[
 \begin{array}{rlrl}
  {  \hat{\bar{\mu}}}_{i,s} \left( {\bf x} \right) &\!\!\!= {  \hat{\bar{\mbox{\boldmath $\alpha$}}}}_{i,s}^{{\rm T}} {  \hat{\bar{\bf k}}}_{i,s} \left( {\bf x} \right), &
  {  \hat{\bar{\sigma}}}_{i,s}^{2} \left( {\bf x} \right) &\!\!\!= {  \hat{\bar{K}}}_{i,s} \left( {\bf x}, {\bf x} \right) - {  \hat{\bar{\bf k}}}_{i,s}^{\rm T} \left( {\bf x} \right) \left( {  \hat{\bar{\bf K}}}_{i,s} + {  \hat{\bar{\sigma}}}^2 {\bf I} \right)^{-1} {  \hat{\bar{\bf k}}}_{i,s} \left( {\bf x} \right) \\
  {\breve{\bar{\mu}}}_{j,s} \left( {\bf x} \right) &\!\!\!= {\breve{\bar{\mbox{\boldmath $\alpha$}}}}_{j,s}^{{\rm T}} {\breve{\bar{\bf k}}}_{j,s} \left( {\bf x} \right), &
  {\breve{\bar{\sigma}}}_{j,s}^{2} \left( {\bf x} \right) &\!\!\!= {\breve{\bar{K}}}_{j,s} \left( {\bf x}, {\bf x} \right) - {\breve{\bar{\bf k}}}_{j,s}^{\rm T} \left( {\bf x} \right) \left( {\breve{\bar{\bf K}}}_{j,s} + {\breve{\bar{\sigma}}}^2 {\bf I} \right)^{-1} {\breve{\bar{\bf k}}}_{j,s} \left( {\bf x} \right) \\
 \end{array}
\]
where:
\[
 \begin{array}{rlrl}
  {  \hat{\bar{\mbox{\boldmath $\alpha$}}}}_{i,s} &\!\!\!= \left( {  \hat{\bar{\bf K}}}_{i,s} + {  \hat{\bar{\sigma}}}^2 {\bf I} \right)^{-1} {\bar{\bf y}}_{s}, &
  {\breve{\bar{\mbox{\boldmath $\alpha$}}}}_{j,s} &\!\!\!= \left( {\breve{\bar{\bf K}}}_{j,s} + {\breve{\bar{\sigma}}}^2 {\bf I} \right)^{-1} {\bar{\bf y}}_{s} \\
  {  \hat{\bar{\bf K}}}_{i,s} &\!\!\!= \left[ \begin{array}{c} {  \hat{\bar{K}}}_{i,s} \left( {\bf x}_t, {\bf x}_{t'} \right) \end{array} \right]_{t,t' \in \mathbb{T}_{\leq s}}, \!\!\!\!&
  {\breve{\bar{\bf K}}}_{j,s} &\!\!\!= \left[ \begin{array}{c} {\breve{\bar{K}}}_{j,s} \left( {\bf x}_t, {\bf x}_{t'} \right) \end{array} \right]_{t,t' \in \mathbb{T}_{\leq s}} \\
  {  \hat{\bar{\bf k}}}_{i,s} \left( {\bf x} \right) &\!\!\!= \left[ \begin{array}{c} {  \hat{\bar{K}}}_{i,s} \left( {\bf x}_t, {\bf x} \right) \end{array} \right]_{t \in \mathbb{T}_{\leq s}}, &
  {\breve{\bar{\bf k}}}_{j,s} \left( {\bf x} \right) &\!\!\!= \left[ \begin{array}{c} {\breve{\bar{K}}}_{j,s} \left( {\bf x}_t, {\bf x} \right) \end{array} \right]_{t \in \mathbb{T}_{\leq s}} \\
  {\bar{\bf y}}_{s} &\!\!\!= \left[ \begin{array}{c} f^\star \left( {\bf x}_t \right) + \nu_t \end{array} \right]_{t \in \mathbb{T}_{\leq s}}, &
  {\bar{\bf f}}_{s} &\!\!\!= \left[ \begin{array}{c} f^\star \left( {\bf x}_t \right) \end{array} \right]_{t \in \mathbb{T}_{\leq s}} \\
 \end{array}
\]
We also occasionally use:
\[
 \begin{array}{rlrl}
  {\bf y}_{t} &\!\!\!= \left[ \begin{array}{c} f^\star \left( {\bf x}_{t'} \right) + \nu_{t'} \end{array} \right]_{t' \leq t}, &
  {\bf f}_{t} &\!\!\!= \left[ \begin{array}{c} f^\star \left( {\bf x}_{t'} \right) \end{array} \right]_{t' \leq t} \\
 \end{array}
\]

As this is a batch algorithm we are concerned with the instantaneous regret for 
the batches, not the individual experiments therein.  The instantaneous regret 
for batch $s$ is:
\[
 \begin{array}{ll}
  \bar{r}_s = \mathop{\rm min}\limits_{t \in \mathbb{T}_s} r_t, &
  r_t = f^\star \left( {\bf x}^\star \right) - f^\star \left( {\bf x}_t \right) \\
 \end{array}
\]
where $r_t$ is the instantaneous regret for experiment $t$.  The cumulative 
regret up to and including batch $S$ is:
\[
 \begin{array}{l}
  \bar{R}_S = \mathop{\sum}\limits_{s \in \mathbb{N}_S+1} \bar{r}_s
 \end{array}
\]

We do not assume $f^\star$ is drawn from any of the GP models for humans or 
AIs, so the problem mis-specified.  So, borrowing from \cite{Bog2}, we assume 
humans and AIs attempt to maximize the ``closest'' (best-in-class) function  
to $f^\star$ in the respective hypothesis spaces, using the shorthand 
${{\hat{\bar{\mathcal{H}}}}_{i,s}} = \mathcal{H}_{{{\hat{\bar{K}}}_{i,s}}}$, 
${{\breve{\bar{\mathcal{H}}}}_{j,s}} = \mathcal{H}_{{{\breve{\bar{K}}}_{j,s}}}$:
\[
 \begin{array}{rl}
  {  \hat{\bar{\mathbb{H}}}}_{i,s} &\!\!\!= \left\{ \left. f \in {{  \hat{\bar{\mathcal{H}}}}_{i,s}} \right| \left\| f \right\|_{{{  \hat{\bar{\mathcal{H}}}}_{i,s}}} \leq {  \hat{B}}_{i} \right\} \\
  {\breve{\bar{\mathbb{H}}}}_{j,s} &\!\!\!= \left\{ \left. f \in {{\breve{\bar{\mathcal{H}}}}_{j,s}} \right| \left\| f \right\|_{{{\breve{\bar{\mathcal{H}}}}_{j,s}}} \leq {\breve{B}}_{j} \right\} \\
 \end{array}
\]
where the closest to optimal approximations of $f^\star$ in the hypothesis spaces 
are:
\[
 \begin{array}{rlrl}
  {  \hat{\bar{f}}}_{i,s}^{\#} &\!\!\!= \mathop{{\rm argmin}}\limits_{f \in {  \hat{\bar{\mathbb{H}}}}_{i,s}} \left\| f - f^\star \right\|_\infty, &
  {\breve{\bar{f}}}_{j,s}^{\#} &\!\!\!= \mathop{{\rm argmin}}\limits_{f \in {\breve{\bar{\mathbb{H}}}}_{j,s}} \left\| f - f^\star \right\|_\infty \\
 \end{array}
\]
As is usual in practice, kernels may be updated when the GP models are updated, 
typically using max-log-likelihood for AIs or something more radical for the 
humans, which modifies the corresponding RKHSs.  The difference between the 
best-in-class approximations and $f^\star$ for batch $s$ are assumed bounded 
as:
\[
 \begin{array}{rlrl}
  \left\| {  \hat{\bar{f}}}_{i,s}^{\#} - f^\star \right\|_\infty \leq {  \hat{\bar{\epsilon}}}_{i,s}, &
  \left\| {\breve{\bar{f}}}_{j,s}^{\#} - f^\star \right\|_\infty \leq {\breve{\bar{\epsilon}}}_{j,s} \\
 \end{array}
\]
After each batch $s$, we have the (nominal) GP models built on (nominal) 
observations of ${\hat{\bar{f}}}_{i,s}^{\#}$ and ${\breve{\bar{f}}}_{j,s}^{\#}$, 
which have the same variance as the (real) models (the posterior variance is 
independent of ${\bf y}$) but different posterior means:
\[
 \begin{array}{rlrl}
  {  \hat{\bar{\mu}}}_{i,s}^{\#} \left( {\bf x} \right) &\!\!\!= {  \hat{\bar{\mbox{\boldmath $\alpha$}}}}_{i,s}^{\#{\rm T}} {  \hat{\bar{\bf k}}}_{i,s} \left( {\bf x} \right), &
  {\breve{\bar{\mu}}}_{j,s}^{\#} \left( {\bf x} \right) &\!\!\!= {\breve{\bar{\mbox{\boldmath $\alpha$}}}}_{j,s}^{\#{\rm T}} {\breve{\bar{\bf k}}}_{j,s} \left( {\bf x} \right) \\
 \end{array}
\]

We assume that test points are generated, either nominally (for humans) or 
directly (for AIs), EC-GP-UCB style \citep{Bog2}, from a sequence of interleaved 
$\beta_t$ sequences by the relevant human $i = t-\hat{\underline{t}}$ if $i \in 
\hat{\mathbb{M}}$ or AI $j = t-\breve{\underline{t}}$ if $j \in 
{\breve{\mathbb{M}}}$ in batch $s_t$, so:
\[
 \begin{array}{rl}
  \beta_t 
  &\!\!\!= \left\{ \begin{array}{ll}
   {  \hat{\beta}}_{i,\underline{t}} & \mbox{if } i = t-{  \hat{\underline{t}}} \in   \hat{\mathbb{M}} \\
   {\breve{\beta}}_{j,\underline{t}} & \mbox{if } j = t-{\breve{\underline{t}}} \in \breve{\mathbb{M}} \\
  \end{array} \right. \\
  \epsilon_t 
  &\!\!\!= \left\{ \begin{array}{ll}
   {  \hat{\epsilon}}_{i,\underline{t}} & \mbox{if } i = t-{  \hat{\underline{t}}} \in   \hat{\mathbb{M}} \\
   {\breve{\epsilon}}_{j,\underline{t}} & \mbox{if } j = t-{\breve{\underline{t}}} \in \breve{\mathbb{M}} \\
  \end{array} \right. \\
  \mu_{t-1} \left( {\bf x} \right) 
  &\!\!\!= \left\{ \begin{array}{ll}
   {  \hat{\bar{\mu}}}_{i,s_t-1} \left( {\bf x} \right) & \mbox{if } i = t-{  \hat{\underline{t}}} \in   \hat{\mathbb{M}} \\
   {\breve{\bar{\mu}}}_{j,s_t-1} \left( {\bf x} \right) & \mbox{if } j = t-{\breve{\underline{t}}} \in \breve{\mathbb{M}} \\
  \end{array} \right. \\
  \sigma_{t-1} \left( {\bf x} \right) 
  &\!\!\!= \left\{ \begin{array}{ll}
   {  \hat{\bar{\sigma}}}_{i,s_t-1} \left( {\bf x} \right) & \mbox{if } i = t-{  \hat{\underline{t}}} \in   \hat{\mathbb{M}} \\
   {\breve{\bar{\sigma}}}_{j,s_t-1} \left( {\bf x} \right) & \mbox{if } j = t-{\breve{\underline{t}}} \in \breve{\mathbb{M}} \\
  \end{array} \right. \\
 \end{array}
\]
using the acquisition function:
\[
 \begin{array}{l}
  \alpha_{t} \left( {\bf x} \right) = 
   \mu_{t-1} \left( {\bf x} \right) + \left( \sqrt{\beta_{t}} + \frac{\epsilon_t}{\sigma} \sqrt{\underline{t}} \right) \sigma_{t-1} \left( {\bf x} \right)
 \end{array}
\]
Generally we cannot control the human's exploitation/exploration trade-off 
sequence ${\hat{\beta}}_{i,t}$, but we assume humans are conservative, so the 
sequence may be assumed small.  We use the AI trade-off sequence 
${\breve{\beta}}_{j,t}$, which we do control, to compensate for the 
conservative tendencies of the humans involved.  We do however assume that the 
humans include at least some exploration in their decisions on the 
understanding that their knowledge is not, in fact, perfect, so 
${\hat{\beta}}_{i,t} \geq 0$ (note that a human who over-estimates their 
abilities may fail to meet this requirement, so care is required to avoid 
this).  For reasons which will become apparent, we assume that:
\[
 \begin{array}{rl}
  {{  \hat{\beta}}_{i,\underline{t}_{s}}} 
  &\!\!\!\in 
  \left( {  \hat{\bar{\zeta}}}_{i,s\downarrow }{{  \hat{\bar{\chi}}}_{i,s}}, {  \hat{\bar{\zeta}}}_{i,s\uparrow } {{  \hat{\bar{\chi}}}_{i,s}} \right) \\
  {{\breve{\beta}}_{j,\underline{t}_{s}}} 
  &\!\!\!\in
  \left( {\breve{\bar{\zeta}}}_{j,s\downarrow }{{\breve{\bar{\chi}}}_{j,s}}, {\breve{\bar{\zeta}}}_{j,s\uparrow } {{\breve{\bar{\chi}}}_{j,s}} \right) \\
 \end{array}
\]
where ${  \hat{\bar{\zeta}}}_{i,s\downarrow} \leq 1$, which we will see makes 
the humans over-exploitative, and ${\breve{\bar{\zeta}}}_{j,s\downarrow} \geq 
1$, which we will see makes the AIs over-explorative, and:
\[
 {{
 \begin{array}{l}
  {  \hat{\bar{\chi}}}_{i,s} = \left( \frac{\Sigma}{\sqrt{\sigma}} \sqrt{ 2 \ln \left( \frac{1}{\delta} \right) + 1 + {  \hat{\bar{\gamma}}}_{i,s}
  } + \left\| {  \hat{\bar{f}}}_{i,s}^{\#} \right\|_{{  \hat{\bar{\mathcal{H}}}}_{i,s}} \right)^2 \\
  {\breve{\bar{\chi}}}_{j,s} = \left( \frac{\Sigma}{\sqrt{\sigma}} \sqrt{ 2 \ln \left( \frac{1}{\delta} \right) + 1 + {\breve{\bar{\gamma}}}_{j,s}
  } + \left\| {\breve{\bar{f}}}_{j,s}^{\#} \right\|_{{\breve{\bar{\mathcal{H}}}}_{j,s}} \right)^2 \\
 \end{array}
 }}
\]
where ${\hat{\bar{\gamma}}}_{i,s}$ and ${\breve{\bar{\gamma}}}_{j,s}$ are, 
respectively, the max-information-gain terms for humans $i \in 
\hat{\mathbb{M}}$ and AIs $j \in \breve{\mathbb{M}}$, as will be described 
shortly.

\subsection{Notes on Maximum Information Gain}

We are assuming $p$ GPs, each of which will have a different information gain 
that is a function of its kernel.  All have the same dataset and get updated at 
the batch boundary.  Thus, after $S$ batches, if we consider human $i \in 
\hat{\mathbb{M}}$:
\[
 \begin{array}{rl}
  \hat{I}_{i} \left( {\bf y}_{Sp} : {\bf f}_{Sp} \right) 

  &\!\!\!= 
    H \left( {\bf y}_{Sp} \right) 
  - \frac{1}{2} \ln \left| 2 \pi e {  \hat{\bar{\sigma}}}^2 {\bf I}_{Sp} \right| \\

  &\!\!\!\!\!\!\!\!\!\!\!\!\!\!\!\!\!\!\!\!\!\!\!\!\!\!\!\!\!\!\!\!\!\!\!\!= 
    H \left( {\bf y}_{Sp-1} \right) 
  + H \left( \left. y_{Sp} \right| {\bf y}_{(S-1)p} \right) 
  - \frac{1}{2} \ln \left| 2 \pi e {  \hat{\bar{\sigma}}}^2 {\bf I}_{Sp} \right| \\

  &\!\!\!\!\!\!\!\!\!\!\!\!\!\!\!\!\!\!\!\!\!\!\!\!\!\!\!\!\!\!\!\!\!\!\!\!= 
    H \left( {\bf y}_{Sp-2} \right) 
  + H \left( \left. y_{Sp} \right| {\bf y}_{(S-1)p} \right) 
  + H \left( \left. y_{Sp-1} \right| {\bf y}_{(S-1)p} \right) 
  - \frac{1}{2} \ln \left| 2 \pi e {  \hat{\bar{\sigma}}}^2 {\bf I}_{Sp} \right| \\

  &\!\!\!\!\!\!\!\!\!\!\!\!\!\!\!\!\!\!\!\!\!\!\!\!\!\!\!\!\!\!\!\!\!\!\!\!= 
  \ldots \\

  &\!\!\!\!\!\!\!\!\!\!\!\!\!\!\!\!\!\!\!\!\!\!\!\!\!\!\!\!\!\!\!\!\!\!\!\!= 
    H \left( {\bf y}_{(S-1)p} \right) 
  + \mathop{\sum}\limits_{t \in \mathbb{T}_S} H \left( \left. y_{t} \right| {\bf y}_{\underline{t}-1} \right) 
  - \frac{1}{2} \ln \left| 2 \pi e {  \hat{\bar{\sigma}}}^2 {\bf I}_{Sp} \right| \\

  &\!\!\!\!\!\!\!\!\!\!\!\!\!\!\!\!\!\!\!\!\!\!\!\!\!\!\!\!\!\!\!\!\!\!\!\!= 
    H \left( {\bf y}_{(S-2)p} \right) 
  + \mathop{\sum}\limits_{t \in \mathbb{T}_S} H \left( \left. y_{t} \right| {\bf y}_{\underline{t}-1} \right) 
  + \mathop{\sum}\limits_{t \in \mathbb{T}_{S-1}} H \left( \left. y_{t} \right| {\bf y}_{\underline{t}-1} \right) 
  - \frac{1}{2} \ln \left| 2 \pi e {  \hat{\bar{\sigma}}}^2 {\bf I}_{Sp} \right| \\

  &\!\!\!\!\!\!\!\!\!\!\!\!\!\!\!\!\!\!\!\!\!\!\!\!\!\!\!\!\!\!\!\!\!\!\!\!= 
  \ldots \\

  &\!\!\!\!\!\!\!\!\!\!\!\!\!\!\!\!\!\!\!\!\!\!\!\!\!\!\!\!\!\!\!\!\!\!\!\!= 
    \mathop{\sum}\limits_{t \in \mathbb{T}_{\leq S}} H \left( \left. y_{t} \right| {\bf y}_{\underline{t}-1} \right) 
  - \frac{1}{2} \ln \left| 2 \pi e {  \hat{\bar{\sigma}}}^2 {\bf I}_{Sp} \right| \\

  &\!\!\!\!\!\!\!\!\!\!\!\!\!\!\!\!\!\!\!\!\!\!\!\!\!\!\!\!\!\!\!\!\!\!\!\!= 
  \frac{1}{2} \mathop{\sum}\limits_{t \in \mathbb{T}_{\leq S}} \ln \left( 2 \pi e \left( {  \hat{\bar{\sigma}}}^2 + {  \hat{\bar{\sigma}}}^{-2} {\hat{\bar{\sigma}}}^2_{i,\underline{t}-1} \left( {\bf x}_t \right) \right) \right) 
  - \frac{1}{2} \ln \left| 2 \pi e {  \hat{\bar{\sigma}}}^2 {\bf I}_{T} \right| \\

  &\!\!\!\!\!\!\!\!\!\!\!\!\!\!\!\!\!\!\!\!\!\!\!\!\!\!\!\!\!\!\!\!\!\!\!\!= 
  \frac{1}{2} \mathop{\sum}\limits_{t \in \mathbb{T}_{\leq S}} \ln \left( 1 + {  \hat{\bar{\sigma}}}^{-2} {\hat{\bar{\sigma}}}^2_{i,\underline{t}-1} \left( {\bf x}_t \right) \right) \\
 \end{array}
\]
where we have used that ${\bf x}_1, \ldots, {\bf x}_{sp}$ are deterministic 
conditioned on ${\bf y}_{(s-1)p}$, and that the variances do not depend on 
${\bf y}$.  The derivation for AI $j$ is essentially identical.  In summary, 
therefore, the information gain is:
\[
 {{
 \begin{array}{l}
  {  \hat{I}}_{i} \left( {\bf y}_{Sp} : {\bf f}_{Sp} \right) = \frac{1}{2} \mathop{\sum}\limits_{t \in \mathbb{T}_{\leq S}} \ln \left( 1 + {  \hat{\bar{\sigma}}}^{-2} {  \hat{\bar{\sigma}}}^2_{i,\underline{t}-1} \left( {\bf x}_t \right) \right) \leq {  \hat{\bar{\gamma}}}_{i,S} \\
  {\breve{I}}_{j} \left( {\bf y}_{Sp} : {\bf f}_{Sp} \right) = \frac{1}{2} \mathop{\sum}\limits_{t \in \mathbb{T}_{\leq S}} \ln \left( 1 + {\breve{\bar{\sigma}}}^{-2} {\breve{\bar{\sigma}}}^2_{j,\underline{t}-1} \left( {\bf x}_t \right) \right) \leq {\breve{\bar{\gamma}}}_{j,S} \\
 \end{array}
 }}
\]
where ${\hat{\bar{\gamma}}}_{i,S}$ and ${\breve{\bar{\gamma}}}_{j,S}$ are, 
respectively, the maximum information gains for human $i$ and AI $j$ over $S$ 
batches.

\subsection{Notes on the Human Model}

We posit that every human $i \in \hat{\mathbb{M}}$ has an {\em evolving} model 
of the system:
\[
 \begin{array}{l}
  {\hat{\bar{f}}}_{i,s} \left( {\bf x} \right) = {\hat{\bar{g}}}_{i,s} \left( {\hat{\bar{\bf p}}}_{i,s} \left( {\bf x} \right) \right)
 \end{array}
\]
where ${\hat{\bar{g}}}_{i,s}$ is in some sense ``simple'' and ${\hat{\bar{\bf 
p}}}_{i,s}: \mathbb{R}^n \to \mathbb{R}^{{\hat{\bar{m}}}_{i,s}}$.  This fits 
into the above scheme if we let the form of ${\hat{ \bar{g}}}_{i,s}$ dictate 
the kernel ${\hat{\bar{K}}}_{i,s}$.  For example if we know that 
${\hat{\bar{g}}}_{i,s}$ is linear - \textit{i.e.,} the human is known to be using some 
heuristic model of the form:
\[
 \begin{array}{l}
  {\hat{\bar{f}}}_{i,s} \left( {\bf x} \right) = {\hat{\bar{\bf w}}}_{i,s}^{\rm T} {\hat{\bar{\bf p}}}_{i,s} \left( {\bf x} \right)
 \end{array}
\]
then we can use a GP with a linear-derived kernel:
\[
 \begin{array}{l}
  {\hat{\bar{K}}}_{i,s} \left( {\bf x}, {\bf x}' \right) = {\hat{\bar{\bf p}}}_{i,s}^{\rm T} \left( {\bf x} \right) {\hat{\bar{\bf p}}}_{i,s} \left( {\bf x}' \right)
 \end{array}
\]
Similarly if the human is using a model ${\hat{\bar{g}}}_{i,s}$ that can be 
captured by a $d^{\rm th}$-order polynomial model:
\[
 \begin{array}{l}
  {\hat{\bar{f}}}_{i,s} \left( {\bf x} \right) = {\hat{\bar{\bf w}}}_{i,s}^{\rm T} \left[ \sqrt{\left( d \atop q \right)} {\hat{\bar{\bf p}}}_{i,s}^{\otimes q} \left( {\bf x} \right) \right]_{q \in \mathbb{N}_{d+1}}
 \end{array}
\]
then we can use a GP with a polynomial-derived kernel:
\[
 \begin{array}{l}
  {\hat{\bar{K}}}_{i,s} \left( {\bf x}, {\bf x}' \right) = \left( 1 + {\hat{\bar{\bf p}}}_{i,s}^{\rm T} \left( {\bf x} \right) {\hat{\bar{\bf p}}}_{i,s} \left( {\bf x}' \right) \right)^d
 \end{array}
\]
Alternatively, if ${\hat{\bar{g}}}_{i,s}$ is more vague (\textit{i.e.,} the researcher 
knows that the factors ${\hat{\bar{\bf p}}}_{i,s}$ are important but not the 
exact form of the relationship) then we might use a GP assuming a 
distance-based model:
\[
 \begin{array}{l}
  {\hat{\bar{K}}}_{i,s} \left( {\bf x}, {\bf x}' \right) = \exp \left( -\frac{1}{2l} \left\| {\hat{\bar{\bf p}}}_{i,s} \left( {\bf x} \right) - {\hat{\bar{\bf p}}}_{i,s} \left( {\bf x}' \right) \right\|_2^2 \right)
 \end{array}
\]
or some similarly generic model that captures the worst-case behavior of the 
human, along, hopefully, with some insight into the thought processes used by 
them.

With regard to maximum information gain, because the human models are evolving, 
${\hat{\bar{m}}}_{i,s}$ will change with $s$, so it is convenient to define 
define ${\hat{\bar{m}}}_{i,s\uparrow} = {\rm max}_{t \in \mathbb{T}_s} 
{\hat{\bar{m}}}_{i,s}$ to capture the worst-case feature-space dimensionality 
over $S$ batches.  We can then bound the maximum information gain for the human 
as the worst-case of these models over all $S$ batches.  So for example, 
depending on the specifics of ${\hat{\bar{g}}}_{i,s}$, we have 
\citep{Sri1,Sce1}:
\[
 \begin{array}{rll}
  \mbox{Linear:} & \hat{\bar{\gamma}}_{i,s} = \mathcal{O} \left( \ln \left( S p \right) \right) & \\
  \mbox{Polynomial:} & \hat{\bar{\gamma}}_{i,s} = \mathcal{O} \left( \ln \left( S p \right) \right) & \\ 
  \mbox{Squared-Exponential:} & \hat{\bar{\gamma}}_{i,s} = \mathcal{O} \left( \ln^{{\hat{\bar{m}}}_{i,s\uparrow}+1} \left( S p \right) \right) & \\
 \end{array}
\]

In general we assume that the asymptotic behavior of the human maximum 
information gain converges more quickly than that of the machine models.  This 
makes intuitive sense of the human is applying a linear or polynomial heuristic 
model ${\hat{\bar{g}}}_{i,s}$, which is captured by a linear or polynomial 
kernel, while the machines use more general GP models with SE or Matern 
type kernels (as is common practice).  Thus in this case the human has a better 
behaved maximum information gain at the cost of a potentially non-zero gap 
${\hat{\bar{\epsilon}}}_{i,s}$ (presumably trending to $0$ as the human gains 
improved insight into the problem and evolves their model to better match the 
problem), while the machine has a worse behaved maximum information gain but 
zero gap (assuming a universal kernel like an SE kernel).

\subsection{Mathematical Preliminaries}

We use $p$-norms extensively, where:
\[
\begin{array}{rll}
  \left\| {\bf x} \right\|_p &\!\!\!\!= \left( \sum_i \left| x_i \right|^p \right)^{\frac{1}{p}} & \mbox{where } {\bf x} \in \mathbb{R}^n \\
  \left\| f \right\|_p &\!\!\!\!= \left( \int_{\bf x} \left| f \left( {\bf x} \right) \right|^p d{\bf x} \right)^{\frac{1}{p}} & \mbox{where } f : \mathbb{R}^n \to \mathbb{R} \\
 \end{array}
\]
with the extensions:
\[
 \begin{array}{rl}
  \!\!\left\| {\bf x} \right\|_{ \infty} &\!\!\!\!\!= \max_i \left| x_i \right| \\
  \!\!\left\| {\bf x} \right\|_{-\infty} &\!\!\!\!\!= \min_i \left| x_i \right| \\
 \end{array}
\]
\[
 \begin{array}{rl}
  \left\| f \right\|_{ \infty} 
                               &\!\!\!\!\!= \inf \left\{ C \geq 0 : \left| f \left( {\bf x} \right) \right| \leq C \mbox{ for almost all } {\bf x} \in \mathbb{R}^n \right\} \\
  \left\| f \right\|_{-\infty} 
                               &\!\!\!\!\!= \sup \left\{ C \geq 0 : \left| f \left( {\bf x} \right) \right| \geq C \mbox{ for almost all } {\bf x} \in \mathbb{R}^n \right\} \\
 \end{array}
\]
This is only a norm for $p \in [1,\infty]$ (though we may occasionally refer to 
it as such in a loose sense), but is well defined for $p \in [-\infty,0) \cup 
(0,\infty]$.  It is not difficult to see that:
\[
 \begin{array}{lll}
  \left\| {\bf x} \right\|_p \geq \left\| {\bf x} \right\|_{p'}, & 
  \forall p,p' \in (0, \infty], p \leq p' \\
  \left\| {\bf x} \right\|_p \geq \left\| {\bf x} \right\|_{p'}, & 
  \forall p,p' \in [-\infty,0), p \leq p' \\
  \left\| {\bf x} \right\|_p \geq \left\| {\bf x} \right\|_{p'}, & 
  \forall p \in (0,\infty], p' \in [-\infty,0) \\
 \end{array}
\]
(the final inequality follows from the first two, noting that $\min_i |x_i| < 
\max_i |x_i|$, and similarly for functions); and furthermore:
\[
 \begin{array}{l}
  \left\| {\bf x} \right\|_p \leq n^{\frac{1}{p}-\frac{1}{p'}} \left\| {\bf x} \right\|_{p'} \;\;\; \forall 1 \leq p \leq p' \leq \infty
 \end{array}
\]
when ${\bf x} \in \mathbb{R}^n$.  
We also use the following inequalities ($\odot$ is the Hadamard product):
\[
 %
\]
where $\frac{1}{p} + \frac{1}{p'} + \ldots + \frac{1}{p''''} = \frac{1}{r}$ and 
$r,p,p',\ldots,p'''' \in (0,\infty]$, with the convention $\frac{1}{\infty} = 
0$; and $q \in (0,\infty)$.

We also use generalized (power) mean \citep{Pec1,Pyu1}, defined as:
\[
 %
\]
where $\theta \in [-\infty, \infty]$.  For example $\theta=-1,0,1$ correspond, 
respectively, to the harmonic, geometric and arithmetic means.  Note that, for 
all $\theta, \theta' \in [-\infty,\infty]$, $\theta \leq \theta'$:
\[
 \begin{array}{l}
  M_\theta \left( \left\{ a_1, a_2, \ldots a_n \right\} \right) \leq M_{\theta'} \left( \left\{ a_1, a_2, \ldots a_n \right\} \right)
 \end{array}
\]
With a minor abuse of notation, we define:
\[
 \begin{array}{l}
  M_\theta \left( {\bf a} \right) = M_\theta \left( \left\{ a_1, a_2, \ldots a_n \right\} \right) 
 \end{array}
\]
for ${\bf a} \in \mathbb{R}^n$, so that, for ${\bf a} \in \mathbb{R}_+^n$, we 
have the connection to the $p$-norms:
\[
 %
\]
The following result is central to our proof:\footnote{This result may be well 
known, but we have been unable to find it in the literature.}
\begin{lem_neginfnorm}
 Let ${\bf a}, {\bf b} \in \mathbb{R}_+^n$, $z \in [-\infty, \infty]$, $q,q' 
 \in [1, \infty]$, $\frac{1}{q} + \frac{1}{q'} = 1$, $\frac{1}{\infty} = 0$.  
 Then we have the following bounds on $M_{-\infty} ({\bf a} \odot {\bf b})$:
 \[
  \begin{array}{rl}
  M_{-\infty} \left( {\bf a} \odot {\bf b} \right) 
  &\!\!\!\leq M_{-z} \left( {\bf a} \right) M_{z} \left( {\bf b} \right) \\ 
  M_{-\infty} \left( {\bf a} \odot {\bf b} \right) 
  &\!\!\!\leq M_{\left| zq \right|} \left( {\bf a} \right) M_{\left| zq' \right|} \left( {\bf b} \right) \\ 
  M_{-\infty} \left( {\bf a} \odot {\bf b} \right) 
  &\!\!\!\leq M_{-\left| z \right|} \left( {\bf a} \right) M_{\frac{q}{q-1} \left| z \right|} \left( {\bf b} \right) \\
  \end{array}
 \]
 \label{lem:neginfnorm}
\end{lem_neginfnorm}
\begin{proof}
Let us suppose that, unlike in the theorem, $z \in ( -\infty, 0) \cup (0, 
\infty )$.  Then, using the generalized mean inequality:
\[
 \begin{array}{rl}
  M_{-\infty} \left( {\bf a} \odot {\bf b} \right) 
  &\!\!\!\leq M_z \left( {\bf a} \odot {\bf b} \right) 
  = n^{-\frac{1}{z}} \left\| {\bf a} \odot {\bf b} \right\|_z 
  = n^{-\frac{1}{z}} \left\| {\bf a}^{\odot z} \odot {\bf b}^{\odot z} \right\|_1^{\frac{1}{z}} 
 \end{array}
\]
If $z > 0$ then, using H{\"o}lder's inequality:
\[
 \begin{array}{rl}
  M_{-\infty} \left( {\bf a} \odot {\bf b} \right) 
  &\!\!\!= n^{-\frac{1}{\left| z \right|}} \left\| {\bf a}^{\odot \left| z \right|} \odot {\bf b}^{\odot \left| z \right|} \right\|_1^{\frac{1}{\left| z \right|}} \\
  &\!\!\!\leq n^{-\frac{1}{\left| z \right|}} \left\| {\bf a}^{\odot \left| z \right|} \right\|_q^{\frac{1}{\left| z \right|}} \left\| {\bf b}^{\odot \left| z \right|} \right\|_{q'}^{\frac{1}{\left| z \right|}} \\
  &\!\!\!= n^{-\frac{1}{\left| z \right|}} \left\| {\bf a}^{\odot \left| zq \right|} \right\|_1^{\frac{1}{\left| zq \right|}} \left\| {\bf b}^{\odot \left| zq' \right|} \right\|_{1}^{\frac{1}{\left| zq' \right|}} \\
  &\!\!\!= n^{-\frac{1}{\left| z \right|}} n^{\frac{1}{\left| zq \right|}} n^{\frac{1}{\left| zq' \right|}} \left\| \frac{1}{n} {\bf a}^{\odot \left| zq \right|} \right\|_1^{\frac{1}{\left| zq \right|}} \left\| \frac{1}{n} {\bf b}^{\odot \left| zq' \right|} \right\|_{1}^{\frac{1}{\left| zq' \right|}} \\
  &\!\!\!= n^{-\frac{1}{\left| zq' \right|}} \left\| {\bf a} \right\|_{\left| zq \right|} n^{-\frac{1}{\left| zq' \right|}} \left\| {\bf b} \right\|_{\left| zq' \right|} \\
  &\!\!\!= M_{\left| zq \right|} \left( {\bf a} \right) M_{\left| zq' \right|} \left( {\bf b} \right) \\
 \end{array}
\]
If $z < 0$ then, using the reverse H{\"o}lder inequality (noting the negative 
exponent):
\[
 \begin{array}{rl}
  M_{-\infty} \left( {\bf a} \odot {\bf b} \right) 
  &\!\!\!= n^{\frac{1}{\left| z \right|}} \left\| {\bf a}^{\odot -\left| z \right|} \odot {\bf b}^{\odot -\left| z \right|} \right\|_1^{-\frac{1}{\left| z \right|}} \\
  &\!\!\!\leq n^{\frac{1}{\left| z \right|}} \left\| {\bf a}^{\odot -\left| z \right|} \right\|_{\frac{1}{q}}^{-\frac{1}{\left| z \right|}} \left\| {\bf b}^{\odot -\left| z \right|} \right\|_{-\frac{1}{q-1}}^{- \frac{1}{\left| z \right|}} \\
  &\!\!\!= n^{\frac{1}{\left| z \right|}} \left\| {\bf a}^{\odot -\frac{\left| z \right|}{q}} \right\|_1^{-\frac{q}{\left| z \right|}} \left\| {\bf b}^{\odot \frac{\left| z \right|}{q-1}} \right\|_1^{\frac{q-1}{\left| z \right|}} \\
  &\!\!\!= n^{\frac{1}{\left| z \right|}} n^{-\frac{q}{\left| z \right|}} n^{\frac{q-1}{\left| z \right|}} \left\| \frac{1}{n} {\bf a}^{\odot -\frac{\left| z \right|}{q}} \right\|_1^{-\frac{q}{\left| z \right|}} \left\| \frac{1}{n} {\bf b}^{\odot \frac{\left| z \right|}{q-1}} \right\|_1^{\frac{q-1}{\left| z \right|}} \\
  &\!\!\!= n^{\frac{q}{\left| z \right|}} \left\| {\bf a} \right\|_{-\frac{\left| z \right|}{q}} n^{-\frac{q-1}{\left| z \right|}} \left\| {\bf b} \right\|_{\frac{\left| z \right|}{q-1}} \\
  &\!\!\!= M_{-\frac{\left| z \right|}{q}} \left( {\bf a} \right) M_{\frac{\left| z \right|}{q-1}} \left( {\bf b} \right) \\
 \end{array}
\]
It is instructive to let $r = |z|/q$.  Then the most recent bound becomes:
\[
 \begin{array}{rl}
  M_{-\infty} \left( {\bf a} \odot {\bf b} \right) 
  &\!\!\!\leq M_{-r} \left( {\bf a} \right) M_{\frac{q}{q-1} r} \left( {\bf b} \right) \\
 \end{array}
\]
which in the limit $q \to \infty$ simplifies to:
\[
 \begin{array}{rl}
  M_{-\infty} \left( {\bf a} \odot {\bf b} \right) 
  &\!\!\!\leq M_{-r} \left( {\bf a} \right) M_{r} \left( {\bf b} \right) \\
 \end{array}
\]
Finally, using the definitions:
\[
 \begin{array}{rl}
  M_{-\infty} \left( {\bf a} \odot {\bf b} \right) 
  &\!\!\!\leq M_0 \left( {\bf a} \odot {\bf b} \right) 
  = M_0 \left( {\bf a} \right) M_0 \left( {\bf b} \right) \\
 \end{array}
\]
completing the proof.
\end{proof}

\subsection{Bogunovic's Lemma}

We have the following from \cite{Bog2}:
\begin{equation}
 \begin{array}{l}
  \left| \mu_{t-1}^{\#} \left( {\bf x} \right) - \mu_{t-1} \left( {\bf x} \right) \right| \leq \frac{\epsilon_t}{\sigma} \sqrt{\underline{t}} \sigma_{t-1} \left( {\bf x} \right) \\
 \end{array}
 \label{eq:boglemma}
\end{equation}
This follows from \cite[Lemma 2]{Bog2} using that all models satisfy this 
result, and that the model is posterior on the observations up to and including 
the previous batch.  It follows from this and the definitions that, for all 
${\bf x} \in \mathbb{X}$:
\begin{equation}
 \begin{array}{rl}
  \left| {  \hat{\bar{\mu}}}_{i,s-1} \left( {\bf x} \right) - f^\star \left( {\bf x} \right) \right| 
  &\!\!\!\leq 
  \left| {  \hat{\bar{\mu}}}_{i,s-1}^{\#} \left( {\bf x} \right) - {  \hat{\bar{f}}}_{i,s-1}^{\#} \left( {\bf x} \right) \right| + 
  \frac{{  \hat{\epsilon}}_{i,\underline{t}_s}}{\sigma} \sqrt{\underline{t}_s} {{  \hat{\bar{\sigma}}}_{i,s-1} \left( {\bf x} \right)} + {  \hat{\epsilon}}_{i,\underline{t}_s} \\

  \left| {\breve{\bar{\mu}}}_{j,s-1} \left( {\bf x} \right) - f^\star \left( {\bf x} \right) \right| 
  &\!\!\!\leq 
  \left| {\breve{\bar{\mu}}}_{j,s-1}^{\#} \left( {\bf x} \right) - {\breve{\bar{f}}}_{j,s-1}^{\#} \left( {\bf x} \right) \right| + 
  \frac{{\breve{\epsilon}}_{j,\underline{t}_s}}{\sigma} \sqrt{\underline{t}_s} {{\breve{\bar{\sigma}}}_{j,s-1} \left( {\bf x} \right)} + {\breve{\epsilon}}_{j,\underline{t}_s} \\
 \end{array}
 \label{eq:normboundyc}
\end{equation}

\subsection{The Regret Bound}

We begin with the following uncertainty bound, which is largely based on the 
bound \cite[Theorem 2]{Cho7} and analogous to \cite[Lemma 5.1]{Sri1} and 
\cite[Lemma 1]{Bog2}:
\begin{lem_boglem1}
Let $\delta \in (0,1)$.  Assume noise variables are $\Sigma$-sub-Gaussian.  Let:
\begin{equation}
 {{
 \begin{array}{l}
  {  \hat{\bar{\chi}}}_{i,s} = \left( \frac{\Sigma}{\sqrt{{  \hat{\bar{\sigma}}}}} \sqrt{ 2 \ln \left( \frac{1}{\delta} \right) + 1 + {  \hat{\bar{\gamma}}}_{i,s}
  } + \left\| {  \hat{\bar{f}}}_{i,s}^{\#} \right\|_{{  \hat{\bar{\mathcal{H}}}}_{i,s}} \right)^2 \\
  {\breve{\bar{\chi}}}_{j,s} = \left( \frac{\Sigma}{\sqrt{{\breve{\bar{\sigma}}}}} \sqrt{ 2 \ln \left( \frac{1}{\delta} \right) + 1 + {\breve{\bar{\gamma}}}_{j,s}
  } + \left\| {\breve{\bar{f}}}_{j,s}^{\#} \right\|_{{\breve{\bar{\mathcal{H}}}}_{j,s}} \right)^2 \\
 \end{array}
 }}
 \label{eq:definechi}
\end{equation}
Then, for all $i \in \hat{\mathbb{M}}$ and $j \in \breve{\mathbb{M}}$:
\[
 \begin{array}{l}
  \Pr \Bigg\{ \forall s, \forall {\bf x} \in \mathbb{X}, 
  \left| {  \hat{\bar{\mu}}}_{i,s-1} \left( {\bf x} \right) - f^\star \left( {\bf x} \right) \right|
  \leq \ldots \\
  \;\;\;\;\;\;\;\;\;\;\;\;\;\;\;\;\;\;\;\;\;\;\;\;\;\;\;\;\;\;\;\;\;\;\;\;\;\;\;\;\left( \sqrt{{  \hat{\bar{\chi}}}_{i,s}} + \frac{{  \hat{\epsilon}}_{i,\underline{t}_s}}{{  \hat{\bar{\sigma}}}} \sqrt{\underline{t}_s} \right)
  {{  \hat{\bar{\sigma}}}_{i,s-1} \left( {\bf x} \right)} 
  + {  \hat{\epsilon}}_{i,\underline{t}_s}
  \Bigg\} 
  \geq 1-\delta \\
  \Pr \Bigg\{ \forall s, \forall {\bf x} \in \mathbb{X}, 
  \left| {\breve{\bar{\mu}}}_{j,s-1} \left( {\bf x} \right) - f^\star \left( {\bf x} \right) \right| 
  \leq \ldots \\
  \;\;\;\;\;\;\;\;\;\;\;\;\;\;\;\;\;\;\;\;\;\;\;\;\;\;\;\;\;\;\;\;\;\;\;\;\;\;\;\;\left( \sqrt{{\breve{\bar{\chi}}}_{j,s}} + \frac{{\breve{\epsilon}}_{j,\underline{t}_s}}{{\breve{\bar{\sigma}}}} \sqrt{\underline{t}_s} \right)
  {{\breve{\bar{\sigma}}}_{j,s-1} \left( {\bf x} \right)}  
  + {\breve{\epsilon}}_{j,\underline{t}_s}
  \Bigg\}
  \geq 1-\delta \\
 \end{array}
\]
\label{lem:boglem1}
\end{lem_boglem1}
\begin{proof}
We start with \cite[Theorem 2]{Cho7}.  This states that, in our setting, with 
probability $\geq 1-\delta$, simultaneously for all $s \geq 1$ and ${\bf x} \in 
D$:
\[
 \begin{array}{l}
  \left| {  \hat{\bar{\mu}}}_{i,s-1}^{\#} \left( {\bf x} \right) - {  \hat{\bar{f}}}_{i,s-1}^{\#} \left( {\bf x} \right) \right| 
  \leq 
  {  \hat{\bar{\sigma}}}_{i,s-1}^{-1} \left( {\bf x} \right) \sqrt{{  \hat{\bar{\chi}}}_{i,s}} \\
  \left| {\breve{\bar{\mu}}}_{j,s-1}^{\#} \left( {\bf x} \right) - {\breve{\bar{f}}}_{j,s-1}^{\#} \left( {\bf x} \right) \right| 
  \leq 
  {\breve{\bar{\sigma}}}_{j,s-1}^{-1} \left( {\bf x} \right) \sqrt{{\breve{\bar{\chi}}}_{j,s}} \\
 \end{array}
\]
for ${  \hat{\bar{\chi}}}_{i,s}$, ${\breve{\bar{\chi}}}_{j,s}$ as specified by 
(\ref{eq:definechi}).  Next, recall (\ref{eq:normboundyc}):
\[
 {\!\!\!\!\!\!\!\!{
 \begin{array}{rl}
  \left| {  \hat{\bar{\mu}}}_{i,s-1} \left( {\bf x} \right) - f^\star \left( {\bf x} \right) \right| 
  &\!\!\!\!\!\leq 
  \left| {  \hat{\bar{\mu}}}_{i,s-1}^{\#} \left( {\bf x} \right) - {  \hat{\bar{f}}}_{i,s-1}^{\#} \left( {\bf x} \right) \right| 
  + \!\frac{{  \hat{\epsilon}}_{i,s}}{{  \hat{\bar{\sigma}}}} \sqrt{\underline{t}_s} {{  \hat{\bar{\sigma}}}_{i,s-1} \left( {\bf x} \right)} + {  \hat{\epsilon}}_{i,s} \\
  \left| {\breve{\bar{\mu}}}_{j,s-1} \left( {\bf x} \right) - f^\star \left( {\bf x} \right) \right| 
  &\!\!\!\!\!\leq 
  \left| {\breve{\bar{\mu}}}_{j,s-1}^{\#} \left( {\bf x} \right) - {\breve{\bar{f}}}_{j,s-1}^{\#} \left( {\bf x} \right) \right| 
  + \!\frac{{\breve{\epsilon}}_{j,s}}{{\breve{\bar{\sigma}}}} \sqrt{\underline{t}_s} {{\breve{\bar{\sigma}}}_{j,s-1} \left( {\bf x} \right)} + {\breve{\epsilon}}_{j,s} \\
 \end{array}
 }\!\!\!\!\!\!\!\!\!\!\!}
\]
and the result follows.
\end{proof}

{\bf Remark:} Alternatively, following \cite[Theorem 1]{Fie1} we can 
use:
\[
 {{
 \begin{array}{l}
  {  \hat{\bar{\chi}}}_{i,s} \geq \left( \frac{\Sigma}{\sqrt{{  \hat{\bar{\sigma}}}}} \sqrt{ 2 \ln \left( \frac{1}{\delta} \right) + \ln \left( \det \left( {  \hat{\bar{\bf K}}}_{i,s-1} + {  \hat{\bar{\sigma}}}^2 {\bf I} \right) \right)
  } + \left\| {  \hat{\bar{f}}}_{i,s}^{\#} \right\|_{{  \hat{\bar{\mathcal{H}}}}_{i,s}} \right)^2 \\
  {\breve{\bar{\chi}}}_{j,s} \geq \left( \frac{\Sigma}{\sqrt{{\breve{\bar{\sigma}}}}} \sqrt{ 2 \ln \left( \frac{1}{\delta} \right) + \ln \left( \det \left( {\breve{\bar{\bf K}}}_{j,s-1} + {\breve{\bar{\sigma}}}^2 {\bf I} \right) \right)
  } + \left\| {\breve{\bar{f}}}_{j,s}^{\#} \right\|_{{\breve{\bar{\mathcal{H}}}}_{j,s}} \right)^2 \\
 \end{array}
 }}
\]

Using the uncertainty bound above, we obtain our first bound on instantaneous 
experiment-wise regret based on \cite[Lemma 5.2]{Sri1} with some techniques 
borrowed from \cite{Bog2}:
\begin{lem_srilem52pre}
 Let $\delta \in (0,1)$.  Assume noise variables are $\Sigma$-sub-Gaussian.  
 Assume that:
 \[
  \begin{array}{rl}
   {{  \hat{\beta}}_{i,\underline{t}_{s}}} 
   &\!\!\!\in 
   \left( {  \hat{\bar{\zeta}}}_{i,s\downarrow }{{  \hat{\bar{\chi}}}_{i,s}}, {  \hat{\bar{\zeta}}}_{i,s\uparrow } {{  \hat{\bar{\chi}}}_{i,s}} \right) \\
   {{\breve{\beta}}_{j,\underline{t}_{s}}} 
   &\!\!\!= 
   \left( {\breve{\bar{\zeta}}}_{j,s\downarrow }{{\breve{\bar{\chi}}}_{j,s}}, {\breve{\bar{\zeta}}}_{j,s\uparrow } {{\breve{\bar{\chi}}}_{j,s}} \right) \\
  \end{array}
 \]
 where:
 \[
  {{
  \begin{array}{l}
   {  \hat{\bar{\chi}}}_{i,s} = \left( \frac{\Sigma}{\sqrt{{  \hat{\bar{\sigma}}}}} \sqrt{ 2 \ln \left( \frac{1}{\delta} \right) + 1 + {  \hat{\bar{\gamma}}}_{i,s}
   } + \left\| {  \hat{\bar{f}}}_{i,s}^{\#} \right\|_{{  \hat{\bar{\mathcal{H}}}}_{i,s}} \right)^2 \\
   {\breve{\bar{\chi}}}_{j,s} = \left( \frac{\Sigma}{\sqrt{{\breve{\bar{\sigma}}}}} \sqrt{ 2 \ln \left( \frac{1}{\delta} \right) + 1 + {\breve{\bar{\gamma}}}_{j,s}
   } + \left\| {\breve{\bar{f}}}_{j,s}^{\#} \right\|_{{\breve{\bar{\mathcal{H}}}}_{j,s}} \right)^2 \\
  \end{array}
  }}
 \]
 Then, simultaneously for all $s \geq 1$, the instantaneous regret is bounded as:
 \[
  {{
  \begin{array}{rl}
   {r}_{{  \hat{\underline{t}}}_s+i} 
   &\!\!\!\leq {  \hat{\bar{r}}}_{i,s\uparrow} 
   + 
   \left( 1 - \sqrt{{  \hat{\bar{\zeta}}}_{i,s\downarrow }} \right) \sqrt{{  \hat{\bar{\chi}}}_{i,s}} 
   {  \hat{\bar{\sigma}}}_{i,s-1} \left( {\bf x}^\star \right) \\
   {r}_{{\breve{\underline{t}}}_s+j} 
   &\!\!\!\leq {\breve{\bar{r}}}_{j,s\uparrow} 
   + 
   \left( 1 - \sqrt{{\breve{\bar{\zeta}}}_{j,s\downarrow }} \right) \sqrt{{\breve{\bar{\chi}}}_{j,s}} 
   {\breve{\bar{\sigma}}}_{j,s-1} \left( {\bf x}^\star \right) \\
  \end{array}
  }}
 \]
 with probability $\geq 1-\delta$, where:
 \[
  {{
  \begin{array}{l}
   {  \hat{\bar{r}}}_{i,s\uparrow} = 
   2 \left( \frac{1+\sqrt{{  \hat{\bar{\zeta}}}_{i,s\uparrow }}}{2} \sqrt{{  \hat{\bar{\chi}}}_{i,s}} + \frac{{  \hat{\epsilon}}_{i,\underline{t}_s}}{{  \hat{\bar{\sigma}}}} \sqrt{\underline{t}_{s}} \right) {{  \hat{\bar{\sigma}}}_{i,s-1} \left( {\bf x}_{{  \hat{\underline{t}}}_s+i} \right)} 
   + 2{  \hat{\epsilon}}_{i,\underline{t}_s} \\
   {\breve{\bar{r}}}_{j,s\uparrow} = 
   2 \left( \frac{1+\sqrt{{\breve{\bar{\zeta}}}_{j,s\uparrow }}}{2} \sqrt{{\breve{\bar{\chi}}}_{j,s}} + \frac{{\breve{\epsilon}}_{j,\underline{t}_s}}{{\breve{\bar{\sigma}}}} \sqrt{\underline{t}_{s}} \right) {{\breve{\bar{\sigma}}}_{j,s-1} \left( {\bf x}_{{\breve{\underline{t}}}_s+j} \right)} 
   + 2{\breve{\epsilon}}_{j,\underline{t}_s} \\
  \end{array}
  }}
 \]
 \label{lem:srilem52pre}
\end{lem_srilem52pre}
\begin{proof}
By the pretext and Lemma \ref{lem:boglem1}:
\[
 \begin{array}{rl}
  f^\star \left( {\bf x}^\star \right) 
  &\!\!\!\leq 
  {  \hat{\bar{\mu}}}_{i,s-1} \left( {\bf x}^\star \right) 
  + 
  \left( \sqrt{{  \hat{\bar{\chi}}}_{i,s}} + \frac{{  \hat{\epsilon}}_{i,\underline{t}_s}}{{  \hat{\bar{\sigma}}}} \sqrt{\underline{t}_{s}} \right) {{  \hat{\bar{\sigma}}}_{i,s-1} \left( {\bf x}^\star \right)} + {  \hat{\epsilon}}_{i,\underline{t}_s} \\
%
 \end{array}
\]
with probability $\geq 1-\delta$.  By definition ${\bf x}_t$ maximizes the 
acquisition function, so:
\[
 \begin{array}{rl}
  {  \hat{\bar{\mu}}}_{i,s-1} \left( {\bf x}^\star \right) 
  &\!\!\!\leq
  {  \hat{\bar{\mu}}}_{i,s-1} \left( {\bf x}_{{  \hat{\underline{t}}}_s+i} \right) 
  + 
  \left( \sqrt{{  \hat{\beta}}_{i,\underline{t}_{s}}} + \frac{{  \hat{\epsilon}}_{i,\underline{t}_s}}{{  \hat{\bar{\sigma}}}} \sqrt{\underline{t}_{s}} \right) {{  \hat{\bar{\sigma}}}_{i,s-1} \left( {\bf x}_{{  \hat{\underline{t}}}_s+i} \right)} \\
  &\;\;\;- 
  \left( \sqrt{{  \hat{\beta}}_{i,\underline{t}_{s}}} + \frac{{  \hat{\epsilon}}_{i,\underline{t}_s}}{{  \hat{\bar{\sigma}}}} \sqrt{\underline{t}_{s}} \right) {{  \hat{\bar{\sigma}}}_{i,s-1} \left( {\bf x}^\star \right)} \\
%
 \end{array}
\]
and hence:
\[
 \begin{array}{rl}
  f^\star \left( {\bf x}^\star \right) 
  -
  f^\star \left( {\bf x}_{{  \hat{\underline{t}}}_s+i} \right) 
  &\!\!\!\leq 
  {  \hat{\bar{\mu}}}_{i,\}s-1} \left( {\bf x}_{{  \hat{\underline{t}}}_s+i} \right) 
  + 
  \left( \sqrt{{  \hat{\beta}}_{i,\underline{t}_{s}}} + \frac{{  \hat{\epsilon}}_{i,\underline{t}_s}}{{  \hat{\bar{\sigma}}}} \sqrt{\underline{t}_{s}} \right) {{  \hat{\bar{\sigma}}}_{i,s-1} \left( {\bf x}_{{  \hat{\underline{t}}}_s+i} \right)} + {  \hat{\epsilon}}_{i,\underline{t}_s} \ldots \\
  &\;\;\;+ 
  \left( \sqrt{{  \hat{\bar{\chi}}}_{i,s}} - \sqrt{{  \hat{\beta}}_{i,\underline{t}_{s}}} \right) {{  \hat{\bar{\sigma}}}_{i,s-1} \left( {\bf x}^\star \right)}
  -
  f^\star \left( {\bf x}_{{  \hat{\underline{t}}}_s+i} \right) \\
%
 \end{array}
\]
It follows that the instantaneous regret is bounded by:
\[
 \begin{array}{rl}
  {r}_{{  \hat{\underline{t}}}_s+i} 
  &\!\!\!\leq 
  {  \hat{\bar{\mu}}}_{i,s-1} \left( {\bf x}_{{  \hat{\underline{t}}}_s+i} \right) 
  + 
  \left( \sqrt{{  \hat{\beta}}_{i,\underline{t}_{s}}} + \frac{{  \hat{\epsilon}}_{i,\underline{t}_s}}{{  \hat{\bar{\sigma}}}} \sqrt{\underline{t}_{s}} \right) {{  \hat{\bar{\sigma}}}_{i,s-1} \left( {\bf x}_{{  \hat{\underline{t}}}_s+i} \right)}
  +
  {  \hat{\epsilon}}_{i,\underline{t}_s} \ldots \\
  &\;\;\;+ 
  \left( \sqrt{{  \hat{\bar{\chi}}}_{i,s}} - \sqrt{{  \hat{\beta}}_{i,\underline{t}_{s}}} \right) {{  \hat{\bar{\sigma}}}_{i,s-1} \left( {\bf x}^\star \right)}
  -
  f^\star \left( {\bf x}_{{  \hat{\underline{t}}}_s+i} \right) \\
%
 \end{array}
\]
So, using our assumptions on $\beta$ we find that:
\[
 \begin{array}{rl}
  {r}_{{  \hat{\underline{t}}}_s+i} 
  &\!\!\!\leq 
  {  \hat{\bar{\mu}}}_{i,s-1} \left( {\bf x}_{{  \hat{\underline{t}}}_s+i} \right) 
  + 
  \left( \sqrt{{  \hat{\bar{\zeta}}}_{i,s\uparrow }} \sqrt{{  \hat{\bar{\chi}}}_{i,s}} + \frac{{  \hat{\epsilon}}_{i,\underline{t}_s}}{{  \hat{\bar{\sigma}}}} \sqrt{\underline{t}_{s}} \right) {{  \hat{\bar{\sigma}}}_{i,s-1} \left( {\bf x}_{{  \hat{\underline{t}}}_s+i} \right)} 
  + 
  {  \hat{\epsilon}}_{i,\underline{t}_s} \ldots \\
  &\;\;\;+ 
  \left( 1 - \sqrt{{  \hat{\bar{\zeta}}}_{i,s\downarrow }} \right) \sqrt{{  \hat{\bar{\chi}}}_{i,s}} {  \hat{\bar{\sigma}}}_{i,s-1} \left( {\bf x}^\star \right)
  -
  f^\star \left( {\bf x}_{{  \hat{\underline{t}}}_s+i} \right) \\
%
 \end{array}
\]
Once again using our pretext and Lemma \ref{lem:boglem1} we see that:
\[
 \begin{array}{rl}
  f^\star \left( {\bf x}_{{  \hat{\underline{t}}}_s+i} \right) 
  &\!\!\!\geq 
  {  \hat{\bar{\mu}}}_{i,s-1} \left( {\bf x}_{{  \hat{\underline{t}}}_s+i} \right) 
  - 
  \left( \sqrt{{  \hat{\bar{\chi}}}_{i,s}} + \frac{{  \hat{\epsilon}}_{i,\underline{t}_s}}{{  \hat{\bar{\sigma}}}} \sqrt{\underline{t}_{s}} \right) {{  \hat{\bar{\sigma}}}_{i,s-1} \left( {\bf x}_{{  \hat{\underline{t}}}_s+i} \right)} 
  - 
  {  \hat{\epsilon}}_{i,\underline{t}_s} \\
%
 \end{array}
\]
Hence:
\[
 \begin{array}{rl}
  {r}_{{  \hat{\underline{t}}}_s+i} 
  &\!\!\!\leq 
  2 \left( \frac{1+\sqrt{{  \hat{\bar{\zeta}}}_{i,s\uparrow }}}{2} \sqrt{{  \hat{\bar{\chi}}}_{i,s}} + \frac{{  \hat{\epsilon}}_{i,\underline{t}_s}}{{  \hat{\bar{\sigma}}}} \sqrt{\underline{t}_{s}} \right) {{  \hat{\bar{\sigma}}}_{i,s-1} \left( {\bf x}_{{  \hat{\underline{t}}}_s+i} \right)} 
  \ldots \\
  &\;\;\;
  + \left( 1 - \sqrt{{  \hat{\bar{\zeta}}}_{i,s\downarrow }} \right) \sqrt{{  \hat{\bar{\chi}}}_{i,s}} {  \hat{\bar{\sigma}}}_{i,s-1} \left( {\bf x}^\star \right)
  + 2{  \hat{\epsilon}}_{i,\underline{t}_s} \\
%
 \end{array}
\]
and the desired result follows from the definitions.  The proof for AI regret 
follows by an analogous argument.
\end{proof}

To extend this to usable batch-wise instantaneous regret bound we need to deal 
with the variance terms ${\hat{\bar{\sigma}}}_{i,s-1} ({\bf x}^\star)$, 
${\breve{\bar{\sigma}}}_{i,s-1} ({\bf x}^\star)$ in the above theorem.  To do 
this, in the following theorem we use the generalized (power) mean, and in 
particular lemma \ref{lem:neginfnorm}, to split the batch-wise risk bound into 
a constraint term containing the free variances and a risk bound that depends 
only on the various parameters of the problem:

\begin{lem_srilem52hmm}
 Fix $\delta > 0$ and ${\bar{\theta}}_s \in [0,\infty]$.\footnote{The proof is 
 true for ${\bar{\theta}}_s \in [-\infty,\infty]$, but the negative 
 ${\bar{\theta}}_s$  case is not of interest here.}  Assume noise variables are 
 $\Sigma$-sub-Gaussian, and that:
 \[
  \begin{array}{rl}
   {{  \hat{\beta}}_{i,\underline{t}_{s}}} 
   &\!\!\!\in 
   \left( {  \hat{\bar{\zeta}}}_{i,s\downarrow }{{  \hat{\bar{\chi}}}_{i,s}}, {  \hat{\bar{\zeta}}}_{i,s\uparrow } {{  \hat{\bar{\chi}}}_{i,s}} \right) \\
   {{\breve{\beta}}_{j,\underline{t}_{s}}} 
   &\!\!\!= 
   \left( {\breve{\bar{\zeta}}}_{j,s\downarrow }{{\breve{\bar{\chi}}}_{j,s}}, {\breve{\bar{\zeta}}}_{j,s\uparrow } {{\breve{\bar{\chi}}}_{j,s}} \right) \\
  \end{array}
 \]
 where ${  \hat{\bar{\zeta}}}_{i,s\downarrow } \leq 1$, ${\breve{\bar{\zeta}}}_{j, 
 s\downarrow } \geq 1$, and:
 \[
  {{
  \begin{array}{l}
   {  \hat{\bar{\chi}}}_{i,s} = \left( \frac{\Sigma}{\sqrt{{  \hat{\bar{\sigma}}}}} \sqrt{ 2 \ln \left( \frac{1}{\delta} \right) + 1 + {  \hat{\bar{\gamma}}}_{i,s}
   } + \left\| {  \hat{\bar{f}}}_{i,s}^{\#} \right\|_{{  \hat{\bar{\mathcal{H}}}}_{i,s}} \right)^2 \\
   {\breve{\bar{\chi}}}_{j,s} = \left( \frac{\Sigma}{\sqrt{{\breve{\bar{\sigma}}}}} \sqrt{ 2 \ln \left( \frac{1}{\delta} \right) + 1 + {\breve{\bar{\gamma}}}_{j,s}
   } + \left\| {\breve{\bar{f}}}_{j,s}^{\#} \right\|_{{\breve{\bar{\mathcal{H}}}}_{j,s}} \right)^2 \\
  \end{array}
  }}
 \]
 If:
 \[
  {{
  \begin{array}{r}
   M_{{\bar{\theta}}_s} \left( \begin{array}{c} \left\{ 
   \exp \left( \sqrt{{  \hat{\bar{\chi}}}_{i,s}} \left( 1 - \sqrt{{  \hat{\bar{\zeta}}}_{i,s\downarrow }} \right)_+ {\bar{\sigma}}_{s-1  \uparrow} \right)
   : {i \in {  \hat{\mathbb{M}}}}
   \right\} \bigcup \ldots \\ \left\{
   \exp \left( \sqrt{{\breve{\bar{\chi}}}_{j,s}} \left( 1 - \sqrt{{\breve{\bar{\zeta}}}_{j,s\downarrow }} \right)_- {\bar{\sigma}}_{s-1\downarrow} \right)
   : {j \in {\breve{\mathbb{M}}}}
   \right\} \end{array} \right) \leq 1
  \end{array}
  }}
 \]
 Then, simultaneously for all $s \geq 1$, the batch-wise instantaneous regret is bounded as:
 \[
  {{
  \begin{array}{rl}
   \bar{r}_s 
   \leq 
   \ln \left( M_{-{\bar{\theta}}_s} \left( \exp \left( {\bar{\bf r}}_{s\uparrow} \right) \right) \right)
  \end{array}
  }}
 \]
 with probability $\geq 1-\delta$, where:
 \[
  \begin{array}{l}
   {\bar{\bf r}}_{s\uparrow} = 
   \left[ \begin{array}{c} 
    \left[ \begin{array}{c}
     2 \left( \frac{1+\sqrt{{  \hat{\bar{\zeta}}}_{i,s\uparrow }}}{2} \sqrt{{  \hat{\bar{\chi}}}_{i,s}} + \frac{{  \hat{\epsilon}}_{i,\underline{t}_s}}{{  \hat{\bar{\sigma}}}} \sqrt{\underline{t}_{s}} \right) {{  \hat{\bar{\sigma}}}_{i,s-1} \left( {\bf x}_{{  \hat{\underline{t}}}_s+i} \right)} 
     + 2{  \hat{\epsilon}}_{i,\underline{t}_s} \\
    \end{array} \right]_{i \in {  \hat{\mathbb{M}}}} \\
    \left[ \begin{array}{c}
     2 \left( \frac{1+\sqrt{{\breve{\bar{\zeta}}}_{j,s\uparrow }}}{2} \sqrt{{\breve{\bar{\chi}}}_{j,s}} + \frac{{\breve{\epsilon}}_{j,\underline{t}_s}}{{\breve{\bar{\sigma}}}} \sqrt{\underline{t}_{s}} \right) {{\breve{\bar{\sigma}}}_{j,s-1} \left( {\bf x}_{{\breve{\underline{t}}}_s+j} \right)} 
     + 2{\breve{\epsilon}}_{j,\underline{t}_s} \\
    \end{array} \right]_{i \in {  \hat{\mathbb{M}}}} \\
   \end{array} \right]
  \end{array}
 \]
 and:
 \[
  \begin{array}{rl}
   {\bar{\sigma}}_{s-1  \uparrow} &\!\!\!= \max \left\{ \mathop{\max}\limits_{i \in {  \hat{\mathbb{M}}}} \left\{ {  \hat{\bar{\sigma}}}_{i,s-1} \left( {\bf x}^\star \right) \right\}, 
                                                        \mathop{\max}\limits_{j \in {\breve{\mathbb{M}}}} \left\{ {\breve{\bar{\sigma}}}_{j,s-1} \left( {\bf x}^\star \right) \right\} \right\} \\
   {\bar{\sigma}}_{s-1\downarrow} &\!\!\!= \min \left\{ \mathop{\min}\limits_{i \in {  \hat{\mathbb{M}}}} \left\{ {  \hat{\bar{\sigma}}}_{i,s-1} \left( {\bf x}^\star \right) \right\}, 
                                                        \mathop{\min}\limits_{j \in {\breve{\mathbb{M}}}} \left\{ {\breve{\bar{\sigma}}}_{j,s-1} \left( {\bf x}^\star \right) \right\} \right\} \\
  \end{array}
 \]
 and $\exp$ is applied element-wise.
 \label{lem:srilem52hmm}
\end{lem_srilem52hmm}
\begin{proof}
It is convenient to re-frame the batch-wise instantaneous regret in log-space, 
and extract the max variance upper bound.  
Recalling that ${\hat{\bar{\zeta}}}_{i,s\downarrow} 
\leq 1$, ${\breve{\bar{\zeta}}}_{j,s\downarrow } \geq 
1$:
\[
{\!\!\!\!\!\!\!\!\!\!\!\!\!\!\!\!\!\!{
 \begin{array}{rl}
  \bar{r}_s 

  \leq 
  \min \Bigg\{
  
  &\!\!\!\mathop{\min}\limits_{i \in {  \hat{\mathbb{M}}}} \left\{
  \begin{array}{r}
  2 \left( \frac{1+\sqrt{{  \hat{\bar{\zeta}}}_{j,s\uparrow }}}{2} \sqrt{{  \hat{\bar{\chi}}}_{i,s}} + \frac{{  \hat{\epsilon}}_{i,\underline{t}_s}}{{  \hat{\bar{\sigma}}}} \sqrt{\underline{t}_{s}} \right) {{  \hat{\bar{\sigma}}}_{i,s-1} \left( {\bf x}_{{  \hat{\underline{t}}}_s+i} \right)} 
  + \ldots \\ \hfill 
  \left( 1 - \sqrt{{  \hat{\bar{\zeta}}}_{i,s\downarrow }} \right) \sqrt{{  \hat{\bar{\chi}}}_{i,s}} {  \hat{\bar{\sigma}}}_{i,s-1} \left( {\bf x}^\star \right)
  + 2{  \hat{\epsilon}}_{i,\underline{t}_s}
  \end{array}
  \right\}, \\

  &\!\!\!\mathop{\min}\limits_{j \in {\breve{\mathbb{M}}}} \left\{
  \begin{array}{r}
  2 \left( \frac{1+\sqrt{{\breve{\bar{\zeta}}}_{j,s\uparrow }}}{2} \sqrt{{\breve{\bar{\chi}}}_{j,s}} + \frac{{\breve{\epsilon}}_{j,\underline{t}_s}}{{\breve{\bar{\sigma}}}} \sqrt{\underline{t}_{s}} \right) {{\breve{\bar{\sigma}}}_{j,s-1} \left( {\bf x}_{{\breve{\underline{t}}}_s+j} \right)} 
  + \ldots \\ \hfill 
  \left( 1 - \sqrt{{\breve{\bar{\zeta}}}_{j,s\downarrow }} \right) \sqrt{{\breve{\bar{\chi}}}_{j,s}} {\breve{\bar{\sigma}}}_{j,s-1} \left( {\bf x}^\star \right)
  + 2{\breve{\epsilon}}_{j,\underline{t}_s}
  \end{array}
  \right\}

  \Bigg\} \\

  \leq 
  \ln \Bigg( \min \Bigg\{
  
  &\!\!\!\mathop{\min}\limits_{i \in {  \hat{\mathbb{M}}}} \left\{ 
  \begin{array}{r}
  \exp \left(
  2 \left( \frac{1+\sqrt{{  \hat{\bar{\zeta}}}_{i,s\uparrow }}}{2} \sqrt{{  \hat{\bar{\chi}}}_{i,s}} + \frac{{  \hat{\epsilon}}_{i,\underline{t}_s}}{{  \hat{\bar{\sigma}}}} \sqrt{\underline{t}_{s}} \right) {{  \hat{\bar{\sigma}}}_{i,s-1} \left( {\bf x}_{{  \hat{\underline{t}}}_s+i} \right)} 
  + 2 {  \hat{\epsilon}}_{i,\underline{t}_s} 
  \right) 
  \ldots \\ \hfill 
  \exp \left(
  \sqrt{{  \hat{\bar{\chi}}}_{i,s}} \left( 1 - \sqrt{{  \hat{\bar{\zeta}}}_{i,s\downarrow }} \right)_+ {\bar{\sigma}}_{s-1  \uparrow}
  \right) 
  \end{array}
  \right\}, \\

  &\!\!\!\mathop{\min}\limits_{j \in {\breve{\mathbb{M}}}} \left\{ 
  \begin{array}{r}
  \exp \left(
  2 \left( \frac{1+\sqrt{{\breve{\bar{\zeta}}}_{j,s \uparrow }}}{2} \sqrt{{\breve{\bar{\chi}}}_{j,s}} + \frac{{\breve{\epsilon}}_{j,\underline{t}_s}}{{\breve{\bar{\sigma}}}} \sqrt{\underline{t}_{s}} \right) {{\breve{\bar{\sigma}}}_{j,s-1} \left( {\bf x}_{{\breve{\underline{t}}}_s+j} \right)} 
  + 2 {\breve{\epsilon}}_{j,\underline{t}_s}
  \right) 
  \ldots \\ \hfill 
  \exp \left(
  \sqrt{{\breve{\bar{\chi}}}_{j,s}} \left( 1 - \sqrt{{\breve{\bar{\zeta}}}_{j,s \downarrow }} \right)_- {\bar{\sigma}}_{s-1\downarrow}
  \right) 
  \end{array}
  \right\}

  \Bigg\} \Bigg) \\

 \end{array}
 }}
\]
which may be re-written:
\[
 \begin{array}{rl}
  \bar{r}_s 
  \leq 
  \ln \left( M_{-\infty} \left( {\bf a} \odot {\bf b} \right) \right) \\
 \end{array}
\]
where:
\[
 \begin{array}{ll}
  {\bf a} = \left[ \begin{array}{c} {  \hat{\bf a}} \\ {\breve{\bf a}} \\ \end{array} \right] \in \mathbb{R}_+^p, &
  {\bf b} = \left[ \begin{array}{c} {  \hat{\bf b}} \\ {\breve{\bf b}} \\ \end{array} \right] \in \mathbb{R}_+^p \\
 \end{array}
\]
\[
{{
 \begin{array}{l}
  {  \hat{\bf a}} = \left[ \begin{array}{c} 
  \exp \left(
  2 \left( \frac{1+\sqrt{{  \hat{\bar{\zeta}}}_{i,s \uparrow }}}{2} \sqrt{{  \hat{\bar{\chi}}}_{i,s}} + \frac{{  \hat{\epsilon}}_{i,\underline{t}_s}}{{  \hat{\bar{\sigma}}}} \sqrt{\underline{t}_{s}} \right) {{  \hat{\bar{\sigma}}}_{i,s-1} \left( {\bf x}_{{  \hat{\underline{t}}}_s+i} \right)} 
  + 2 {  \hat{\epsilon}}_{i,\underline{t}_s} 
  \right)
  \end{array} \right]_{i \in {  \hat{\mathbb{M}}}} \\ 
  
  {  \hat{\bf b}} = \left[ \begin{array}{c} 
  \exp \left(
   \sqrt{{  \hat{\bar{\chi}}}_{i,s}} \left( 1 - \sqrt{{  \hat{\bar{\zeta}}}_{i,s\downarrow }} \right)_+ {\bar{\sigma}}_{s-1  \uparrow}
  \right)
  \end{array} \right]_{i \in {  \hat{\mathbb{M}}}} \\ 
  
  {\breve{\bf a}} = \left[ \begin{array}{c} 
  \exp \left(
  2 \left( \frac{1+\sqrt{{\breve{\bar{\zeta}}}_{j,s\uparrow }}}{2} \sqrt{{\breve{\bar{\chi}}}_{j,s}} + \frac{{\breve{\epsilon}}_{j,\underline{t}_s}}{{\breve{\bar{\sigma}}}} \sqrt{\underline{t}_{s}} \right) {{\breve{\bar{\sigma}}}_{j,s-1} \left( {\bf x}_{{\breve{\underline{t}}}_s+j} \right)} 
  + 2 {\breve{\epsilon}}_{j,\underline{t}_s}
  \right) 
  \end{array} \right]_{j \in {\breve{\mathbb{M}}}} \\ 
  
  {\breve{\bf b}} = \left[ \begin{array}{c} 
  \exp \left(
   \sqrt{{\breve{\bar{\chi}}}_{j,s}} \left( 1 - \sqrt{{\breve{\bar{\zeta}}}_{j,s\downarrow }} \right)_- {\bar{\sigma}}_{s-1\downarrow}
  \right)
  \end{array} \right]_{j \in {\breve{\mathbb{M}}}} \\ 
 \end{array}
}}
\]
So, by lemma \ref{lem:neginfnorm}, we have that:
\[
 \begin{array}{rl}
  \bar{r}_s 
  \leq 
  \ln \left( M_{-{\bar{\theta}}_s} \left( {\bf a} \right) M_{{\bar{\theta}}_s} \left( {\bf b} \right) \right) 
  = 
  \ln \left( M_{-{\bar{\theta}}_s} \left( {\bf a} \right) \right) + \ln \left( M_{{\bar{\theta}}_s} \left( {\bf b} \right) \right) \\
 \end{array}
\]
and by assumption $M_{{\bar{\theta}}_s} \left( {\bf b} \right) \leq 1$, so:
\[
 \begin{array}{rl}
  \bar{r}_s 
  \leq 
  \ln \left( M_{-{\bar{\theta}}_s} \left( {\bf a} \right) \right) \\
 \end{array}
\]
and the result follows by the definition.
\end{proof}

The next step is to convert this bound into a bound on the total regret.  To 
obtain such a bound we need some additional assumptions regarding ``gap'' 
parameters $\epsilon$ and the explorative/exploitative nature of the humans 
and AIs.  We do this with the following theorem:
\begin{th_totregbndb}
 Fix $\delta > 0$ and ${\bar{\theta}}_s \in [0,\infty]$.\footnote{The proof is 
 true for ${\bar{\theta}}_s \in [-\infty,\infty]$, but the negative 
 ${\bar{\theta}}_s$  case is not of interest here.}  Assume noise variables are 
 $\Sigma$-sub-Gaussian, and that:
 \[
\begin{array}{rl}
   {{  \hat{\beta}}_{i,\underline{t}_{s}}} 
   &\!\!\!\in 
   \left( {  \hat{\bar{\zeta}}}_{i,s\downarrow }{{  \hat{\bar{\chi}}}_{i,s}}, {  \hat{\bar{\zeta}}}_{i,s\uparrow } {{  \hat{\bar{\chi}}}_{i,s}} \right) \\
   {{\breve{\beta}}_{j,\underline{t}_{s}}} 
   &\!\!\!= 
   \left( {\breve{\bar{\zeta}}}_{j,s\downarrow }{{\breve{\bar{\chi}}}_{j,s}}, {\breve{\bar{\zeta}}}_{j,s\uparrow } {{\breve{\bar{\chi}}}_{j,s}} \right) \\
  \end{array}%
 \]
 If ${\hat{\epsilon}}_{i,\underline{t}_s}, {\breve{\epsilon}}_{j,\underline{t}_s} 
 = \mathcal{O} (s^{-q})$ for some $q \in (\frac{1}{2},\infty)$ then, with 
 probability $\geq 1-\delta$:
 \[
{\!\!\!\!\!\!\!\!\!\!\!{
  \begin{array}{l}
   \frac{1}{S} {\bar{R}}_S

   \leq 
  
   \ln \Bigg( M_{-{\bar{\theta}}_\downarrow} \Bigg( \exp \Bigg( \Bigg\{
  
   \left\{ \left.
   \sqrt{ 8 {  \hat{\bar{\sigma}}}^2 \left( \sqrt{{  \hat{\bar{\chi}}}_{i,S}} + \mathcal{O} \left( 1 \right) \right)^2 {  \hat{C}}_{i,2} \frac{1}{S} {  \hat{\bar{\gamma}}}_{i,S} +
   \mathcal{O} \left( 1 \right) } 
   \right| i \in {  \hat{\mathbb{M}}} \right\} \bigcup \ldots  \\

   \hfill \left\{ \left.
   \sqrt{ 8 {\breve{\bar{\sigma}}}^2 \left( \frac{1+\sqrt{{\breve{\bar{\zeta}}}_{j\sim}}}{2} \sqrt{{\breve{\bar{\chi}}}_{j,S}} + \mathcal{O} \left( 1 \right) \right)^2 {\breve{C}}_{j,2} \frac{1}{S} {\breve{\bar{\gamma}}}_{j,S} +
   \mathcal{O} \left( 1 \right) } 
   \right| j \in {\breve{\mathbb{M}}} \right\}
  
   \Bigg\} \Bigg) \Bigg) \Bigg) \\
  \\
  \end{array}
 }}
 \]
 where ${\hat{C}}_{i,2}= \max_s \frac{{\hat{\bar{\sigma}}}^{-2} {\hat{\bar{K}}}_{i,s 
 \uparrow}}{\ln \left( 1 + {\hat{\bar{\sigma}}}^{-2} {\hat{\bar{K}}}_{i,s 
 \uparrow} \right)}$ and ${\breve{C}}_{j,2} = \max_s \frac{{\breve{\bar{\sigma}}}^{-2} 
 {\breve{\bar{K}}}_{j,s\uparrow}}{\ln \left( 1 + {\breve{\bar{\sigma}}}^{-2} 
 {\breve{\bar{K}}}_{j,\}s\uparrow} \right)}$; and:
 \[
  \begin{array}{rl}
   {\bar{\theta}}_\downarrow &\!\!\!\!= \mathop{\min}\limits_{s \in \mathbb{N}_S+1} {\bar{\theta}}_s \\
   {\breve{\bar{\zeta}}}_{j\sim} &\!\!\!\!= \mathop{\max}\limits_{s \in \mathbb{N}_S+1} {\breve{\bar{\zeta}}}_{j,s}
  \end{array}
 \]
\label{th:totregbndb}
\end{th_totregbndb}
\begin{proof}
Using the assumptions and Lemma \ref{lem:srilem52hmm}, we have that, with 
probability $\geq 1-\delta$, simultaneously for all $s \geq 1$:
\[
 {{
 \begin{array}{rl}
  \bar{r}_s 
  \leq 
  \ln \left( M_{-{\theta}_s} \left( \exp \left( {\bar{\bf r}}_{s\uparrow} \right) \right) \right)
  \leq 
  \ln \left( M_{-{\bar{\theta}}_s} \left( \exp \left( {\bar{\bf r}}_{s\uparrow} \right) \right) \right)
 \end{array}
 }}
\]
where ${\bar{\bf r}}_{s\uparrow} \geq {\bf 0}$.  Using the definition of 
$M_{-{\bar{\theta}}_s}$:
\[
 {\!\!\!\!\!\!\!\!\!\!\!\!\!\!\!\!\!\!\!\!\!\!\!\!\!\!\!\!\!\!\!\!\!{
 \begin{array}{rl}
  \mathop{\sum}\limits_{s \in \mathbb{N}_S+1} 
  \bar{r}_s 
  &\!\!\!\leq 
  \mathop{\sum}\limits_{s \in \mathbb{N}_S+1} 
  \ln \left( M_{-{\bar{\theta}}_\downarrow} \left( \exp \left( {\bar{\bf r}}_{s\uparrow} \right) \right) \right) \\
  &\!\!\!=
  \ln \left( \mathop{\prod}\limits_{s \in \mathbb{N}_S+1} M_{-{\bar{\theta}}_\downarrow} \left( \exp \left( {\bar{\bf r}}_{s\uparrow} \right) \right) \right) \\
  &\!\!\!=
  \ln \left( \mathop{\prod}\limits_{s \in \mathbb{N}_S+1} \left( \frac{1}{p} \left\| \exp \left( {\bar{\bf r}}_{s\uparrow} \right) \right\|_{-{\bar{\theta}}_\downarrow} \right)^{-\frac{1}{{\bar{\theta}}_\downarrow}} \right) \\
  &\!\!\!=
  \ln \left( \left( \mathop{\prod}\limits_{s \in \mathbb{N}_S+1} p^{\frac{S}{{\bar{\theta}}_\downarrow}} \left\| \left( \exp \left( {\bar{\bf r}}_{s\uparrow} \right) \right)^{\odot -{\bar{\theta}}_\downarrow} \right\|_1^{-\frac{1}{{\bar{\theta}}_\downarrow}} \right) \right) \\
  &\!\!\!=
  \ln \left( \left( \mathop{\prod}\limits_{s \in \mathbb{N}_S+1} p^{\frac{S}{{\bar{\theta}}_\downarrow}} \left( \left\| \left( \left( \exp \left( {\bar{\bf r}}_{s\uparrow} \right) \right)^{\odot -\frac{{\bar{\theta}}_\downarrow}{S}} \right)^{\odot S} \right\|_1^{\frac{1}{S}} \right)^{-\frac{S}{{\bar{\theta}}_\downarrow}} \right) \right) \\
  &\!\!\!=
  \ln \left( \left( \mathop{\prod}\limits_{s \in \mathbb{N}_S+1} p^{\frac{S}{{\bar{\theta}}_\downarrow}} \left\| \left( \exp \left( {\bar{\bf r}}_{s\uparrow} \right) \right)^{\odot -\frac{{\bar{\theta}}_\downarrow}{S}} \right\|_S^{-\frac{S}{{\bar{\theta}}_\downarrow}} \right) \right) \\
 \end{array}
 }}
\]
Using the generalized H{\"o}lder inequality and the definition of 
$M_{-{\bar{\theta}}_\downarrow}$:
\[
 {\!\!\!\!\!\!\!\!\!\!\!\!\!\!\!\!\!\!\!\!\!\!{
 \begin{array}{rl}
  \mathop{\sum}\limits_{s \in \mathbb{N}_S+1} 
  \bar{r}_s 
  &\!\!\!\leq 
  \ln \left( \left( \mathop{\prod}\limits_{s \in \mathbb{N}_S+1} p^{\frac{S}{{\bar{\theta}}_\downarrow}} \left\| \left( \exp \left( {\bar{\bf r}}_{s\uparrow} \right) \right)^{\odot -\frac{{\bar{\theta}}_\downarrow}{S}} \right\|_S^{-\frac{S}{{\bar{\theta}}_\downarrow}} \right) \right) \\
  &\!\!\!\leq
  \ln \left( \left( \left\| \mathop{\bigodot}\limits_{s \in \mathbb{N}_S+1} p^{\frac{S}{{\bar{\theta}}_\downarrow}} \left( \exp \left( {\bar{\bf r}}_{s\uparrow} \right) \right)^{\odot -\frac{{\bar{\theta}}_\downarrow}{S}} \right\|_1^{-\frac{S}{{\bar{\theta}}_\downarrow}} \right) \right) \\
  &\!\!\!=
  \ln \left( \left( \left( \frac{1}{p} \left\| \left( \mathop{\bigodot}\limits_{s \in \mathbb{N}_S+1} \exp \left( {\bar{\bf r}}_{s\uparrow} \right) \right)^{\odot -\frac{{\bar{\theta}}_\downarrow}{S}} \right\|_1 \right)^{-\frac{S}{{\bar{\theta}}_\downarrow}} \right) \right) \\
  &\!\!\!=
  \ln \left( M_{-\frac{{\bar{\theta}}_\downarrow}{S}} \left( \mathop{\bigodot}\limits_{s \in \mathbb{N}_S+1} \exp \left( {\bar{\bf r}}_{s\uparrow} \right) \right) \right) \\
 \end{array}
 }}
\]
and so, again recalling that ${\bar{\bf r}}_{s\uparrow} \geq {\bf 0}$:
\[
 {{
 \begin{array}{rl}
  \mathop{\sum}\limits_{s \in \mathbb{N}_S+1} 
  \bar{r}_s 
  &\!\!\!\leq 
  \ln \left( M_{-\frac{{\bar{\theta}}_\downarrow}{S}} \left( \mathop{\bigodot}\limits_{s \in \mathbb{N}_S+1} \exp \left( {\bar{\bf r}}_{s\uparrow} \right) \right) \right) \\
  &\!\!\!=
  \ln \left( M_{-\frac{{\bar{\theta}}_\downarrow}{S}} \left( \exp \left( \mathop{\sum}\limits_{s \in \mathbb{N}_S+1} {\bar{\bf r}}_{s\uparrow} \right) \right) \right) \\
 \end{array}
 }}
\]
Next, using that $\chi$ is increasing with $s$, and noting our restricted range 
in $\zeta$, note that:
\[
 \begin{array}{rl}
  {  \hat{\bar{r}}}_{i,s\uparrow}^2 
  &\!\!\!\leq 
  4 \left( \left( \sqrt{{  \hat{\bar{\chi}}}_{i,S}} + \frac{{  \hat{\epsilon}}_{i,\underline{t}_s}}{{  \hat{\bar{\sigma}}}} \sqrt{\underline{t}_{s}} \right) {{  \hat{\bar{\sigma}}}_{i,s-1} \left( {\bf x}_{{  \hat{\underline{t}}}_s+i}
   \right)} 
  + {  \hat{\epsilon}}_{i,\underline{t}_s} \right)^2 \\
  &\!\!\!\leq 
  8 \max \left\{ \left( \sqrt{{  \hat{\bar{\chi}}}_{i,S}} + \frac{{  \hat{\epsilon}}_{i,\underline{t}_s}}{{  \hat{\bar{\sigma}}}} \sqrt{\underline{t}_{s}} \right)^2 {{  \hat{\bar{\sigma}}}^2_{i,s-1} \left( {\bf x}_{{  \hat{\underline{t}}}_s+i} \right)}, 
  {  \hat{\epsilon}}^2_{i,\underline{t}_s} \right\} \\
  {\breve{\bar{r}}}_{j,s\uparrow}^2 
  &\!\!\!\leq 
  4 \left( \left( \frac{1+\sqrt{{\breve{\bar{\zeta}}}_{j\sim}}}{2} \sqrt{{\breve{\bar{\chi}}}_{j,S}} + \frac{{\breve{\epsilon}}_{j,\underline{t}_s}}{{\breve{\bar{\sigma}}}} \sqrt{\underline{t}_{s}} \right) {{\breve{\bar{\sigma}}}_{j,s-1} \left( {\bf x}_{{\breve{\underline{t}}}_s+j}
   \right)} 
  + {\breve{\epsilon}}_{j,\underline{t}_s} \right)^2 \\
  &\!\!\!\leq 
  8 \max \left\{ \left( \frac{1+\sqrt{{\breve{\bar{\zeta}}}_{j\sim}}}{2} \sqrt{{\breve{\bar{\chi}}}_{j,S}} + \frac{{\breve{\epsilon}}_{j,\underline{t}_s}}{{\breve{\bar{\sigma}}}} \sqrt{\underline{t}_{s}} \right)^2 {{\breve{\bar{\sigma}}}^2_{j,s-1} \left( {\bf x}_{{\breve{\underline{t}}}_s+j} \right)}, 
  {\breve{\epsilon}}^2_{j,\underline{t}_s} \right\} \\
 \end{array}
\]
and so:
\[
 \begin{array}{rl}
  \mathop{\sum}\limits_{s \in \mathbb{N}_S+1} {  \hat{\bar{r}}}_{i,s\uparrow}^2 
  &\!\!\!\leq 
  8 \mathop{\sum}\limits_{s \in \mathbb{N}_S+1} \max \left\{ \left( \sqrt{{  \hat{\bar{\chi}}}_{i,S}} + \frac{{  \hat{\epsilon}}_{i,\underline{t}_s}}{{  \hat{\bar{\sigma}}}} \sqrt{\underline{t}_{s}} \right)^2 {{  \hat{\bar{\sigma}}}^2_{i,s-1} \left( {\bf x}_{{  \hat{\underline{t}}}_s+i} \right)}, 
  {  \hat{\epsilon}}^2_{i,\underline{t}_s} \right\} \\
  &\!\!\!\leq 
  8 \mathop{\sum}\limits_{s \in \mathbb{N}_S+1} \left( \sqrt{{  \hat{\bar{\chi}}}_{i,S}} + \frac{{  \hat{\epsilon}}_{i,\underline{t}_s}}{{  \hat{\bar{\sigma}}}} \sqrt{\underline{t}_{s}} \right)^2 {{  \hat{\bar{\sigma}}}^2_{i,s-1} \left( {\bf x}_{{  \hat{\underline{t}}}_s+i} \right)} +  
  8 \mathop{\sum}\limits_{s \in \mathbb{N}_S+1} {  \hat{\epsilon}}^2_{i,\underline{t}_s} \\
  \mathop{\sum}\limits_{s \in \mathbb{N}_S+1} {\breve{\bar{r}}}_{j,s\uparrow}^2 
  &\!\!\!\leq 
  8 \mathop{\sum}\limits_{s \in \mathbb{N}_S+1} \max \left\{ \left( \frac{1+\sqrt{{\breve{\bar{\zeta}}}_{j\sim}}}{2} \sqrt{{\breve{\bar{\chi}}}_{j,S}} + \frac{{\breve{\epsilon}}_{j,\underline{t}_s}}{{\breve{\bar{\sigma}}}} \sqrt{\underline{t}_{s}} \right)^2 {{\breve{\bar{\sigma}}}^2_{j,s-1} \left( {\bf x}_{{\breve{\underline{t}}}_s+j} \right)}, 
  {\breve{\epsilon}}^2_{j,\underline{t}_s} \right\} \\
  &\!\!\!\leq 
  8 \mathop{\sum}\limits_{s \in \mathbb{N}_S+1} \left( \frac{1+\sqrt{{\breve{\bar{\zeta}}}_{j\sim}}}{2} \sqrt{{\breve{\bar{\chi}}}_{j,S}} + \frac{{\breve{\epsilon}}_{j,\underline{t}_s}}{{\breve{\bar{\sigma}}}} \sqrt{\underline{t}_{s}} \right)^2 {{\breve{\bar{\sigma}}}^2_{j,s-1} \left( {\bf x}_{{\breve{\underline{t}}}_s+j} \right)} +  
  8 \mathop{\sum}\limits_{s \in \mathbb{N}_S+1} {\breve{\epsilon}}^2_{j,\underline{t}_s} \\
 \end{array}
\]
Recalling our assumption ${  \hat{\epsilon}}_{i,\underline{t}_s}, 
{\breve{\epsilon}}_{j,\underline{t}_s} = \mathcal{O} (s^{-q})$ for 
some $q > \frac{1}{2}$ we find that, as $\sum_{s\in\mathbb{N}} \frac{1}{s^{2 
\hat{q}_{i}}} = \zeta (2{  \hat{q}_{i}}) < \infty$ 
($\zeta$ here is the Reimann zeta function):
\[
 \begin{array}{rl}
  \mathop{\sum}\limits_{s \in \mathbb{N}_S+1} {  \hat{\bar{r}}}_{i,s\uparrow}^2 
  &\!\!\!\leq 
  8 \mathop{\sum}\limits_{s \in \mathbb{N}_S+1} \left( \sqrt{{  \hat{\bar{\chi}}}_{i,S}} + \mathcal{O} \left( 1 \right) \right)^2 {{  \hat{\bar{\sigma}}}^2_{i,s-1} \left( {\bf x}_{{  \hat{\underline{t}}}_s+i} \right)} +  
  \mathcal{O} \left( 1 \right) \\
  \mathop{\sum}\limits_{s \in \mathbb{N}_S+1} {\breve{\bar{r}}}_{j,s\uparrow}^2 
  &\!\!\!\leq 
  8 \mathop{\sum}\limits_{s \in \mathbb{N}_S+1} \left( \frac{1+\sqrt{{\breve{\bar{\zeta}}}_{j\sim}}}{2} \sqrt{{\breve{\bar{\chi}}}_{j,S}} + \mathcal{O} \left( 1 \right) \right)^2 {{\breve{\bar{\sigma}}}^2_{j,s-1} \left( {\bf x}_{{\breve{\underline{t}}}_s+j} \right)} +  
  \mathcal{O} \left( 1 \right) \\
 \end{array}
\]
Now, by the standard procedure:
\[
 \begin{array}{rl}
  \mathop{\sum}\limits_{s \in \mathbb{N}_S+1} {  \hat{\bar{\sigma}}}^{-2} {{  \hat{\bar{\sigma}}}_{i,s-1}^2 \left( {\bf x}_{\underline{t}_s+i} \right)} 
  &\!\!\!\leq 
  {  \hat{C}}_{i,2} 
  \mathop{\sum}\limits_{s \in \mathbb{N}_S+1} \ln \left( 1 + {  \hat{\bar{\sigma}}}^{-2} {{  \hat{\bar{\sigma}}}_{i,s-1}^2 \left( {\bf x}_{\underline{t}_s+i} \right)} \right) \\
  &\!\!\!\leq 
  {  \hat{C}}_{i,2} 
  \mathop{\sum}\limits_{t \in \mathbb{T}_{\leq S}} \ln \left( 1 + {  \hat{\bar{\sigma}}}^{-2} {{  \hat{\bar{\sigma}}}_{i,s_t-1}^2 \left( {\bf x}_t \right)} \right) \\
  &\!\!\!\leq 
  {  \hat{C}}_{i,2} {  \hat{\bar{\gamma}}}_{i,S} \\

  \mathop{\sum}\limits_{s \in \mathbb{N}_S+1} {\breve{\bar{\sigma}}}^{-2} {{\breve{\bar{\sigma}}}_{j,s-1}^2 \left( {\bf x}_{\underline{t}_s+j} \right)} 
  &\!\!\!\leq 
  {\breve{C}}_{j,2} 
  \mathop{\sum}\limits_{s \in \mathbb{N}_S+1} \ln \left( 1 + {\breve{\bar{\sigma}}}^{-2} {{\breve{\bar{\sigma}}}_{j,s-1}^2 \left( {\bf x}_{\underline{t}_s+j} \right)} \right) \\
  &\!\!\!\leq 
  {\breve{C}}_{j,2} 
  \mathop{\sum}\limits_{t \in \mathbb{T}_{\leq S}} \ln \left( 1 + {\breve{\bar{\sigma}}}^{-2} {{\breve{\bar{\sigma}}}_{j,s_t-1}^2 \left( {\bf x}_t \right)} \right) \\
  &\!\!\!\leq 
  {\breve{C}}_{j,2} {\breve{\bar{\gamma}}}_{j,S} \\
 \end{array}
\]
where:
\[
 \begin{array}{l}
  {  \hat{C}}_{i,2} = \mathop{\max}\limits_{s \in \mathbb{N}_S+1} \frac{{  \hat{\bar{\sigma}}}^{-2} {  \hat{\bar{K}}}_{i,s\uparrow}}{\ln \left( 1 + {  \hat{\bar{\sigma}}}^{-2} {  \hat{\bar{K}}}_{i,s\uparrow} \right)} \\
  {\breve{C}}_{j,2} = \mathop{\max}\limits_{s \in \mathbb{N}_S+1} \frac{{\breve{\bar{\sigma}}}^{-2} {\breve{\bar{K}}}_{j,s\uparrow}}{\ln \left( 1 + {\breve{\bar{\sigma}}}^{-2} {\breve{\bar{K}}}_{j,s\uparrow} \right)} \\
 \end{array}
\]
and so:
\[
 \begin{array}{rl}
  \mathop{\sum}\limits_{s \in \mathbb{N}_S+1} {  \hat{\bar{r}}}_{i,s\uparrow}^2 
  &\!\!\!\leq 
  8 {  \hat{\bar{\sigma}}}^2 \left( \sqrt{{  \hat{\bar{\chi}}}_{i,S}} + \mathcal{O} \left( 1 \right) \right)^2 {  \hat{C}}_{i,2} {  \hat{\bar{\gamma}}}_{i,S} +  
  \mathcal{O} \left( 1 \right) \\
  \mathop{\sum}\limits_{s \in \mathbb{N}_S+1} {\breve{\bar{r}}}_{j,s\uparrow}^2 
  &\!\!\!\leq 
  8 {\breve{\bar{\sigma}}}^2 \left( \frac{1+\sqrt{{\breve{\bar{\zeta}}}_{j\sim}}}{2} \sqrt{{\breve{\bar{\chi}}}_{j,S}} + \mathcal{O} \left( 1 \right) \right)^2 {\breve{C}}_{j,2} {\breve{\bar{\gamma}}}_{j,S} +  
  \mathcal{O} \left( 1 \right) \\
 \end{array}
\]
Recalling that $\| \cdot \|_1 \leq \sqrt{S} \| \cdot \|_2$ (in $S$ dimensions):
\[
 \begin{array}{rl}
  \mathop{\sum}\limits_{s \in \mathbb{N}_S+1} {  \hat{\bar{r}}}_{i,s\uparrow} 
  &\!\!\!\leq 
  \sqrt{S} \sqrt{ 8 {  \hat{\bar{\sigma}}}^2 \left( \sqrt{{  \hat{\bar{\chi}}}_{i,S}} + \mathcal{O} \left( 1 \right) \right)^2 {  \hat{C}}_{i,2} {  \hat{\bar{\gamma}}}_{i,S} +  
  \mathcal{O} \left( 1 \right) } \\
  \mathop{\sum}\limits_{s \in \mathbb{N}_S+1} {\breve{\bar{r}}}_{j,s\uparrow} 
  &\!\!\!\leq 
  \sqrt{S} \sqrt{ 18 {\breve{\bar{\sigma}}}^2 \left( \frac{1+\sqrt{{\breve{\bar{\zeta}}}_{j\sim}}}{2} \sqrt{{\breve{\bar{\chi}}}_{j,S}} + \mathcal{O} \left( 1 \right) \right)^2 {\breve{C}}_{j,2} {\breve{\bar{\gamma}}}_{j,S} +  
  \mathcal{O} \left( 1 \right) } \\
 \end{array}
\]
and so:
\[
{\!\!\!\!\!\!\!\!\!\!\!\!\!\!\!\!\!\!\!\!\!\!\!\!\!\!\!\!\!\!{
 \begin{array}{l}
  \mathop{\sum}\limits_{s \in \mathbb{N}_S+1} {  \hat{\bar{r}}}_{i,s\uparrow} +
  \mathop{\sum}\limits_{s \in \mathbb{N}_S+1} {\breve{\bar{r}}}_{j,s\uparrow} 

  \ldots \\ 
  \;\;\;\leq
  
  \ln \Bigg( M_{-\frac{{\bar{\theta}}_\downarrow}{S}} \Bigg( \exp \Bigg( \Bigg\{
  
  \left\{ \left.
  \sqrt{S} \sqrt{ 8 {  \hat{\bar{\sigma}}}^2 \left( \mathcal{O} \left( 1 \right) + \sqrt{{  \hat{\bar{\chi}}}_{i,S}} \right)^2 {  \hat{C}}_{i,2} {  \hat{\bar{\gamma}}}_{i,S} +
  \mathcal{O} \left( 1 \right) } 
  \right| i \in {  \hat{\mathbb{M}}} \right\} \bigcup \ldots  \\

  \hfill \left\{ \left.
  \sqrt{S} \sqrt{ 8 {\breve{\bar{\sigma}}}^2 \left( \mathcal{O} \left( 1 \right) + \frac{1+\sqrt{{\breve{\bar{\zeta}}}_{j\sim}}}{2} \sqrt{{\breve{\bar{\chi}}}_{j,S}} \right)^2 {\breve{C}}_{j,2} {\breve{\bar{\gamma}}}_{j,S} +
  \mathcal{O} \left( 1 \right) } 
  \right| j \in {\breve{\mathbb{M}}} \right\}
  
  \Bigg\} \Bigg) \Bigg) \Bigg) \\
 \end{array}
 }}
\]
Finally, noting that, for ${\bf a} > {\bf 0}$:
\[
 \begin{array}{rl}
  \frac{1}{S} \ln \left( M_{-\frac{{\bar{\theta}}_\downarrow}{S}} \left( \exp \left( {\bf a} \right) \right) \right)
  &\!\!\!=
  \ln \left( M_{-\frac{{\bar{\theta}}_\downarrow}{S}}^{\frac{1}{S}} \left( \exp \left( {\bf a} \right) \right) \right) \\
  &\!\!\!=
  \ln \left( \left( \frac{1}{p} \sum_i \left( e^{a_i} \right)^{-\frac{{\bar{\theta}}_\downarrow}{S}} \right)^{-\frac{1}{{\bar{\theta}}_\downarrow}} \right) \\
  &\!\!\!=
  \ln \left( \left( \frac{1}{p} \sum_i \left( e^{\frac{1}{S} a_i} \right)^{-{\bar{\theta}}_\downarrow} \right)^{-\frac{1}{{\bar{\theta}}_\downarrow}} \right) \\
  &\!\!\!=
  \ln \left( M_{-{\bar{\theta}}_\downarrow} \left( \exp \left( \frac{1}{S} {\bf a} \right) \right) \right) \\
 \end{array}
\]
we obtain:
\[
{\!\!\!\!\!\!\!\!\!\!\!\!\!\!\!\!\!\!\!\!{
 \begin{array}{l}
  \frac{1}{S} \left(
  \mathop{\sum}\limits_{s \in \mathbb{N}_S+1} {  \hat{\bar{r}}}_{i,s\uparrow} +
  \mathop{\sum}\limits_{s \in \mathbb{N}_S+1} {\breve{\bar{r}}}_{j,s\uparrow} 
  \right)

  \ldots \\ 
  \;\;\;\leq 
  
  \ln \Bigg( M_{-{\bar{\theta}}_\downarrow} \Bigg( \exp \Bigg( \Bigg\{
  
  \left\{ \left.
  \frac{1}{\sqrt{S}} \sqrt{ 8 {  \hat{\bar{\sigma}}}^2 \left( \mathcal{O} \left( 1 \right) + \sqrt{{  \hat{\bar{\chi}}}_{i,S}} \right)^2 {  \hat{C}}_{i,2} {  \hat{\bar{\gamma}}}_{i,S} +
  \mathcal{O} \left( 1 \right) } 
  \right| i \in {  \hat{\mathbb{M}}} \right\} \bigcup \ldots  \\

  \hfill \left\{ \left.
  \frac{1}{\sqrt{S}} \sqrt{ 8 {\breve{\bar{\sigma}}}^2 \left( \mathcal{O} \left( 1 \right) + \frac{1+\sqrt{{\breve{\bar{\zeta}}}_{j\sim}}}{2} \sqrt{{\breve{\bar{\chi}}}_{j,S}} \right)^2 {\breve{C}}_{j,2} {\breve{\bar{\gamma}}}_{j,S} +
  \mathcal{O} \left( 1 \right) } 
  \right| j \in {\breve{\mathbb{M}}} \right\}
  
  \Bigg\} \Bigg) \Bigg) \Bigg) \\
 \end{array}
 }}
\]
as required.
\end{proof}

Finally, we consider the bounds on the posterior variance at ${\bf x}^\star$.  We 
have the following result:
\begin{lem_sigmalimit}
 Fix $j' \in {\breve{\mathbb{M}}}$.  We have the bounds for all $i \in 
 \hat{\mathbb{M}}$, $j \in \breve{\mathbb{M}}$:
 \[
  \begin{array}{rcl}
   {  \hat{\bar{\sigma}}}_{i,s\downarrow}^\star 
   &\!\!\!\leq
   {  \hat{\bar{\sigma}}}_{i,s} \left( {\bf x}^\star \right) 
   &\!\!\!\leq
   {  \hat{\bar{\sigma}}}_{i,s\uparrow}^\star \\

   {\breve{\bar{\sigma}}}_{j,s\downarrow}^\star 
   &\!\!\!\leq
   {\breve{\bar{\sigma}}}_{j,s} \left( {\bf x}^\star \right) 
   &\!\!\!\leq
   {\breve{\bar{\sigma}}}_{j,s\uparrow}^\star \\
  \end{array}
 \]
 where:
 \[
  \begin{array}{rl}
   {  \hat{\bar{\sigma}}}_{i,s\downarrow}^{\star 2} 
   &\!\!\!=
   \frac{{  \hat{\bar{\sigma}}}^2 + 2 ps {  \hat{\bar{D}}}_{i,s,\downarrow} - ps \frac{1}{{  \hat{\bar{K}}}_{i,s}^{\star\star}} {  \hat{\bar{D}}}_{i,s,\downarrow}^2}
        {{  \hat{\bar{\sigma}}}^2 + ps {  \hat{\bar{K}}}_{i,s}^{\star\star}} 
   {  \hat{\bar{K}}}_{i,s}^{\star\star} \\

   {\breve{\bar{\sigma}}}_{j,s\downarrow}^{\star 2}
   &\!\!\!=
   \frac{{\breve{\bar{\sigma}}}^2 + 2 ps {\breve{\bar{D}}}_{j,s,\downarrow} - ps \frac{1}{{\breve{\bar{K}}}_{j,s}^{\star\star}} {\breve{\bar{D}}}_{j,s,\downarrow}^2}
        {{\breve{\bar{\sigma}}}^2 + ps {\breve{\bar{K}}}_{j,s}^{\star\star}} 
   {\breve{\bar{K}}}_{j,s}^{\star\star} \\

   {  \hat{\bar{\sigma}}}_{i,s\uparrow}^{\star 2}
   &\!\!\!=
   \frac{{  \hat{\bar{\sigma}}}^2 + 2 \sum_{t \in \mathbb{T}_{\leq s}} {  \hat{\bar{D}}}_{i,s,t} - \frac{1}{{  \hat{\bar{K}}}_{i,s}^{\star\star}} \sum_{t \in \mathbb{T}_{\leq s}} {  \hat{\bar{D}}}_{i,s,t}^2}
        {{  \hat{\bar{\sigma}}}^2 + ps {  \hat{\bar{K}}}_{i,s}^{\star\star}} 
   {  \hat{\bar{K}}}_{i,s}^{\star\star} \\
\end{array}
 \]
 and:
 \[
  \begin{array}{rlrl}
   {  \hat{\bar{K}}}_{i,s}^{\star\star} &\!\!\!\!= {  \hat{\bar{K}}}_{i,s} \left( {\bf x}^\star, {\bf x}^\star \right), &
   {\breve{\bar{K}}}_{j,s}^{\star\star} &\!\!\!\!= {\breve{\bar{K}}}_{j,s} \left( {\bf x}^\star, {\bf x}^\star \right) \\
   {  \hat{\bar{K}}}_{i,s,t,\star}      &\!\!\!\!= {  \hat{\bar{K}}}_{i,s} \left( {\bf x}^\star, {\bf x}_t     \right), &
   {\breve{\bar{K}}}_{j,s,t,\star}      &\!\!\!\!= {\breve{\bar{K}}}_{j,s} \left( {\bf x}^\star, {\bf x}_t     \right) \\ 
   {  \hat{\bar{K}}}_{i,s,t,t}          &\!\!\!\!= {  \hat{\bar{K}}}_{i,s} \left( {\bf x}_t,     {\bf x}_t     \right), &
   {\breve{\bar{K}}}_{j,s,t,t}          &\!\!\!\!= {\breve{\bar{K}}}_{j,s} \left( {\bf x}_t,     {\bf x}_t     \right) \\
  \end{array}
 \]
 and:
 \[
  \begin{array}{rlrl}
   \left| {  \hat{\bar{K}}}_{i,s,t,t} \right| &\!\!\!\!= {  \hat{\bar{K}}}_{i,s}^{\star\star} - {  \hat{\bar{D}}}_{i,s,t}, &
   \left| {\breve{\bar{K}}}_{j,s,t,t} \right| &\!\!\!\!= {\breve{\bar{K}}}_{j,s}^{\star\star} - {\breve{\bar{D}}}_{i,s,t} \\
   {  \hat{\bar{D}}}_{i,s,\downarrow} &\!\!\!\!= \mathop{\min}\limits_{t \in \mathbb{T}_{\leq s}} \left\{ {  \hat{\bar{D}}}_{i,s,t} \right\}, &
   {\breve{\bar{D}}}_{i,s,\downarrow} &\!\!\!\!= \mathop{\min}\limits_{t \in \mathbb{T}_{\leq s}} \left\{ {\breve{\bar{D}}}_{i,s,t} \right\} \\
  \end{array}
 \]
 \label{lem:sigmalimit}
\end{lem_sigmalimit}
\begin{proof}
Using the definition of GP posterior variance and $K$ bounds we can minimize 
the posterior variance within the constraints given as:
\[
{{
 \begin{array}{rl}
  {  \hat{\bar{\sigma}}}_{i,s}^2 \left( {\bf x}^\star \right) 

  &\!\!\!\geq
  {  \hat{\bar{K}}}_{i,s}^{\star\star} - {  \hat{\bar{K}}}_{i,s,\uparrow,\star}^2 {\bf 1}_{ps}^{\rm T} \left( {  \hat{\bar{K}}}_{i,s\uparrow} {\bf 1}_{ps} {\bf 1}_{ps}^{\rm T} + {  \hat{\bar{\sigma}}}^2 {\bf I}_{ps} \right)^{-1} {\bf 1}_{ps} \\

  &\!\!\!=
  {  \hat{\bar{K}}}_{i,s}^{\star\star} \left( 1 - \frac{{  \hat{\bar{K}}}_{i,s,\uparrow,\star}^2}{{  \hat{\bar{K}}}_{i,s}^{\star\star}} \frac{ps}{{  \hat{\bar{\sigma}}}^2 + ps {  \hat{\bar{K}}}_{i,s\uparrow}} \right) \\

  &\!\!\!=
  {  \hat{\bar{K}}}_{i,s}^{\star\star} \left( 1 - \frac{ps {  \hat{\bar{K}}}_{i,s,\uparrow,\star}^2}{{  \hat{\bar{\sigma}}}^2 {  \hat{\bar{K}}}_{i,s}^{\star\star} + ps {  \hat{\bar{K}}}_{i,s\uparrow} {  \hat{\bar{K}}}_{i,s}^{\star\star}} \right) \\

  &\!\!\!=
  {  \hat{\bar{K}}}_{i,s}^{\star\star} \left( \frac{{  \hat{\bar{\sigma}}}^2 {  \hat{\bar{K}}}_{i,s}^{\star\star} + ps {  \hat{\bar{K}}}_{i,s\uparrow} {  \hat{\bar{K}}}_{i,s}^{\star\star} - ps {  \hat{\bar{K}}}_{i,s,\uparrow,\star}^2}{{  \hat{\bar{\sigma}}}^2 {  \hat{\bar{K}}}_{i,s}^{\star\star} + ps {  \hat{\bar{K}}}_{i,s\uparrow} {  \hat{\bar{K}}}_{i,s}^{\star\star}} \right) \\

  &\!\!\!=
  {  \hat{\bar{K}}}_{i,s}^{\star\star} \left( \frac{{  \hat{\bar{\sigma}}}^2 {  \hat{\bar{K}}}_{i,s}^{\star\star} + ps {  \hat{\bar{K}}}_{i,s\uparrow} {  \hat{\bar{K}}}_{i,s}^{\star\star}  - ps {  \hat{\bar{K}}}_{i,s}^{\star\star 2} + 2 ps {  \hat{\bar{K}}}_{i,s}^{\star\star} D_{i,s,\downarrow} - ps D_{i,s,\downarrow}^2}
                                                   {{  \hat{\bar{\sigma}}}^2 {  \hat{\bar{K}}}_{i,s}^{\star\star} + ps {  \hat{\bar{K}}}_{i,s\uparrow} {  \hat{\bar{K}}}_{i,s}^{\star\star}} \right) \\
\end{array}
}}
\]
and likewise for the AI posterior variances.  
For the upper bound, we may pessimise the bound by first using the maximum 
eigenvalue:
\[
{{
 \begin{array}{rl}
  {  \hat{\bar{\sigma}}}_{i,s}^2 \left( {\bf x}^\star \right) 

  &\!\!\!\leq
  {  \hat{\bar{K}}}_{i,s}^{\star\star}
  - 
  \frac{\sum_{t \in \mathbb{T}_{\leq s}} {  \hat{\bar{K}}}_{i,s,t,\star}^2}
       {\lambda_{\rm max} \left( {  \hat{\bar{\bf K}}}_{i,s} \right) + {  \hat{\bar{\sigma}}}^2}
 \end{array}
}}
\]
and by Gershgorin's circle theorem $\lambda_{\rm max} ({\hat{\bar{\bf K}}}_{i,s}) 
\leq \sum_{t \in \mathbb{T}_{\leq s}} {  \hat{\bar{K}}}_{i,s,t_s,t_s}$, so:
\[
{{
 \begin{array}{rl}
  {  \hat{\bar{\sigma}}}_{i,s}^2 \left( {\bf x}^\star \right) 

  &\!\!\!\leq
  {  \hat{\bar{K}}}_{i,s}^{\star\star}
  - 
  \frac{
        \sum_{t \in \mathbb{T}_{\leq s}} 
        {  \hat{\bar{K}}}_{i,s,t,\star}^2
       }{
        {  \hat{\bar{\sigma}}}^2
        +
        \sum_{t \in \mathbb{T}_{\leq s}} 
        {  \hat{\bar{K}}}_{i,s,t_s,t_s}
       } \\

  &\!\!\!=
  \frac{
        {  \hat{\bar{\sigma}}}^2
        {  \hat{\bar{K}}}_{i,s}^{\star\star}
        +
        \sum_{t \in \mathbb{T}_{\leq s}} 
        {  \hat{\bar{K}}}_{i,s,t_s,t_s} 
        {  \hat{\bar{K}}}_{i,s}^{\star\star}
        -
        \sum_{t \in \mathbb{T}_{\leq s}} 
        {  \hat{\bar{K}}}_{i,s,t,\star}^2
       }{
        {  \hat{\bar{\sigma}}}^2
        {  \hat{\bar{K}}}_{i,s}^{\star\star}
        +
        \sum_{t \in \mathbb{T}_{\leq s}} 
        {  \hat{\bar{K}}}_{i,s,t_s,t_s}
        {  \hat{\bar{K}}}_{i,s}^{\star\star}
       } 
       {  \hat{\bar{K}}}_{i,s}^{\star\star} \\
 \end{array}
}}
\]
Now, we have that ${  \hat{\bar{K}}}_{i,s,t,\star} = {  \hat{\bar{K}}}_{i,s}^{\star\star} 
- {  \hat{\bar{D}}}_{i,s,t}$, where ${  \hat{\bar{D}}}_{i,s,t} \in [0, 
{  \hat{\bar{K}}}_{i,s}^{\star\star}]$, so:
\[
{{
 \begin{array}{rl}
  {  \hat{\bar{\sigma}}}_{i,s}^2 \left( {\bf x}^\star \right) 

  &\!\!\!\leq
  \frac{
        {  \hat{\bar{\sigma}}}^2
        {  \hat{\bar{K}}}_{i,s}^{\star\star}
        +
        \sum_{t \in \mathbb{T}_{\leq s}} 
        {  \hat{\bar{K}}}_{i,s,t_s,t_s} 
        {  \hat{\bar{K}}}_{i,s}^{\star\star}
        -
        ps 
        {  \hat{\bar{K}}}_{i,s}^{\star\star 2}
        +
        2
        {  \hat{\bar{K}}}_{i,s}^{\star\star}
        \sum_{t \in \mathbb{T}_{\leq s}} 
        {  \hat{\bar{D}}}_{i,s,t}
        -
        \sum_{t \in \mathbb{T}_{\leq s}} 
        {  \hat{\bar{D}}}_{i,s,t}^2
       }{
        {  \hat{\bar{\sigma}}}^2
        {  \hat{\bar{K}}}_{i,s}^{\star\star}
        +
        \sum_{t \in \mathbb{T}_{\leq s}} 
        {  \hat{\bar{K}}}_{i,s,t_s,t_s}
        {  \hat{\bar{K}}}_{i,s}^{\star\star}
       } 
       {  \hat{\bar{K}}}_{i,s}^{\star\star} \\

  &\!\!\!\leq
  \frac{
        {  \hat{\bar{\sigma}}}^2
        {  \hat{\bar{K}}}_{i,s}^{\star\star}
        +
        \sum_{t \in \mathbb{T}_{\leq s}} 
        {  \hat{\bar{K}}}_{i,s,t_s,t_s} 
        {  \hat{\bar{K}}}_{i,s}^{\star\star}
        -
        ps 
        {  \hat{\bar{K}}}_{i,s}^{\star\star 2}
        +
        2
        {  \hat{\bar{K}}}_{i,s}^{\star\star}
        \sum_{t \in \mathbb{T}_{\leq s}} 
        {  \hat{\bar{D}}}_{i,s,t}
        -
        \sum_{t \in \mathbb{T}_{\leq s}} 
        {  \hat{\bar{D}}}_{i,s,t}^2
       }{
        {  \hat{\bar{\sigma}}}^2
        {  \hat{\bar{K}}}_{i,s}^{\star\star}
        +
        \sum_{t \in \mathbb{T}_{\leq s}} 
        {  \hat{\bar{K}}}_{i,s,t_s,t_s}
        {  \hat{\bar{K}}}_{i,s}^{\star\star}
       } 
       {  \hat{\bar{K}}}_{i,s}^{\star\star} \\
\end{array}
}}
\]
and likewise for AI variances.
\end{proof}

In this theorem the sequence ${  \hat{\bar{D}}}_{i,s,t} = {  \hat{\bar{K}}}_{i,s} 
({\bf x}^\star,{\bf x}^\star) - |{  \hat{\bar{K}}}_{i,s} ({\bf x}^\star, {\bf x}_t)| 
\geq 0$ is a proxy for convergence, being minimized for ${\bf x}_t = {\bf x}^\star$ 
and increasing as ${\bf x}_t$ becomes (a-posterior) less correlated with ${\bf 
x}^\star$.  If we consider the case considered in the paper - namely $1$ human and 
$1$ AI with zero gap and a trade-off sequence meeting the conditions of GP-UCB - 
then we know that the AI, operating alone, suffices to ensure convergence; and 
moreover adding additional observations (the human recommendations) will not prevent 
this.  From here, it is not difficult to see that the upper and lower bounds in the 
above theorem converge not just to $0$ but to one another, and if we further assume 
that ${  \hat{\bar{K}}}_{i,s}^{\star\star} = {\breve{\bar{K}}}_{j,s}^{\star\star}$ 
then the ratio of any upper bound on the posterior variance to the lower bound on 
any posterior variance will converge to $1$.


\subsection{Ablation Studies\label{subsec:AblationStudy}}
\subsubsection{Varying degrees of Exploitation-Exploration}\label{subsec:ablation_beta}

\begin{figure}
\begin{centering}
	\subfloat[]{\includegraphics[width=0.44\columnwidth]{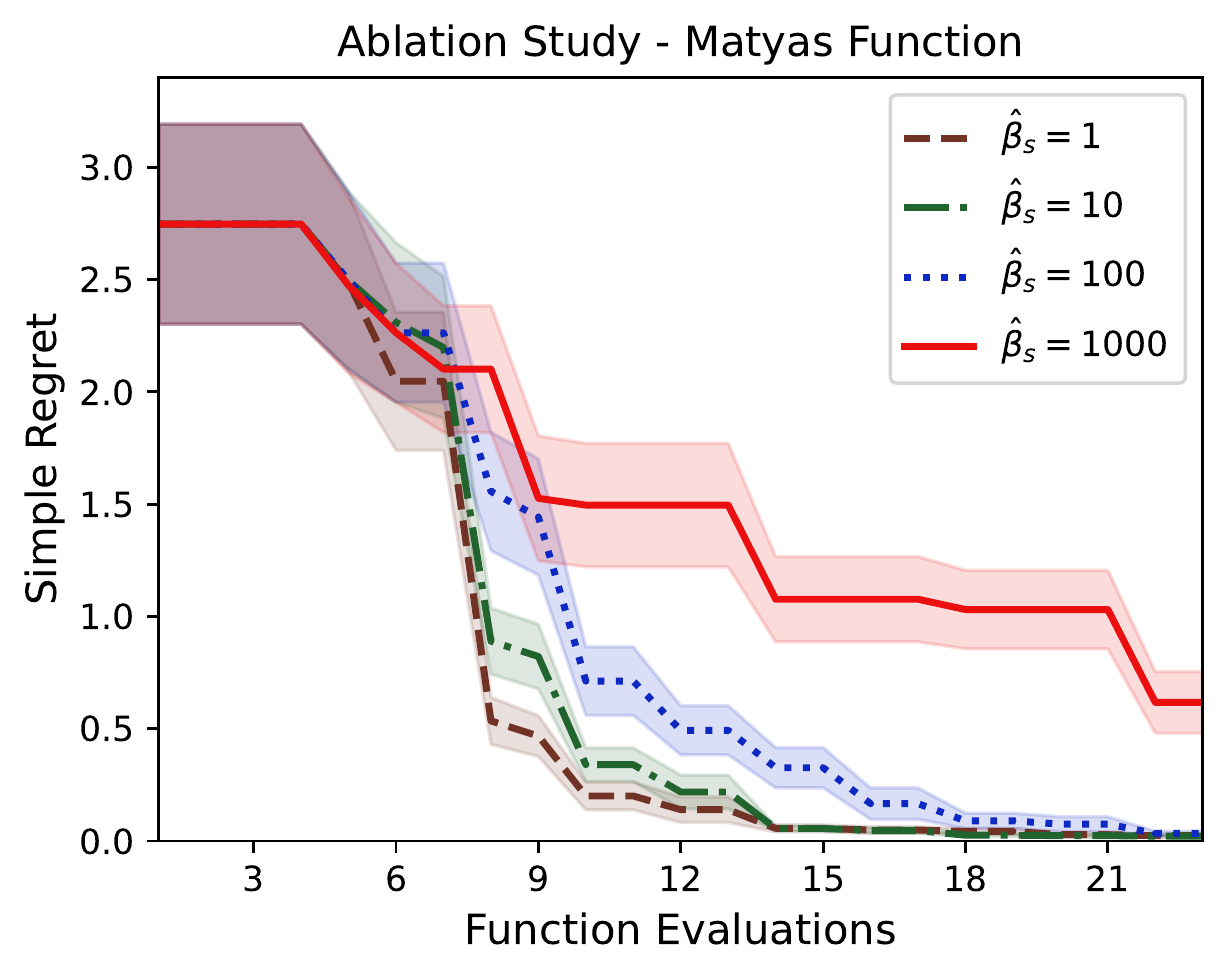}\label{fig:ablation-matyas}
		
	}\subfloat[]{\includegraphics[width=0.44\columnwidth]{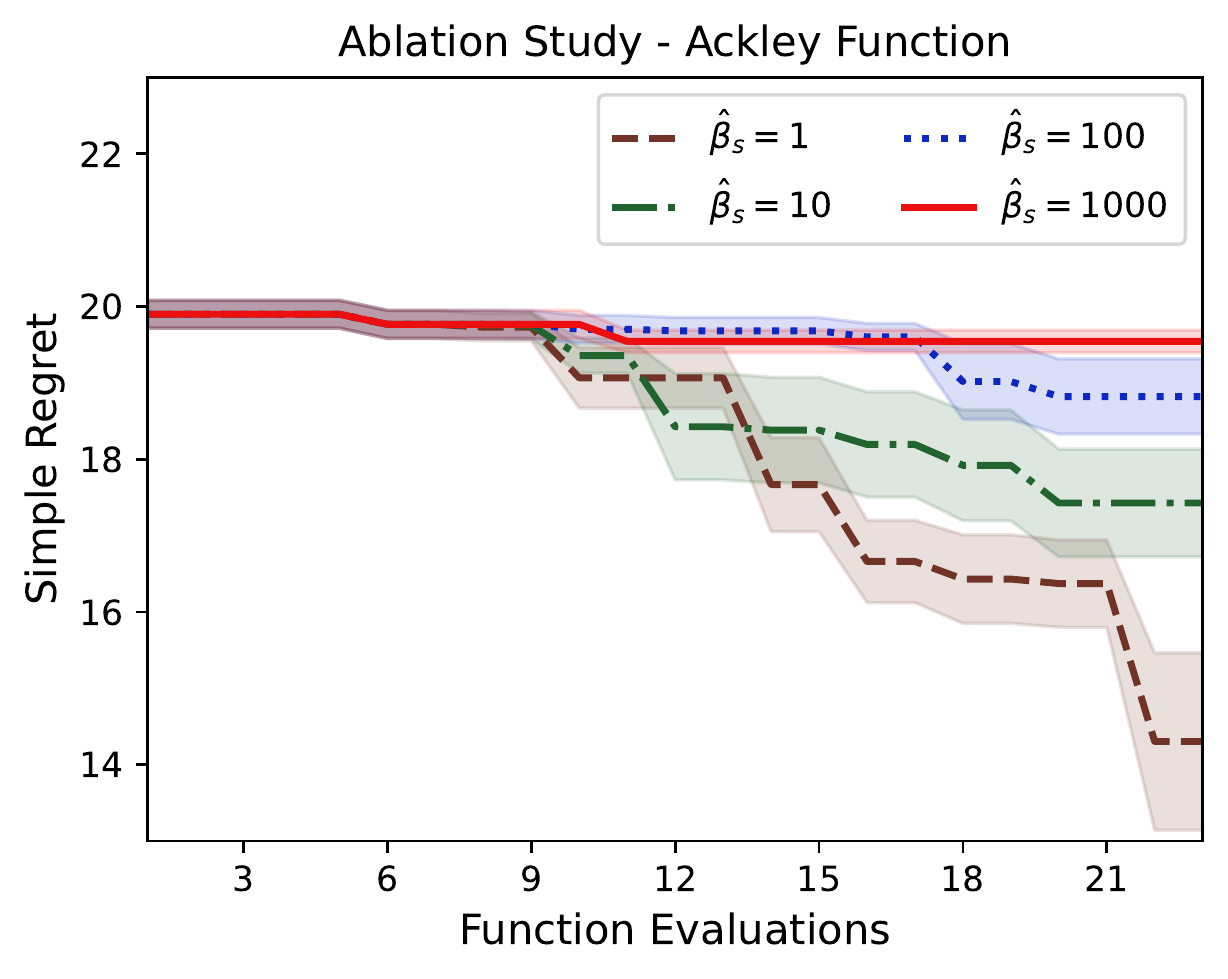}\label{fig:ablation-ack}}

\subfloat[]{\includegraphics[width=0.44\columnwidth]{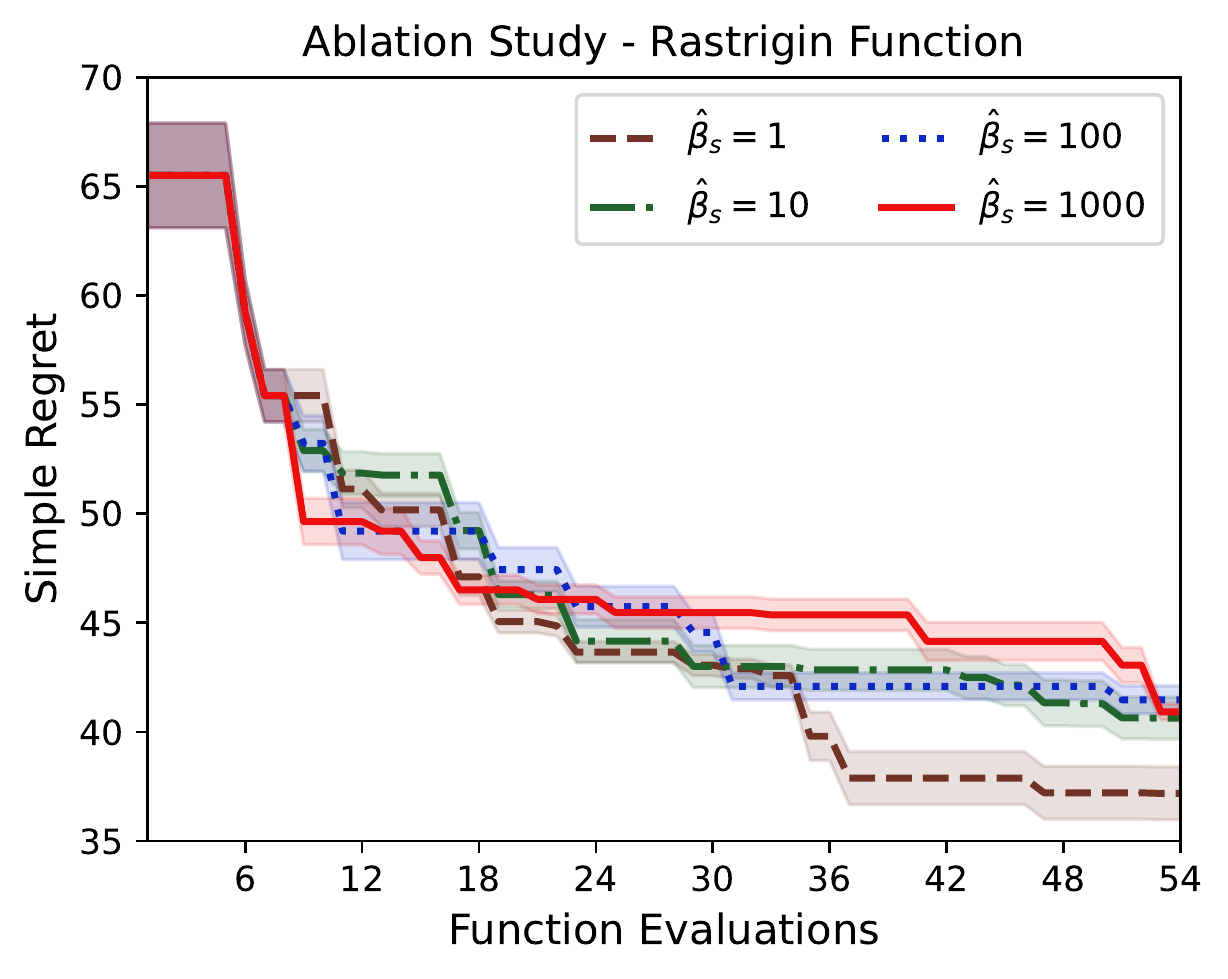}\label{fig:ablation-levy}}	
\subfloat[]{\includegraphics[width=0.44\columnwidth]{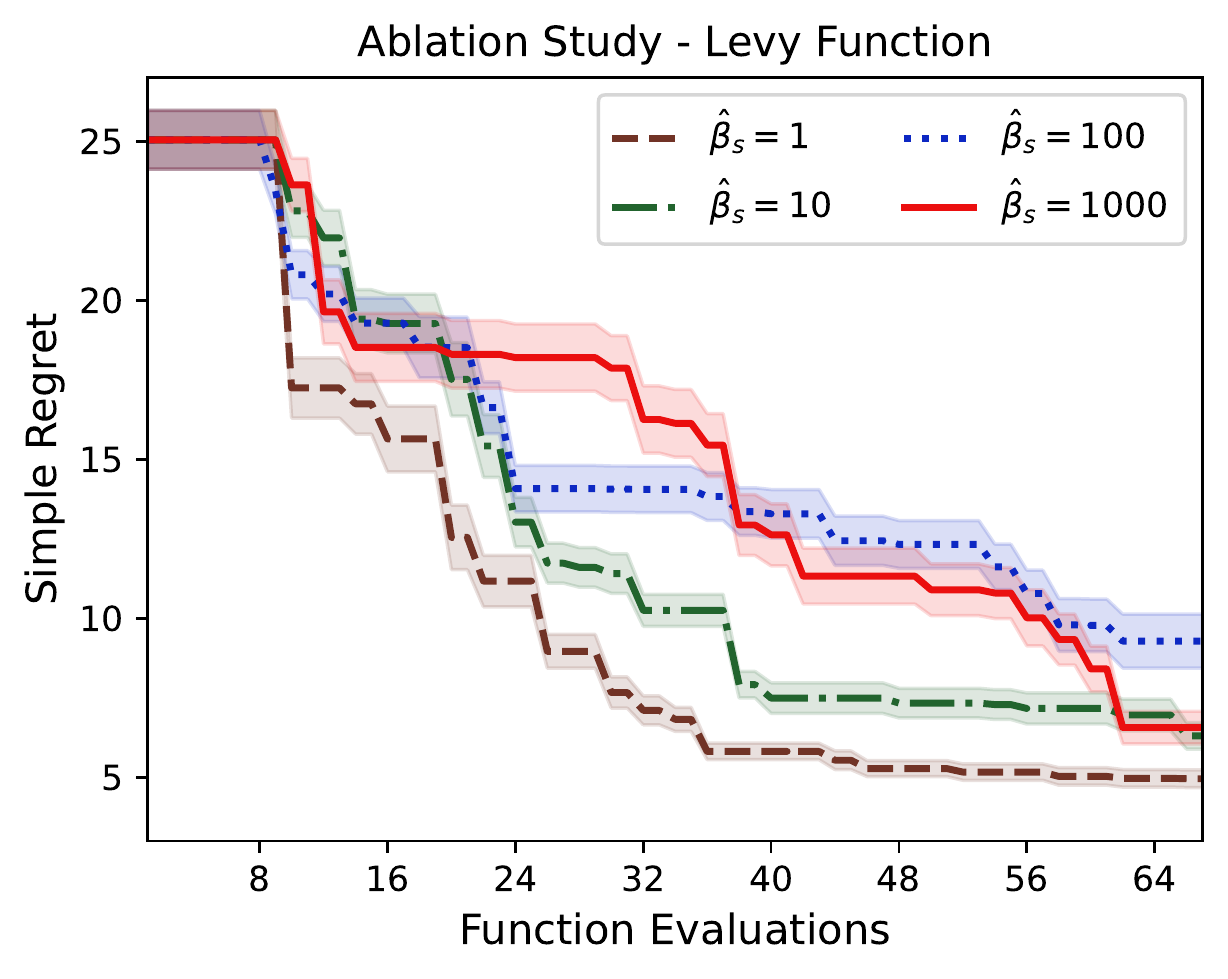}\label{fig:ablation-rastrigin}}
		
	\caption{Ablation study with varying degrees of exploration-exploitation parameter
	($\hat{\beta}_{s}$) obtained for (a) Matyas-2D, (b) Ackley-4D, (c) Rastrigin-5D, and (d) Levy-6D functions.\label{fig:ablationstudy-1}}
	
\end{centering}
\end{figure}

We study the sensitivity of the exploitation-exploration parameter ($\hat{\beta}_{s}$) in our proposed BO-Muse framework and compare the optimization performance. We vary the   exploitation-exploration parameter ($\hat{\beta}_{s}$) in the exponent range of $[0,3]$ i.e., $\hat{\beta}_{s} \in 10^{[0,3]}$ to cover the whole spectrum from over-exploitative experts ($\hat{\beta}_{s}=1$) to over-explorative experts ($\hat{\beta}_{s}=1000$). We have tuned the Squared Exponential (SE) kernel hyper-parameters of the inherent GP surrogate models using maximum-likelihood estimation. The empirical results obtained for various synthetic functions are depicted in Figure \ref{fig:ablationstudy-1}. It is evident from the empirical results that BO-Muse teamed up with an expert following more of an exploitation strategy ($\hat{\beta}_{s}=1$) has better convergence when compared to its counterpart teamed with pure explorative expert ($\hat{\beta}_{s}=1000$).

\subsubsection{BO-Muse with Expected Improvement Acquisition Function}\label{subsec:ablation_ei}

We have conducted an additional experiment to study the behavior of our BO-Muse framework with different acquisition function strategies for the human experts. Expected Improvement (EI) acquisition function \citep{wilson2018maximizing} guides the search for optima 
by taking into account the expected improvement over the current best solution.
If $f^\star(\mathbf{x}^{+})$ is the best value observed, then the next
best query point is obtained by maximizing the EI acquisition function
$\hat{a}^{\text{EI}}_{s}({\bf x})$, given by:

\begin{equation*}
	\hat{a}^{\text{EI}}_{s}({\bf x})=\left\{ \begin{aligned}(\hat{\mu}_{s}({\bf x})-f^\star(\mathbf{x}^{+}))\;\Phi(\mathcal{Z})+\hat{\sigma}_{s}({\bf x})\;\phi(\mathcal{Z}) \quad \text{if}\;\hat{\sigma}_{s}({\bf x})\!>\!0\\
		0\qquad\qquad\qquad\qquad\qquad\qquad\quad\quad\text{if}\;\hat{\sigma}_{s}({\bf x})\!=\!0
	\end{aligned}
	\right.
	\label{eq:acq-ei}
\end{equation*}
\[
\mathcal{Z}=\frac{\hat{\mu}_{s}({\bf x})-f^\star(\mathbf{x}^{+})}{\hat{\sigma}_{s}({\bf x})}
\]

where $\Phi(\mathcal{Z})$ and $\phi(\mathcal{Z})$ represents the Cumulative Distribution Function (CDF) and the Probability Density Function (PDF) of the
standard normal distribution, respectively. 

\begin{figure}
	
	\centering{}\includegraphics[width=0.6\columnwidth]{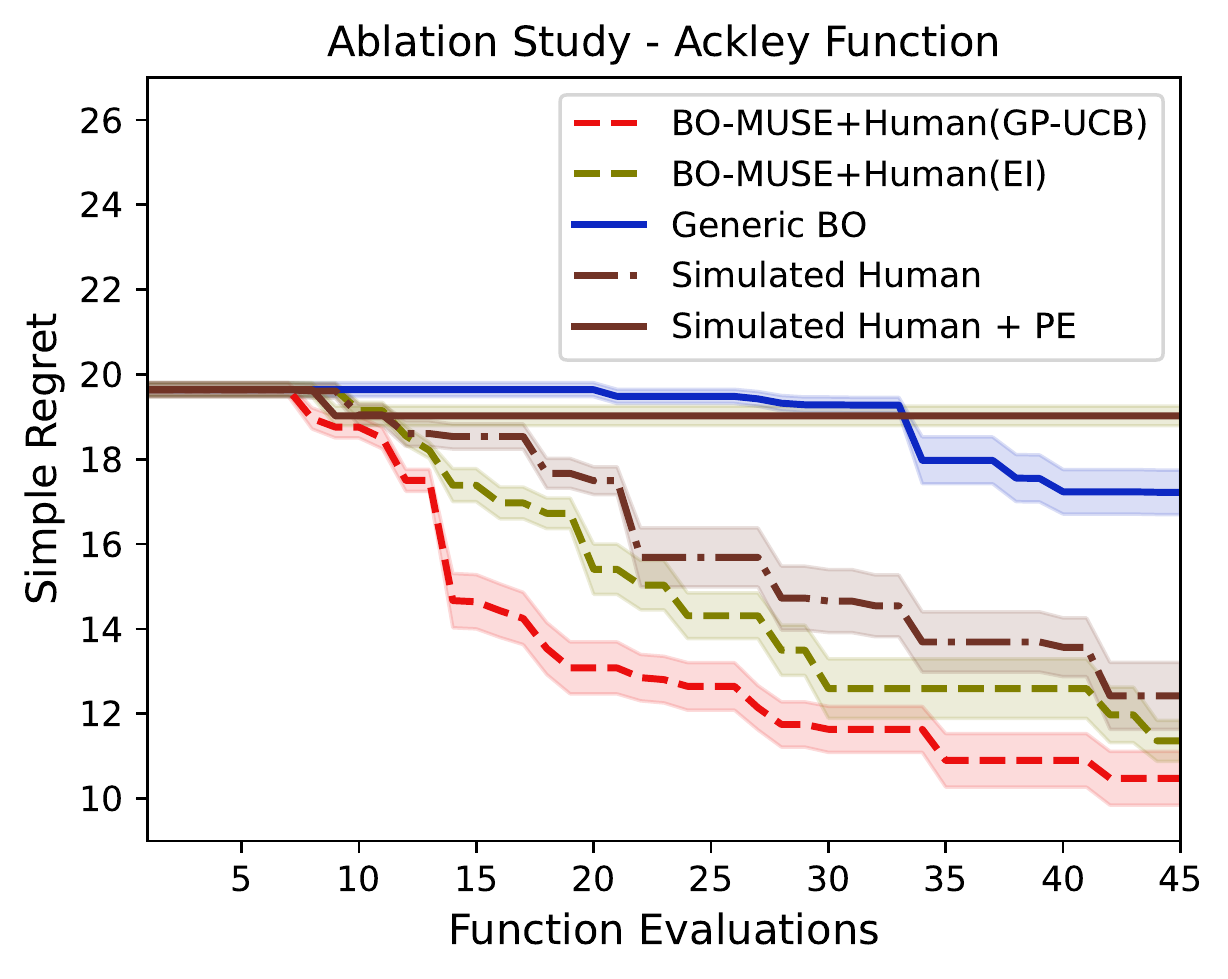}\label{fig:ack-ei}\caption{Results of the ablation study with EI acquisition function obtained for Ackley function.\label{fig:ablation-ei}}
	
\end{figure}

In this experiment, the GP-UCB acquisition function used by the human expert in BO-Muse framework is now replaced with the Expected Improvement acquisition function. We compare this new baseline (BO-Muse + Human (EI)) with BO-Muse + Human (GP-UCB) and all the other competing baselines. The empirical results obtained for the experiment with EI acquisition function is depicted in Figure \ref{fig:ablation-ei}. As expected BO-Muse with the EI acquisition function still outperforms the standard baselines considered. However, BO-Muse with GP-UCB acquisition function has superior performance when compared to its counterpart with the EI acquisition function.

\subsection{Additional Details of Classification Experiments\label{subsec:AdditionalDetailsCLF}}
We have considered two real-world classification tasks using Support
Vector Machines (SVMs) and Random Forests (RFs). This experiment involves
hyper-parameter tuning of SVMs and RFs operating on the \textit{Biodeg}
dataset to classify biodegradable and non-biodegradable materials.
We used publicly available Biodeg dataset from the UCI data repository
\citep{Dua_2019}. Biodeg dataset consists of 1056 instances with
$41$ features. We randomly split the dataset into $80/20$ train/test
splits. Each time a hyper-parameter set (design) is chosen, the model
needs to retrained and evaluated on a held out set. The goal is to
reach to the hyper-parameter set that leads to a classification model
with the minimum test classification error. 

We have created two groups (arms) with $4$ members ($2$ students
and $2$ postdoctoral researchers) randomly allocated in each group.
Each expert in the first group teams up with AI as per BO-Muse, while
each expert in the second group (baseline) tunes the classifier completely
on their own. For each of the classification tasks, the two groups
use the same tuning budget ($3$ random initial designs + $30$ further
iterations. The aforementioned real-world task is suitable for our
case: (1) It is easier to find multiple human experts for this task
as AI graduate students and post-doctoral researchers have a good
understanding of classification (SVM and RF) models and understand
how its hyper-parameters generally influence the model fitting, (2)
This task is familiar to the machine learning community. 

%

\subsection{Spacecraft Shielding Design Experiment\label{subsec:SpacecraftShieldingDesign}}

\section*{}

Our third experiment is to team with an expert to design spacecraft
shields to protect from impact by orbital debris particles. 

\subsubsection{Experimental problem}

\begin{figure}
\centering
\includegraphics[width=0.8\textwidth]{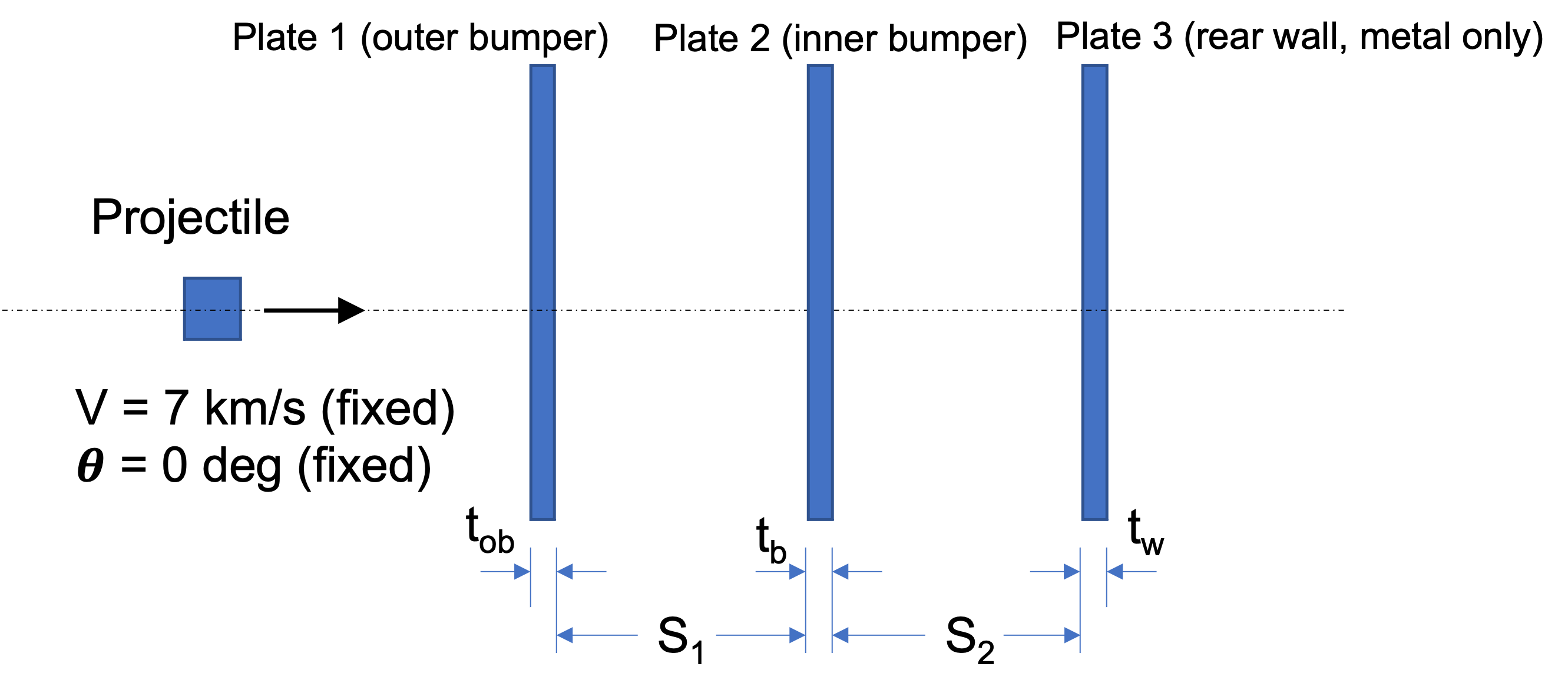}

{\caption{Schematic of the spec debris shield design problem. Our objective
is a design solution that will prevent perforation of the spacecraft
hull (Plate 3, rear wall) by modifying the plate materials, plate
thicknesses, and plate spacing (S).\label{fig:whipple-problem-schematic}
}
}
\end{figure}

Here we consider the design of a two- or three-wall shield for protection
against a cubic steel projectile impacting face on, normal to the
surface of the target plates, at an impact velocity of 7.0 km/s. There
exists no state-of-the-art solution for such an impact threat, however
for protecting against a spherical aluminium projectiles in this velocity
domain the state-of-the-art solution would be a ``stuffed Whipple
shield'' after \cite{christiansen1995}, consisting of an outer aluminium
plate, inner layers of aramid and ceramic fabrics, and a rear wall
(pressure hull) of aluminium. US, Japanese, and European modules on
the ISS all utilize stuffed Whipple shield designs \citet{MMODhandbook2009}.
The design space is schematically shown in Figure \ref{fig:whipple-problem-schematic}.
Design variables include: (1) plate material - AA6061-T651 (``AL''),
4340 steel (``ST''), Kevlar/epoxy (``KE''), and ultra-high molecular
weight polyethylene (``PE''); (2) plate thickness - 0.1 cm to 1.0
cm in 0.1 cm increments; (3) plate spacing, S - 0.0 cm to 10 cm in
1.0 cm increments, an; (4) number of plates, 2 or 3 (\textit{i.e.,
}the 'outer bumper' plate may or may not be used). Only metal plates
(i.e., ``AL'' or ``ST'') may be used for the 3rd plate. The full
factorial design space includes 577,365 options.

\subsubsection{Background}

Spacecraft are subject to impact by natural micrometeoroid and man-made
orbital debris particles, collectively referred to as space debris,
during their orbital lifetime. The impact of such particles (typically
at velocities above 10 km/s) is a significant risk to the safe operation
of spacecraft and the fulfillment of mission objectives. Indeed, for
manned spacecraft such as the International Space Station (ISS) and
Space Shuttle Orbiter, space debris impact is the top mission risk
(see e.g., \cite{hamlin2013}). As such, all manned spacecraft and
some robotic spacecraft carry dedicated protective shields. The most
common shield configuration is a simple design known as a Whipple
shield with two thin plates separated by a gap. Meteoroid and debris
particles, upon impact with the outer plate, fragment into a cloud
of solid, molten, and vaporized particles. This debris cloud expands
as it propagates through the gap, resulting in a dispersed and substantially
less lethal load upon the spacecraft hull. Multiple variants of the
Whipple shield exist, including stuffed Whipple shields which utilize
intermediate layers of high-strength and high-impedance fabric (see
e.g., \cite{christiansen1995}) and multi-shock shields which induce
multiple impact shocks into the projectile to promote maximum melting
and vaporization \citet{courpalais1990}.

Spacecraft debris shields are typically designed using a combination
of semi-analytical equations, numerical simulations, and experimental
testing. Simulations are performed in either explicit finite element
solvers, e.g., ANSYS LS-DYNA, or shock physics solvers, e.g., CTH
from Sandia National Laboratory. Modeling hypervelocity impact in
those simulation codes requires substantial expertise to accurately
projectile and target kinematics together with material response.
Furthermore, such simulations can be computationally expensive, requiring
hundreds of CPU hours depending on the geometric discretization of
the model. Experimentation is typically performed on laboratory accelerators
known as two-stage light gas guns. The number of such facilities that
can perform experiments with millimeter and centimeter sized projectiles
up to impact velocities of $7+$ km/s is very limited (estimated to be
$<20$ globally). Such experiments are also expensive, costing on the
order of thousands of dollars, with a low through-put of approximately
1 experiment per day. In the design of space debris shielding, in
order to minimize the number of experiments and simulations required,
space debris is typically simplified to spherical aluminium particles.
In reality, of course, the debris environment consists of a range
of materials, both metallic and non-metallic, the properties of which
influence their impact lethality. Similarly, for robotic and manned
spacecraft the majority of impact risk is represented by millimeter-sized
objects, the majority of which are fragmentation debris that have
been generated by catastrophic breakup of a satellite or rocket body
and are thus highly irregular in shape, see \cite{rivero2016}. Until
recently the engineering environment models used to predict mission
risk to space debris impact have also simplified the debris population
as spherical aluminium objects, thus there was little incentive to
introduce the added complexity of projectile shape and material effects
in shielding design or characterization studies. However, recent improvements
in orbital debris environment engineering models, e.g., ORDEM 3.0
\citet{krisko2014}, and planned improvements to debris population
source models, e.g., via DebriSat \citet{rivero2016}, aim to address
some of these deficiencies. Shield design and characterization, therefore,
must also begin to account for projectile shape and material effects. 

\subsubsection{Details of experiment}

The spacecraft shielding design experiment utilizes synthetic data
generated via numerical simulation. This section provides additional
information on the simulation setup and evaluation. Simulations are
performed in the explicit structural mechanics solver LS-DYNA from
ANSYS \citet{dynamanual}. Simulations are performed in 3D using a
smooth particle hydrodynamics (SPH) discretization scheme, which enables
projectile fragmentation to be modeled without arbitrary numerical
erosion that would otherwise be required for a mesh-based Lagrangian
scheme. SPH elements of 0.05 mm diameter are used to discretize all
simulated parts. The metallic materials, AA6061-T651 (``AL'') and
4340 steel (``ST''), utilize a Gruneisen equation of state (EoS)
\citet{gruneisen1959}, a Johnson-Cook viscoplasticity model \citet{johnsonCook1983}
and a Johnson-Cook fracture model \citet{johnsonCook1985}, the constants
for which are given in Tables \ref{tab:6061-model-constants} and
\ref{tab:4340-model-constants}. The aramid composite (``KE'') is
modeled as a continuum using the elastic-plastic orthotropic strength
with failure model from LS-DYNA (MAT\_059) and a linear EoS, the constants
for which are given in Table \ref{tab:kevlar-model-constants}. The
ultra-high molecular weight polyethylene (``PE''), specifically
Dyneema HB26, is modeled using the orthotropic non-linear model and
material constants from \citet{nguyen2016}. 

\begin{table}
\begin{centering}
	\caption{Constitutive model parameters for AA6061-T651 used in the numerical
		simulations.}
	\label{tab:6061-model-constants}
\begin{tabular}{|l|c|c|c|}
\hline 
Parameter & Value & Units & Source\tabularnewline
\hline 
\hline 
\textit{EoS: Gruneisen} &  &  & \tabularnewline
\hline 
Density, $\rho$ & 2700 & $kg/m^{3}$ & \citet{MIL-HDBK-5J}\tabularnewline
\hline 
Shear modulus, $G$ & 26.2 & $GPa$ & \citet{MIL-HDBK-5J}\tabularnewline
\hline 
Elastic modulus, $E$ & 68.3 & $GPa$ & \citet{MIL-HDBK-5J}\tabularnewline
\hline 
Poisson's ratio, $\nu$ & 0.33 & - & \citet{MIL-HDBK-5J}\tabularnewline
\hline 
Melting temperature, $T_{m}$ & 930 & K & \citet{corbett2006}\tabularnewline
\hline 
Bulk soundspeed, $c_{0}$ & 5240 & $m/s$ & \citet{corbett2006}\tabularnewline
\hline 
Parameter $S_{1}$ & 1.4 & - & \citet{corbett2006}\tabularnewline
\hline 
Parameter $S_{2}$ & 0.0 & - & \citet{corbett2006}\tabularnewline
\hline 
Parameter $S_{3}$ & 0.0 & - & \citet{corbett2006}\tabularnewline
\hline 
Gruneisen gamma, $\varGamma$ & 1.97 & - & \citet{corbett2006}\tabularnewline
\hline 
Specific heat, $c_{p}$ & 885 & $J/kgK$ & \citet{corbett2006}\tabularnewline
\hline 
 &  &  & \tabularnewline
\hline 
\textit{Strength: Johnson-Cook} &  &  & \tabularnewline
\hline 
Yield stress, $A$ & 324 & $MPa$ & \citet{anderson2006}\tabularnewline
\hline 
Hardening constant, $B$ & 114 & $MPa$ & \citet{anderson2006}\tabularnewline
\hline 
Hardening exponent, $n$ & 0.42 & - & \citet{anderson2006}\tabularnewline
\hline 
Strain rate constant, $C$ & 0.002 & - & \citet{anderson2006}\tabularnewline
\hline 
Thermal softening exponent, $m$ & 1.34 & - & \citet{anderson2006}\tabularnewline
\hline 
Reference strain rate, $\dot{\varepsilon_{0}}$ & 1.0 & $/s$ & \citet{anderson2006}\tabularnewline
\hline 
 &  &  & \tabularnewline
\hline 
\textsl{Failure: Johnson-Cook} &  &  & \tabularnewline
\hline 
Failure constant, $d_{1}$ & 0.0 & - & \citet{anderson2006}\tabularnewline
\hline 
Triaxiality constant, $d_{2}$ & 1.11 & - & \citet{anderson2006}\tabularnewline
\hline 
Triaxiality exponent, $d_{3}$ & -1.5 & - & \citet{anderson2006}\tabularnewline
\hline 
Strain rate constant, $d_{4}$ & 0.0 & - & \citet{anderson2006}\tabularnewline
\hline 
Thermal softening constant, $d_{5}$ & 0.0 & - & \citet{anderson2006}\tabularnewline
\hline 
\end{tabular}
\par\end{centering}
\end{table}

\begin{table}
\begin{centering}
	\caption{Constitutive model parameters for 4340 steel used in the numerical
		simulations.}
	\label{tab:4340-model-constants}
\begin{tabular}{|l|c|c|c|}
\hline 
Parameter & Value & Units & Source\tabularnewline
\hline 
\hline 
\textit{EoS: Gruneisen} &  &  & \tabularnewline
\hline 
Density, $\rho$ & 7850 & $kg/m^{3}$ & \citet{4340specification}\tabularnewline
\hline 
Shear modulus, $G$ & 80.0 & $GPa$ & \citet{4340specification}\tabularnewline
\hline 
Elastic modulus, $E$ & 205.0 & $GPa$ & \citet{4340specification}\tabularnewline
\hline 
Poisson's ratio, $\nu$ & 0.29 & - & \citet{4340specification}\tabularnewline
\hline 
Melting temperature, $T_{m}$ & 1700 & K & \citet{4340specification}\tabularnewline
\hline 
Bulk soundspeed, $c_{0}$ & 3935 & $m/s$ & \citet{banerjee2007}\tabularnewline
\hline 
Parameter $S_{1}$ & 1.578 & - & \citet{banerjee2007}\tabularnewline
\hline 
Parameter $S_{2}$ & 0.0 & - & \citet{banerjee2007}\tabularnewline
\hline 
Parameter $S_{3}$ & 0.0 & - & \citet{banerjee2007}\tabularnewline
\hline 
Gruneisen gamma, $\varGamma$ & 1.69 & - & \citet{banerjee2007}\tabularnewline
\hline 
Specific heat, $c_{p}$ & 475 & $J/kgK$ & \citet{4340specification}\tabularnewline
\hline 
 &  &  & \tabularnewline
\hline 
\textit{Strength: Johnson-Cook} &  &  & \tabularnewline
\hline 
Yield stress, $A$ & 910 & $MPa$ & \citet{holmquist2001}\tabularnewline
\hline 
Hardening constant, $B$ & 586 & $MPa$ & \citet{holmquist2001}\tabularnewline
\hline 
Hardening exponent, $n$ & 0.26 & - & \citet{holmquist2001}\tabularnewline
\hline 
Strain rate constant, $C$ & 0.014 & - & \citet{holmquist2001}\tabularnewline
\hline 
Thermal softening exponent, $m$ & 1.03 & - & \citet{holmquist2001}\tabularnewline
\hline 
Reference strain rate, $\dot{\varepsilon_{0}}$ & 1.0 & $/s$ & \citet{holmquist2001}\tabularnewline
\hline 
 &  &  & \tabularnewline
\hline 
\textsl{Failure: Johnson-Cook} &  &  & \tabularnewline
\hline 
Failure constant, $d_{1}$ & -0.80 & - & \citet{holmquist2001}\tabularnewline
\hline 
Triaxiality constant, $d_{2}$ & 2.10 & - & \citet{holmquist2001}\tabularnewline
\hline 
Triaxiality exponent, $d_{3}$ & -0.50 & - & \citet{holmquist2001}\tabularnewline
\hline 
Strain rate constant, $d_{4}$ & 0.002 & - & \citet{holmquist2001}\tabularnewline
\hline 
Thermal softening constant, $d_{5}$ & 0.61 & - & \citet{holmquist2001}\tabularnewline
\hline 
\end{tabular}
\par\end{centering}
\end{table}

\begin{table}
\begin{centering}
	\caption{Constitutive model parameters for woven aramid composite used in the
		numerical simulations.}
	\label{tab:kevlar-model-constants}
\begin{tabular}{|l|c|c|c|}
\hline 
Parameter & Value & Units & Source\tabularnewline
\hline 
\hline 
\textit{EoS: Linear} &  &  & \tabularnewline
\hline 
Density, $\rho$ & 1230 & $kg/m^{3}$ & \citet{vanHoof1999}\tabularnewline
\hline 
Shear modulus, $G_{11}$ & 0.77 & $GPa$ & \citet{vanHoof1999}\tabularnewline
\hline 
Shear modulus, $G_{22}$ & 1.36 & $GPa$ & \citet{vanHoof1999}\tabularnewline
\hline 
Shear modulus, $G_{33}$ & 1.36 & $GPa$ & \citet{vanHoof1999}\tabularnewline
\hline 
Elastic modulus, $E_{11}$ & 18.5 & $GPa$ & \citet{vanHoof1999}\tabularnewline
\hline 
Elastic modulus, $E_{22}$ & 18.5 & $GPa$ & \citet{vanHoof1999}\tabularnewline
\hline 
Elastic modulus, $E_{33}$ & 6.0 & $GPa$ & \citet{vanHoof1999}\tabularnewline
\hline 
Poisson's ratio, $\nu_{21}$ & 0.25 & - & \citet{vanHoof1999}\tabularnewline
\hline 
Poisson's ratio, $\nu_{31}$ & 0.33 & - & \citet{vanHoof1999}\tabularnewline
\hline 
Poisson's ratio, $\nu_{32}$ & 0.33 & - & \citet{vanHoof1999}\tabularnewline
\hline 
 &  &  & \tabularnewline
\hline 
\textit{Strength: Composite Failure} &  &  & \tabularnewline
\hline 
In-plane shear strength, $S_{21}$ & 9.0 & $MPa$ & \citet{vanHoof1999}\tabularnewline
\hline 
Transverse shear strength, $S_{31}$ & 271.5 & $MPa$ & \citet{vanHoof1999}\tabularnewline
\hline 
Transverse shear strength, $S_{32}$ & 271.5 & $MPa$ & \citet{vanHoof1999}\tabularnewline
\hline 
Longitudinal compressive strength, $C_{11}$ & 221.0 & $MPa$ & \citet{silcock2006}\tabularnewline
\hline 
Transverse compressive strength, $C_{22}$ & 221.0 & $MPa$ & \citet{silcock2006}\tabularnewline
\hline 
Normal compressive strength, $C_{33}$ & 1200.0 & $MPa$ & \citet{vanHoof1999}\tabularnewline
\hline 
Longitudinal tensile strength, $T_{11}$ & 740.0 & $MPa$ & \citet{vanHoof1999}\tabularnewline
\hline 
Transverse tensile strength, $T_{22}$ & 740.0 & $MPa$ & \citet{vanHoof1999}\tabularnewline
\hline 
Normal tensile strength, $T_{33}$ & 34.5 & $MPa$ & \citet{vanHoof1999}\tabularnewline
\hline 
\end{tabular}
\par\end{centering}
\end{table}

\begin{figure}[t]
\begin{centering}
\includegraphics[scale=0.75]{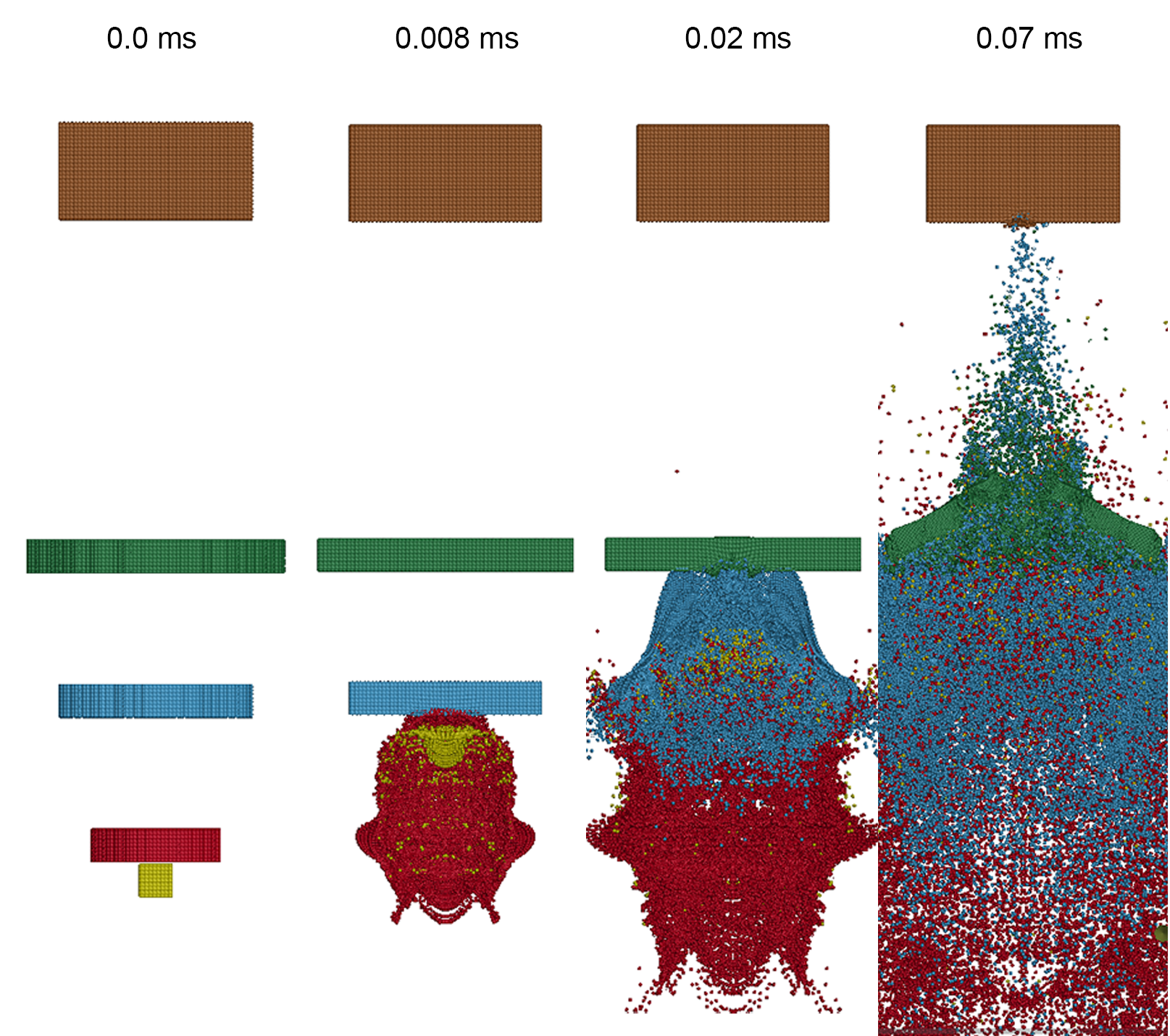}
\par\end{centering}
\caption{Series of frames from an LS-DYNA simulation showing the impact of
a steel cubic particle (yellow) at 7 km/s against a shield consisting
of three 1.0 cm thick plates of AA6061-T651 (red - outer bumper, blue
- inner bumper, green - rear wall) with a 3.0 mm thick AA6061-T651
witness plate (brown) positioned 10 cm behind the rearmost surface
of the shield. The projectile is shown to fragment into a dispersed
cloud of projectile and shield fragments which radially disperse before
impacting upon the rear wall. The rear wall is perforated and fragments
are observed to crater the witness plate, providing a non-zero depth
of penetration measurement.}

\label{fig:dynaFrames}
\end{figure}

In Figure \ref{fig:dynaFrames} a series of frames from a representative
LS-DYNA simulation are provided, depicting the impact of the cubic
steel projectile against a three-wall shield design. 

Simulations are performed on AMD EPYC servers with 64 CPU cores (2.25
GHz) and 1 TB of RAM. All simulations are performed in parallel on
4 CPU cores and require 20-40 CPU hours to run, depending on the complexity
(in this case thickness) of the target plates. The simulation models
included a 3.0 cm thick aluminium alloy witness plate located 10.0
cm from the rear surface of the target rear wall (Plate 3 in Figure
\ref{fig:whipple-problem-schematic}) to measure residual penetration
in the event that the shield was perforated. Results were recorded
as a binary pass/fail related to non-perforation or perforation of
the target rear wall, respectively, together with a continuous Depth
of Penetration (DoP) measurement into the witness plate. Our objective
is to design a protective shield that can defeat the projectile threat
for minimal weight.

Three initial designs are evaluated by the human expert, the details
of which are provided in Table \ref{tab:Results} together with the
simulation results. Based on these results we assume that target designs
with areal weights significantly less than 5 $kg/m^{2}$ are likely
infeasible, while designs significantly heavier ($>15 kg/m^{2}$)
are not of interest. Within this weight range the design space includes
577,365 potential options. We perform the optimization in iterative
batches of size 2, with one suggestion from the BO and one from the
human expert.

\subsubsection{Difference to a classical CS setting }

We work with one expert and this differs from usual CS settings where
multiple experts perform the same task because: 
\begin{enumerate}
\item \emph{Level of expertise required is high, and access is difficult:
}The design of shields for protecting against space debris impact
at hypervelocity is a highly specialized discipline typically limited
to national space agencies or their primary contractors. For instance,
shielding onboard the ISS was predominantly developed by NASA and
Boeing for US modules, ROSCOSMOS and RKK Energia for Russian modules,
and JAXA for the Japanese module. Shielding on the European Columbus
module was designed primarily by Alenia Aerospazio under contract
to the European Space Agency, but borrowed heavily from the NASA designs
(see e.g.,\cite{destefanis1999}). Therefore, recruiting multiple
experts is a formidable task. 
\item \emph{The cost of experiments is high: }Due to the cost and access
limitations on experimental facilities, we utilize numerical simulations
for this design study. Such simulations are difficult to design and
validate, our expert (with 20 years experience on such codes) required
about 120 hours to build the simulation models. In addition, the simulations
can be computationally expensive, requiring on the order of 100-200
CPU hours per simulation for a moderate CPU. 
\item \emph{The design problem is hard: }There exists no state-of-the-art
solution for the defined problem. The nearest analogue is a shield
designed for a spherical aluminium projectile, for which the state-of-the-art
is a stuffed Whipple shield. This shielding configuration has been
used to define our optimization variables. Existing semi-analytical
penetration laws, such as those in \cite{christiansen1995}, are not
valid for application with non-spherical or non-aluminium projectiles. 
\item \emph{The expert cannot repeat the same design task }as they learn
during the first experiment. 
\end{enumerate}
All these factors mean that for a real experiment we can only show
how BO-Muse helps a single expert for a new problem for which there
is no state-of-art solution.

\subsubsection{Results}

Results of the experiment are provided in Table \ref{tab:Results}.
We can observe that the human expert initially explored designs similar
to the stuffed Whipple concept (i.e., metallic outer bumper, KE or
PE inner bumper, and metallic rear wall, with the inner bumper being
roughly located at the mid-point between the two metallic plates).
By design ID 9 the expert had identified a stuffed Whipple design
that was successful at defeating the projectile, with an areal weight
of 14.3 $kg/m^{2}$. Between design IDs 9 and 19 we can observe the
expert exploiting this successful design to identify a lower weight
solution, without success. Up to this point the expert does not seem
to have been influenced by the BO suggestions. By design ID 21, however,
we can observe the expert beginning to exploit BO suggestions, with
design ID 21 a modification of ID 5 and 14, design ID 23 a modification
of 18, and so on. Design ID 27, an expert suggestion that is an exploitation
of the BO suggested ID 22, was successful at defeating the projectile,
albeit at a higher weight than the previously identified solutions
at ID 9 and 11 (16.9 $kg/m^{2}$). The final expert design, ID 29,
is a further exploitation of ID 27 intended to reduce weight and
is found to be the best solution identified during the experiment,
with an areal weight of 13.8 $kg/m^{2}$.

The BO-Muse design ID 22 and subsequent human exploitation (design
IDs 27 and 29) are, according to our human expert, highly unusual
configurations for spacecraft debris shielding that they would not
have otherwise considered if not for the BO suggestion. The conventional
design methodology based on typical shields used in flight hardware
suggest that the outer bumper have a density and a shock impedance
comparable to that of the projectile and be sized such that the shock
rarefaction and tensile release wave superimpose towards the back
of the projectile, maximizing fragmentation and radial dispersion.
An internal fabric layer, such as that used in stuffed Whipple shields
(see e.g., \cite{christiansen1995}) is then intended to catch and
decelerate projectile fragments prior to impact upon the shield rear
wall. This general design principle has been established through ongoing
investigation since the Apollo program and matured for the International
Space Station with significant, proven success. Design ID 22 and the
subsequent designs (IDs 27 and 29) are a substantial deviation from
these established principles and at current it is unclear why they
have been successful - further investigation is needed.

\textbf{}
\begin{table}
\begin{centering}
	\textbf{\caption{Results of the spacecraft protective design optimization experiment. Bold indicates best. (SC - space claim).\label{tab:Results}}}
{\tiny{}}%
\begin{tabular}{|c|c|c|c|c|c|c|c|c|c|c|c|c|c|}
\hline 
\multirow{2}{*}{{\tiny{}ID}} & \multirow{2}{*}{{\tiny{}Source}} & \multicolumn{2}{c|}{{\tiny{}Plate 1}} & {\tiny{}Gap1} & \multicolumn{2}{c|}{{\tiny{}Plate 2}} & {\tiny{}Gap2} & \multicolumn{2}{c|}{{\tiny{}Plate 3}} & {\tiny{}SC} & {\tiny{}Weight} & \textbf{\tiny{}Result} & \multicolumn{1}{c|}{{\tiny{}DoP }}\tabularnewline
 &  & {\tiny{}Mat} & {\tiny{}$t$(cm)} & {\tiny{}(cm)} & {\tiny{}Mat} & {\tiny{}$t$(cm)} & {\tiny{}(cm)} & {\tiny{}Mat} & {\tiny{}t(cm)} & {\tiny{}(cm)} & {\tiny{}($kg/m^{2})$} & {\tiny{}{[}1/0{]}} & {\tiny{}(cm)}\tabularnewline
\hline 
\hline 
{\tiny{}1} & {\tiny{}expert} & {\tiny{}ST} & {\tiny{}0.2} & {\tiny{}2.0} & {\tiny{}ST} & {\tiny{}0.2} & {\tiny{}5.0} & {\tiny{}ST} & {\tiny{}0.5} & {\tiny{}7.9} & {\tiny{}7.1} & {\tiny{}1} & {\tiny{}2.89}\tabularnewline
\hline 
{\tiny{}2} & {\tiny{}expert} & {\tiny{}AL} & {\tiny{}1.0} & {\tiny{}3.5} & {\tiny{}AL} & {\tiny{}1.0} & {\tiny{}3.0} & {\tiny{}AL} & {\tiny{}1.0} & {\tiny{}9.5} & {\tiny{}8.1} & {\tiny{}1} & {\tiny{}0.31}\tabularnewline
\hline 
{\tiny{}3} & {\tiny{}expert} & {\tiny{}ST} & {\tiny{}1.0} & {\tiny{}6.0} & {\tiny{}AL} & {\tiny{}1.0} & {\tiny{}1.0} & {\tiny{}AL} & {\tiny{}1.0} & {\tiny{}10.0} & {\tiny{}13.3} & {\tiny{}1} & {\tiny{}0.01}\tabularnewline
\hline 
\textcolor{blue}{\tiny{}4} & {\tiny{}BO} & {\tiny{}ST} & {\tiny{}0.9} & {\tiny{}0.0} & {\tiny{}AL} & {\tiny{}0.9} & {\tiny{}7.0} & {\tiny{}AL} & {\tiny{}0.6} & {\tiny{}9.4} & {\tiny{}11.1} & {\tiny{}1} & {\tiny{}0.43}\tabularnewline
\hline 
\textcolor{blue}{\tiny{}5} & {\tiny{}expert} & {\tiny{}AL} & {\tiny{}0.5} & {\tiny{}5.0} & {\tiny{}KE} & {\tiny{}1.0} & {\tiny{}2.0} & {\tiny{}AL} & {\tiny{}1.0} & {\tiny{}9.5} & {\tiny{}5.3} & {\tiny{}1} & {\tiny{}0.22}\tabularnewline
\hline 
\textcolor{blue}{\tiny{}6} & {\tiny{}BO} & {\tiny{}ST} & {\tiny{}0.3} & {\tiny{}8.0} & {\tiny{}PE} & {\tiny{}0.6} & {\tiny{}0.0} & {\tiny{}AL} & {\tiny{}0.9} & {\tiny{}9.8} & {\tiny{}5.4} & {\tiny{}1} & {\tiny{}0.31}\tabularnewline
\hline 
\textcolor{blue}{\tiny{}7} & {\tiny{}expert} & {\tiny{}ST} & {\tiny{}0.3} & {\tiny{}5.0} & {\tiny{}KE} & {\tiny{}1.0} & {\tiny{}2.0} & {\tiny{}ST} & {\tiny{}1.0} & {\tiny{}9.3} & {\tiny{}11.4} & {\tiny{}1} & {\tiny{}0.07}\tabularnewline
\hline 
\textcolor{blue}{\tiny{}8} & {\tiny{}BO} & {\tiny{}ST} & {\tiny{}0.5} & {\tiny{}7.0} & {\tiny{}ST} & {\tiny{}0.9} & {\tiny{}0.0} & {\tiny{}AL} & {\tiny{}0.9} & {\tiny{}9.3} & {\tiny{}13.4} & {\tiny{}1} & {\tiny{}0.68}\tabularnewline
\hline 
\textbf{\textcolor{blue}{\tiny{}9}} & \textbf{\tiny{}expert} & \textbf{\tiny{}ST} & \textbf{\tiny{}0.7} & \textbf{\tiny{}4.0} & \textbf{\tiny{}PE} & \textbf{\tiny{}1.0} & \textbf{\tiny{}3.0} & \textbf{\tiny{}ST} & \textbf{\tiny{}1.0} & \textbf{\tiny{}9.7} & \textbf{\tiny{}14.3} & \textbf{\tiny{}0} & \textbf{\tiny{}0.0}\tabularnewline
\hline 
\textcolor{blue}{\tiny{}10} & {\tiny{}BO} & {\tiny{}AL} & {\tiny{}0.9} & {\tiny{}0.0} & {\tiny{}KE} & {\tiny{}0.2} & {\tiny{}8.0} & {\tiny{}ST} & {\tiny{}0.5} & {\tiny{}9.6} & {\tiny{}6.6} & {\tiny{}1} & {\tiny{}1.90}\tabularnewline
\hline 
\textbf{\textcolor{blue}{\tiny{}11}} & \textbf{\tiny{}expert} & \textbf{\tiny{}ST} & \textbf{\tiny{}0.7} & \textbf{\tiny{}4.0} & \textbf{\tiny{}KE} & \textbf{\tiny{}1.0} & \textbf{\tiny{}3.0} & \textbf{\tiny{}ST} & \textbf{\tiny{}1.0} & \textbf{\tiny{}9.7} & \textbf{\tiny{}14.6} & \textbf{\tiny{}0} & \textbf{\tiny{}0.0}\tabularnewline
\hline 
\textcolor{brown}{\tiny{}12} & {\tiny{}BO} & {\tiny{}PE} & {\tiny{}0.2} & {\tiny{}8.0} & {\tiny{}AL} & {\tiny{}0.9} & {\tiny{}0.0} & {\tiny{}ST} & {\tiny{}0.9} & {\tiny{}10.0} & {\tiny{}9.7} & {\tiny{}1} & {\tiny{}2.84}\tabularnewline
\hline 
\textcolor{brown}{\tiny{}13} & {\tiny{}expert} & {\tiny{}ST} & {\tiny{}0.6} & {\tiny{}4.0} & {\tiny{}KE} & {\tiny{}1.0} & {\tiny{}3.0} & {\tiny{}ST} & {\tiny{}0.9} & {\tiny{}9.5} & {\tiny{}13.0} & {\tiny{}1} & {\tiny{}0.01}\tabularnewline
\hline 
\textcolor{brown}{\tiny{}14} & {\tiny{}BO} & {\tiny{}-} & {\tiny{}-} & {\tiny{}0.0} & {\tiny{}KE} & {\tiny{}0.9} & {\tiny{}0.0} & {\tiny{}ST} & {\tiny{}0.9} & {\tiny{}1.8} & {\tiny{}8.2} & {\tiny{}1} & {\tiny{}0.84}\tabularnewline
\hline 
\textcolor{brown}{\tiny{}15} & {\tiny{}expert} & {\tiny{}ST} & {\tiny{}0.5} & {\tiny{}4.0} & {\tiny{}KE} & {\tiny{}1.0} & {\tiny{}3.0} & {\tiny{}ST} & {\tiny{}1.0} & {\tiny{}9.5} & {\tiny{}13.0} & {\tiny{}1} & {\tiny{}0.01}\tabularnewline
\hline 
\textcolor{brown}{\tiny{}16} & {\tiny{}BO} & {\tiny{}-} & {\tiny{}-} & {\tiny{}0.0} & {\tiny{}ST} & {\tiny{}0.9} & {\tiny{}9.0} & {\tiny{}AL} & {\tiny{}0.1} & {\tiny{}10.0} & {\tiny{}7.3} & {\tiny{}1} & {\tiny{}0.32}\tabularnewline
\hline 
\textcolor{brown}{\tiny{}17} & {\tiny{}expert} & {\tiny{}ST} & {\tiny{}0.7} & {\tiny{}4.0} & {\tiny{}KE} & {\tiny{}1.0} & {\tiny{}3.0} & {\tiny{}ST} & {\tiny{}0.8} & {\tiny{}9.5} & {\tiny{}13.0} & {\tiny{}1} & {\tiny{}0.01}\tabularnewline
\hline 
\textcolor{brown}{\tiny{}18} & {\tiny{}BO} & {\tiny{}AL} & {\tiny{}0.9} & {\tiny{}8.0} & {\tiny{}ST} & {\tiny{}0.1} & {\tiny{}0.0} & {\tiny{}AL} & {\tiny{}0.9} & {\tiny{}9.9} & {\tiny{}5.6} & {\tiny{}1} & {\tiny{}0.39}\tabularnewline
\hline 
\textcolor{brown}{\tiny{}19} & {\tiny{}expert} & {\tiny{}ST} & {\tiny{}0.5} & {\tiny{}6.0} & {\tiny{}PE} & {\tiny{}1.0} & {\tiny{}1.0} & {\tiny{}ST} & {\tiny{}1.0} & {\tiny{}9.5} & {\tiny{}12.8} & {\tiny{}1} & {\tiny{}0.06}\tabularnewline
\hline 
\textcolor{green}{\tiny{}20} & {\tiny{}BO} & {\tiny{}PE} & {\tiny{}0.0} & {\tiny{}0.0} & {\tiny{}ST} & {\tiny{}0.4} & {\tiny{}8.0} & {\tiny{}AL} & {\tiny{}0.9} & {\tiny{}9.3} & {\tiny{}5.6} & {\tiny{}1} & {\tiny{}0.31}\tabularnewline
\hline 
\textcolor{green}{\tiny{}21} & {\tiny{}expert} & {\tiny{}ST} & {\tiny{}0.7} & {\tiny{}0.0} & {\tiny{}KE} & {\tiny{}1.0} & {\tiny{}7.0} & {\tiny{}ST} & {\tiny{}1.0} & {\tiny{}9.7} & {\tiny{}14.6} & {\tiny{}1} & {\tiny{}0.33}\tabularnewline
\hline 
\textcolor{green}{\tiny{}22} & {\tiny{}BO} & {\tiny{}KE} & {\tiny{}0.9} & {\tiny{}0.0} & {\tiny{}ST} & {\tiny{}0.9} & {\tiny{}8.0} & {\tiny{}ST} & {\tiny{}0.1} & {\tiny{}9.9} & {\tiny{}9.0} & {\tiny{}1} & {\tiny{}0.26}\tabularnewline
\hline 
\textcolor{green}{\tiny{}23} & {\tiny{}expert} & {\tiny{}ST} & {\tiny{}1.0} & {\tiny{}5.0} & {\tiny{}KE} & {\tiny{}1.0} & {\tiny{}2.0} & {\tiny{}AL} & {\tiny{}1.0} & {\tiny{}10.0} & {\tiny{}11.8} & {\tiny{}1} & {\tiny{}0.11}\tabularnewline
\hline 
\textcolor{green}{\tiny{}24} & {\tiny{}BO} & {\tiny{}-} & {\tiny{}-} & {\tiny{}0.0} & {\tiny{}ST} & {\tiny{}0.9} & {\tiny{}0.0} & {\tiny{}ST} & {\tiny{}0.1} & {\tiny{}1.0} & {\tiny{}7.9} & {\tiny{}1} & {\tiny{}0.82}\tabularnewline
\hline 
\textcolor{green}{\tiny{}25} & {\tiny{}expert} & {\tiny{}ST} & {\tiny{}0.4} & {\tiny{}6.0} & {\tiny{}AL} & {\tiny{}1.0} & {\tiny{}1.0} & {\tiny{}ST} & {\tiny{}1.0} & {\tiny{}9.4} & {\tiny{}13.7} & {\tiny{}1} & {\tiny{}0.03}\tabularnewline
\hline 
\textcolor{green}{\tiny{}26} & {\tiny{}BO} & {\tiny{}-} & {\tiny{}-} & {\tiny{}0.0} & {\tiny{}PE} & {\tiny{}0.1} & {\tiny{}9.0} & {\tiny{}ST} & {\tiny{}0.9} & {\tiny{}10.0} & {\tiny{}7.2} & {\tiny{}1} & {\tiny{}1.53}\tabularnewline
\hline 
\textbf{\textcolor{green}{\tiny{}27}} & \textbf{\tiny{}expert} & \textbf{\tiny{}KE} & \textbf{\tiny{}1.0} & \textbf{\tiny{}0.0} & \textbf{\tiny{}ST} & \textbf{\tiny{}1.0} & \textbf{\tiny{}7.0} & \textbf{\tiny{}ST} & \textbf{\tiny{}1.0} & \textbf{\tiny{}10.0} & \textbf{\tiny{}16.9} & \textbf{\tiny{}0} & \textbf{\tiny{}0.0}\tabularnewline
\hline 
{\tiny{}28} & {\tiny{}BO} & {\tiny{}AL} & {\tiny{}0.9} & {\tiny{}0.0} & {\tiny{}ST} & {\tiny{}0.1} & {\tiny{}8.0} & {\tiny{}AL} & {\tiny{}0.9} & {\tiny{}9.9} & {\tiny{}5.6} & {\tiny{}1} & {\tiny{}1.75}\tabularnewline
\hline 
\textbf{\tiny{}29} & \textbf{\tiny{}expert} & \textbf{\tiny{}KE} & \textbf{\tiny{}1.0} & \textbf{\tiny{}0.0} & \textbf{\tiny{}ST} & \textbf{\tiny{}0.8} & \textbf{\tiny{}7.0} & \textbf{\tiny{}ST} & \textbf{\tiny{}0.8} & \textbf{\tiny{}9.6} & \textbf{\tiny{}13.8} & \textbf{\tiny{}0} & \textbf{\tiny{}0.0}\tabularnewline
\hline 
\end{tabular}{\tiny\par}
\par\end{centering}

\end{table}

Evaluating this experiment - the role of BO-Muse was to inject novelty
in the design process, from which the human expert could take inspiration
and perform exploitation. We consider this to have been successfully
demonstrated in a real applied engineering design experiment

\subsection{Discussion of Limitations}

In this section we present a brief discussion of the limitations of
our work. With regard to the human expert, we have assumed that ``Cognitive
entrenchment'' behavior occurs, which is backed by recent studies
\citet{dane2010reconsidering,daw2006cortical}. This may not hold
strictly in all cases, which may cause the algorithm's sample efficiency
to be lower than expected as BO may over-compensate for expected expert
over-exploitation that does not eventuate. Similarly, we assume that
the expert is able to improve their model as the number of observations
available to them increases. However a less skilled expert may fail
to do this, and subsequently the algorithm's sample efficiency may
suffer as, after a point, expert suggestions may cease to be useful.
Finally, the human expert may not behave precisely like a GP-UCB model
would suggest, again resulting in lower efficiency. We note, however,
that even when the human is unable to perform as expected, BO-Muse
will still have sub-linear convergence. In such a worst-case scenario
every second experiment will, in effect, be wasted. However the data
from these ``wasted'' experiments will still provide additional observations
of $f^{\star}$ for the machine's GP model, which can only improve
the model's accuracy. The behavior of the algorithm in this case
can therefore be analyzed in the ``machine running GP-UCB plus improved
prior due to additional data'' regime - that is, a GP-UCB algorithm
with AI-generated suggestions, an exploration parameter $\beta$ that
is increased by a constant multiplicative factor, and a stream of
additional (harmless or even potentially informative) human-guided
experimental observations - which suffices to ensure sub-linear convergence
\citet{Sri1}. We also note that the aforesaid worst-case scenario
is highly unlikely on the assumption that the human expert is knowledgeable
in the relevant field.

\end{document}